\documentclass[manuscript,nonacm]{acmart}

\setcopyright{acmcopyright}
\copyrightyear{2024}
\acmYear{2024}
\acmDOI{XXXXXXX.XXXXXXX}

\usepackage{subcaption}
\usepackage{cleveref}
\usepackage{csquotes}
\usepackage{listings}
\usepackage{xcolor}
\usepackage{amsmath}
\usepackage{tabularx}
\usepackage{wrapfig}
\usepackage{makecell}
\usepackage[colorinlistoftodos]{todonotes}
\usepackage{pdfpages}
\usepackage{tcolorbox}

\definecolor{pillblue}{HTML}{E0F2FF} %
\definecolor{pillborder}{HTML}{5B9BD5} %
\definecolor{pilltext}{HTML}{1F4E79} %

\newcommand{\inlinequote}[2]{P#1:~``\emph{#2}''}
\newcommand\quoteparticipant[2]{\begin{quote}\inlinequote{#1}{#2}\end{quote}}

\newcommand{\pill}[1]{%
  \fcolorbox{pillborder}{pillblue}{\textcolor{pilltext}{\footnotesize\textbf{#1}}}%
}

\definecolor{darkblue}{HTML}{1F4E79}
\definecolor{mutedgreen}{RGB}{66,124,65} %
\definecolor{amber}{RGB}{255,191,0}
\definecolor{lightgray}{gray}{0.95}

\lstdefinelanguage{cnd}{
    keywords={left, right, above, below, directlyLeft, directlyRight, directlyAbove, directlyBelow, clockwise, counterclockwise},
    keywords=[2]{group, orientation, cyclic},
    keywords=[3]{icon, projection, attribute, flag},
    keywords=[4]{hideDisconnected, hideDisconnectedBuiltIn},
    sensitive=false,
    keywordstyle=[2]\color{mutedgreen}\bfseries,
    keywordstyle=[3]\color{darkblue}\bfseries,
    basicstyle=\tiny\ttfamily
}

\begin{document}

\title{Grounded Language Design for \\ Lightweight Diagramming for Formal Methods}

\newcommand{\cnd}{\texttt{CnD}}
\newcommand{\cndFull}{Cope and Drag}
\newcommand{\lfs}{LFS}
\newcommand{\ASV}{Alloy Default Visualizer}
\newcommand{\ASVs}{Alloy Default Visualizers}
\newcommand{\SDT}{Sterling-with-D3}

\author{Siddhartha Prasad}
\orcid{0000-0001-7936-8147}
\affiliation{
  \department{Computer Science Department}
  \institution{Brown University}
  \city{Providence}
  \state{Rhode Island}
  \postcode{02912}
  \country{USA}
}
\email{siddhartha_prasad@brown.edu}

\author{Ben Greenman}
\orcid{0000-0001-7078-9287}
\affiliation{
  \department{Kahlert School of Computing}
  \institution{University of Utah}
  \city{Salt Lake City}
  \state{Utah}
  \postcode{84112}
  \country{USA}
}
\email{benjamin.l.greenman@gmail.com}

\author{Tim Nelson}
\orcid{0000-0002-9377-9943}
\affiliation{
  \department{Computer Science Department}
  \institution{Brown University}
  \city{Providence}
  \state{Rhode Island}
  \postcode{02912}
  \country{USA}
}
\email{timothy_nelson@brown.edu}

\author{Shriram Krishnamurthi}
\orcid{0000-0001-5184-1975}
\affiliation{
  \department{Computer Science Department}
  \institution{Brown University}
  \city{Providence}
  \state{Rhode Island}
  \postcode{02912}
  \country{USA}
}
\email{shriram@brown.edu}

\newcommand{\code}[1]{\texttt{#1}}

\begin{abstract}

  Model finding, as embodied by SAT solvers and similar tools, is used
  widely, both in embedding settings and as a tool in its own
  right. For instance, tools like Alloy target SAT to enable users to
  incrementally define, explore, verify, and diagnose sophisticated specifications
  for a large number of complex systems.

  These tools critically include a visualizer that lets users graphically
  explore these generated models. As we show, however, default
  visualizers, which know nothing about the domain, are unhelpful and
  even actively violate presentational and cognitive principles. At
  the other extreme, full-blown visualizations require significant
  effort as well as knowledge a specifier might not possess; they can
  also exhibit bad failure modes (including silent failure).

  Instead, we need a \emph{language} to capture essential domain
  information for \emph{lightweight} diagramming. We ground our
  language design in both the
  cognitive science literature on diagrams and on a large number of example
  custom visualizations. This identifies the key elements of lightweight
  diagrams. We distill these into a small set of orthogonal
  primitives. We extend an Alloy-like tool to support these
  primitives. We evaluate
  the effectiveness of the produced diagrams,
  finding them good for reasoning. We then compare this against many
  other drawing languages and tools to show that this work defines a
  new niche that is lightweight, effective, and driven by sound
  principles.
\end{abstract}



\maketitle

\begin{tcolorbox}[colback=blue!5!white, colframe=blue!75!black, title=This system is in continuous development.]
  This is version \textbf{1} of a paper about a system that is in continuous development.
  Before referencing this work, please check for a more recent version of the paper and the
  latest updates on the system to ensure the information is current.
\end{tcolorbox}

\begin{figure*}[ht]
    \centering
    \includegraphics[width=\textwidth]{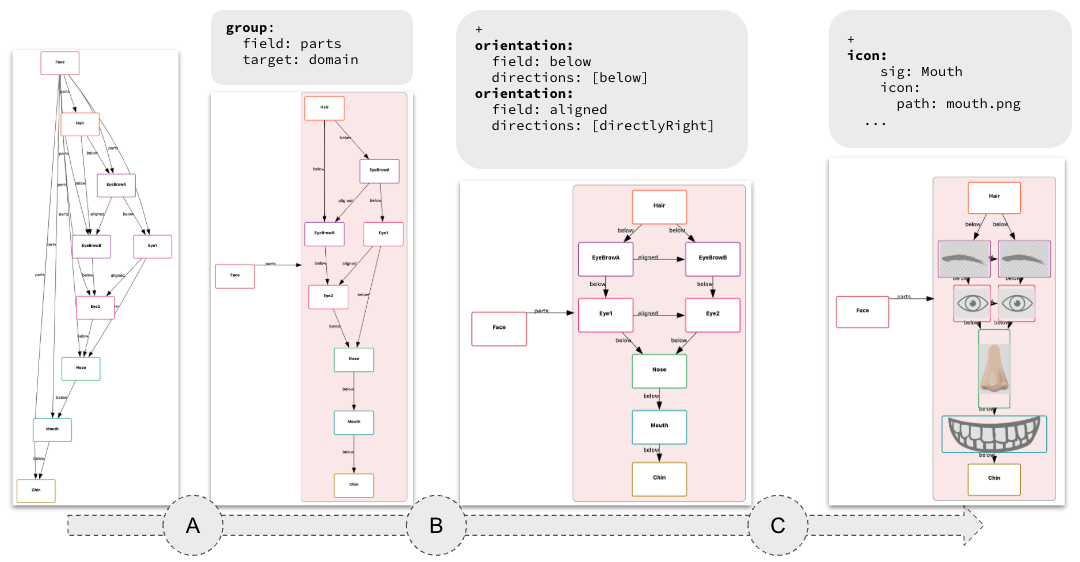}
    \caption{\cndFull{} is a language built to support lightweight diagramming for lightweight formal methods.
    A small set of orthogonal primitives allows users to incrementally refine
    model finder output into more effective diagrams. 
    This figure shows how grouping (A), orientation (B), and icons (C) can be incrementally applied to model-finder output describing a face.}
    \label{f:teaser}
\end{figure*}

\section{Introduction}
\label{s:intro}

SAT solvers are used to model and reason about increasingly
sophisticated domains. We especially focus on their embodiment in
tools like Alloy, a declarative modeling language that combines
first-order logic with transitive closure and a relational calculus
~\cite{j-alloy-2012}. Alloy embraces the \emph{lightweight formal
  methods} strategy described by Jackson and Wing~\cite{jackson1996lightweight}. In this
approach, users build models incrementally, exploring the design space
as they proceed. The automation provided by SAT is critical to this
process.

A central attribute of lightweight exploration is that users can
reason informally about their designs before doing so formally. 
Whereas traditional verification tools require \emph{properties}
before the tools can do anything effective, tools like Alloy can work
with just system descriptions: they use a SAT-solver to generate
instances of the system and present these to the user, most commonly
in visual form. (\cref{f:sterling-default-bt} shows an example.) This has two
benefits. First, users can (hopefully quickly) spot ways in which the
output does not conform to their intent, and improve their
specification. Second, seeing these violations
suggests desirable and undesirable \emph{properties}, which can then
be formalized and turned into verification conditions. Thus, realizing
lightweightness hinges heavily on the quality of visualizations. In
turn, once properties exist, counterexamples to their success are also
presented using the same mechanisms.

Alloy, and its sibling tool Forge~\cite{nelson2024forge}, rely on a visualizer that
draws relations as graphs~\cite{j-alloy-2012} (see \cref{f:filesystem-asv} for an example).
Recent work on a
new visualizer for both tools, Sterling~\cite{dyer2021sterling}, provides some
improvements (see \cref{f:sterling-default-bt} for an example)
but is largely indistinguishable from the perspective of
our work: both are \emph{generic}, in that they do not embody any
knowledge of the domain. While these do not demand any effort by the
user to create visualizations, research shows that users struggle to
make sense of Alloy visualizations as models grow in size and
complexity~\cite{mansoor2023empirical}.

In contrast, recent work~\cite{nelson2024forge} exploits the fact that Sterling is
programmable to create \emph{custom} visualizations. These suffer from
the dual problem. They can be narrowly crafted to a domain, and can be
quite visually attractive (as figure 11 in \cite{nelson2024forge} shows). However, as we
discuss in \cref{s:pain-barrier}, users must: know the details of drawing libraries,
which are orthogonal to formal methods; expend a great deal of effort;
and not inadvertently run afoul of good visual design principles.
Furthermore, they suffer from critical failure modes (\cref{s:when-viz-fail}) that are
especially dangerous in this context.

This paper views the visualization task as a \emph{language design
  problem}. That is, we want the power of customization that
programming provides, but we want it to: be lightweight; avoid rich
library dependence; embody good visual design principles; and prevent
bad failure modes. By embodying these in a programming language, we
can be confident that all programs written in that
language enjoy these traits.

We therefore present a new language, \cndFull{} (\cnd{}), for this
purpose. \cnd{} is driven both top-down, by cognitive science
principles of visualization, and bottom-up, by manually distilling
ideas from dozens of actual visualizations. We present a language that
is small, requires little annotation, and can be used
incrementally. We show through experiments that the features of the
language result in visualizations that are more effective than the
default visualizer. While this work falls in the realm of recent
language designs for visual output, we show why this work is novel and
particularly well-suited for this context.

Furthermore, we believe this work has much broader
applicability. While we have focused on lightweight formal methods in
Alloy, many other formal methods tools also employ
diagrams. Model-checkers present counterexample traces with states that beg to be
visualized; indeed, some like ProB~\cite{leuschel+butler:fme03:prob} already support 
graphical visualization of states. Similarly,
proof-assistants like Lean~\cite{demoura+ullrich:cade21:lean4} provide domain-specific
visualizations for structures like graphs, trees, 
and so on~\cite{lean:user-widgets,nawrocki++:itp23:proofwidgets} 
and for proofs as well~\cite{wren2021animate}. We believe the ideas of this paper can
profitably be explored in those settings as well.

\subsection*{Vocabulary}

\paragraph{Specifications, Instances, Bad-Instances}
A \textbf{specification} (or, in the lightweight spirit, \emph{spec}) is a formal description of a system's behavior (e.g., Alloy or Forge code).
An \textbf{instance} is a concrete assignment of values to variables that 
satisfies the constraints and properties defined by the spec.
A \textbf{bad instance} is an instance,
but fails to satisfy the \emph{implicit} spec that the author is \emph{trying} to write.
A bad instance reflects a discrepancy between what the explicit spec currently represents
and what the author intends it to represent. We use the word ``bad'' intentionally, since it is
human judgment that identifies such a discrepancy. The discovery and exploration of bad-instances
while driving towards the implicit spec is a key part of the lightweight formal methods process.

\paragraph{\ASVs{}}

The Alloy language and its variants support multiple default visualizers, including the
Alloy Visualizer~\cite{j-alloy-2012} and Sterling~\cite{dyer2021sterling}.
These default visualizers are agnostic to the domain being modeled,
and communicate instances via directed graphs. We discuss the
differences between these two tools in \cref{s:asv}, but for the
purposes of this paper, these tools are effectively the same, so we
refer to them by the general label of \ASVs{}.

\paragraph{Interactive Diagrams}
A powerful aspect of the \cnd{} diagrams we reference in this paper is their interactive nature.
While we include screenshots in this paper, we also provide access to these interactive 
diagrams in the accompanying supplement. The term \pill{name} indicates a diagram's availability in the
supplement: see \cref{s:supplement} for details.

\section{Language Design}
\label{s:design}

Languages can be designed from many perspectives, such as personal
taste, carefully subjecting every feature to a user study~\cite{maloney2010scratch,quorumlanguage,sime1977scope},
or programmer opinion, which can be broadly expressed or narrowly
targeted~\cite{tunnell2017can,tunnell2018behavior,stefik2013empirical}. While these all have their validity, we focus on
\emph{grounded} design: designs that are derived from, and can be
substantiated by, external artifacts.

There are roughly two ways to proceed. One is bottom-up: start with a
large collection of desired artifacts, cluster their commonalities and
differences, identify their principal dimensions, and use those to
inform the design. Another is top-down: start with general principles
established by experts and concretize them into operations. 
Because these approaches complement each other well, we use both methods.
We start top-down, distilling key principles from the diagramming literature (\cref{s:top-down}),
and then examine a corpus of concrete examples in the context of these principles (\cref{s:bottom-up}).
We then align these perspectives to derive a set of diagramming primitives
 that are both theoretically grounded and practically applicable in the context of lightweight formal methods (\cref{s:alignment}).

\subsection{Top Down}
\label{s:top-down}

We began our language design by drawing on established principles in cognitive science, visualization, and diagramming.

\subsubsection{Visual Principles}
\label{s:visual-principles}
A diagram is a visual representation that uses shapes and symbols to convey information or ideas. In this role as an information presentation tool, 
effective diagrams rely on perceptual properties common in information visualization: pre-attentive processing and the Gestalt principles~\cite{chen2017information}.

\paragraph{Pre-Attentive Processing}

Pre-attentive processing is a cognitive function that allows
humans to quickly detect certain visual features (e.g., color, shape, and size)
of objects without conscious effort.
This rapid processing can both help and hinder understanding.

The Stroop effect, for instance, demonstrates that the brain often automatically processes
demonstrates that the brain often automatically processes text, such as word recognition, faster than
stimuli such as color~\cite{stroop1935studies} and 
spatiality~\cite{spatialstroop}. 
In the context of diagrams, this means that understanding is hindered when visual elements conflict with textual content.
By laying out a binary tree's \texttt{left} children to the right and \texttt{right} children to the left, 
\cref{f:sterling-default-bt} lends itself to spatial Stroop effects.
Other common features that might cause undesirable pre-attentive processing include
irrelevant visual elements~\cite{tufte1983visual}, unnecessary overlapping lines ~\cite{purchase1997aesthetic},
or ambiguous labeling~\cite{chen2017information,naturaldiagramming}.
Effective diagrams, therefore, use visual features to highlight important information
while being mindful of extraneous cognitive load~\cite{sweller1991evidence}.

\paragraph{Gestalt Principles}

The Gestalt principles describe the ways in which humans perceive and organize visual information~\cite{koffka1922perception, chen2017information}:
\begin{enumerate}
  \item \textbf{Proximity:} Elements that are close to each other are perceived as a group.
  \item \textbf{Similarity:} Similar elements are perceived as part of the same group.
  \item \textbf{Closure:} Incomplete shapes are perceived as whole, even when part of the information is missing.
  \item \textbf{Continuity:} Elements arranged on a line or curve are perceived as related or belonging together.
  \item \textbf{Figure-Ground:} The visual field is perceived as divided into a foreground and a background.
  \item \textbf{Symmetry and Order:} Objects are perceived as symmetrical and organized around a central point.
  \item \textbf{Common Fate:} Elements moving in the same direction are perceived as part of a collective or unit.
\end{enumerate}
These principles play a crucial role in visual design, user interface design, and data visualization, 
helping to create more intuitive and easily comprehensible visual representations.

\subsubsection{Diagramming Principles}
\label{s:diagramming-lit}

While visualizations focus on displaying information, diagrams act as cognitive tools that 
facilitate deeper comprehension and analysis of complex 
systems~\cite{larkin1987diagram,spatialschemastversky, naturaldiagramming}.

A diagram's effectiveness stems not only from its visual appeal, but also from its ability 
to organize information in a way that facilitates
efficient cognitive processes~\cite{larkin1987diagram}.
These processes depend on the task one is trying to perform.

\begin{figure}[t]
  \centering
  \includegraphics[width=0.5\textwidth]{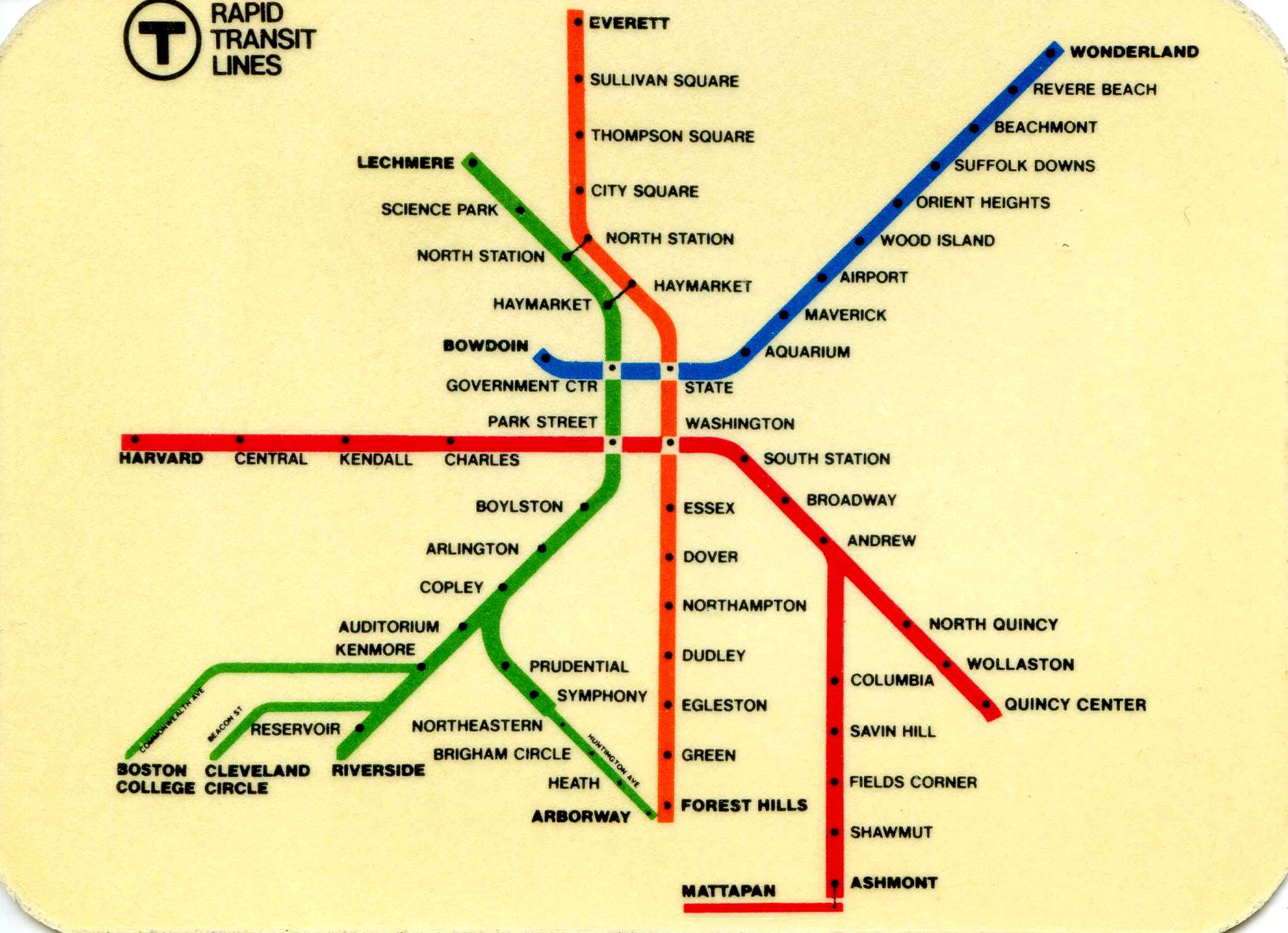}
  \caption{The Massachusetts Bay Transportation Authority (MBTA) transit map~\cite{MBTA1973Map} (public domain image from Wikimedia Commons)
  prioritizes ease of navigation over geographic precision.}
  \label{f:subway-mbta}
\end{figure}

Subway maps (\cref{f:subway-mbta}), for example, leverage the observation that semantic similarity
is intuitively tied to spatial proximity~\cite{goldstone1994efficient}.
By grouping together stations that are functionally related 
(e.g., transfer points or lines serving the same neighborhoods),
they serve as effective diagrams for planning public transportation routes.
However, this layout is poorly suited for understanding the actual geographic 
distances between stations, as the map distorts spatial relationships in 
favor of simplifying the navigation task.

Thus, while a ``good'' diagram must be well laid out,
it must also relate well to the \emph{problem} it is 
designed to address~\cite{naturaldiagramming, chen2017information}.
Diagrams are of help when they:
\begin{enumerate} 
  \item Leverage spatial metaphors and natural correspondences in the source domain~\cite{spatialschemastversky,larkin1987diagram}.
  \item Convey patterns and relations in the source domain in spatial terms~\cite{goldstone1994efficient, larkin1987diagram}.
  \item Group together all information that will be used together~\cite{larkin1987diagram,goldstone1994efficient}.
\end{enumerate}

\subsubsection{Diagramming Tools}
\label{s:diagramming-tools}

Having understood what makes a good diagram, we now examine the literature on how we can best support their creation.
In an exploration of how domain experts create diagrams, Ma'ayan et al. define a set of behaviors that should be supported 
by a successful diagramming tool~\cite{naturaldiagramming}. 

\begin{enumerate}
  \item Exploration Support: Users should be able to rapidly prototype and iterate on diagrams, both at a conceptual level and in terms of visual details. 
  The tool should allow users to easily undo, redo, and modify their diagrams as their understanding evolves.  
  \item Representation Salience: There should be a clear correspondence between conceptual elements in the spec and their visual representation.
  \item Live Engagement: Tools should allow users to directly manipulate diagram elements and see real-time updates.
  \item Vocabulary Correspondence:  The tool's interface and functionality should align closely with domain-specific concepts familiar to users.
\end{enumerate}

\subsubsection{Core Diagramming Operations}

From the above literature, we identify that users require the following operations in order to create effective diagrams:

\paragraph{Relative Positioning} Users should be able to specify the relative positions of diagram elements.
This allows diagram elements to be spatially arranged to reflect the underlying structure of the domain, relating to the Gestalt principles
of continuity, closure, and proximity. Placing graph elements consistently in relation to others can imply diagrammatic flow, groupings, 
and even underlying shapes. We identify two key aspects of relative positioning:
\begin{enumerate}
  \item Directional relationships between diagram elements. For example, specifying that elements should be placed above, below, to the left, or to the right of one another.
  \item Angular flow to describe element positions. For example, specifying that elements should be arranged in a circle, or that they should form a shape.
\end{enumerate}

\paragraph{Grouping} Users should be able to group related diagram elements together. Leveraging the Gestalt principle of proximity, 
grouping related elements together can help users to better understand the relationships between them. This also allows diagrammers to group together
elements that will be used together, reducing extraneous cognitive load. Grouping relationships in interactive diagrams can
also help users reason about higher-level structures and relationships, leveraging the Gestalt principle of common fate.

\paragraph{Styling} Users should be able to control some visual aspects of how diagram elements
are rendered. We identify two ways in which users might want to style their diagrams: theming and iconography.
Theming involves influencing the aesthetic properties of diagram elements, such as
the color, size, and shape of elements. This allows users to create diagrams that tap 
into the Gestalt principles of similarity and figure-ground. 

Iconography, on the other hand, involves the use of pictorial 
elements in diagrams that themselves carry semantic meaning: e.g.,
representing fruit using a fruit emoji (e.g., \cref{f:fruit-icons}). Icons and symbols in diagrams 
can tap into pre-attentive processing, allowing users to quickly understand 
what a diagram element represents without needing to examine the diagram in detail.

With these primitives, diagrammers should be able to create artifacts that:
\begin{itemize}
  \item Focus on important parts of the problem. 
  \item Are tools for knowledge presentation. That is, they offer insight into instances of the spec being explored.
  \item Act as tools for knowledge discovery. That is, they make salient key aspects of bad-instances of the spec.
\end{itemize}

\subsection{Bottom Up}
\label{s:bottom-up}

While the top-down approach provides a theoretical foundation for our language design, much of the literature
draws on abstract principles that may not be directly applicable to the formal methods domain.

To study how these principles manifest in practice, we contacted the Forge~\cite{nelson2024forge}
team, who kindly gave us access to anonymized data on visualizations that students created in their 
upper-level undergraduate course on formal methods, as well as the results of a follow-up survey on
students' visualization experiences. Students had given permission
for this use.\footnote{Some of these visualizations are shown in figure 11 of ~\cite{nelson2024forge}.}

The dataset contained anonymized retrospective comments on 58 final group projects
(21 from 2022, 21 from 2023, 
and 16 from 2024), most of which were written in Forge~\cite{nelson2024forge}, and visualized in either 
an \ASV{} or using D3 to customize Sterling. As a result, we were
able to gain insight into how both the \ASVs{} and \SDT{}
work for users who have roughly a semester of experience with formal modeling.

Projects covered a wide range of domains, from games like Solitaire to cryptographic protocols and 
group theory. Each project involved a code submission (a spec and optional visualization) as well as a recorded 
presentation or written summary explaining the domain being modeled, the design decisions made, and an optional 
demonstration of the visualization.

\begin{figure}[t]
  \centering
  \includegraphics[width=0.6\textwidth]{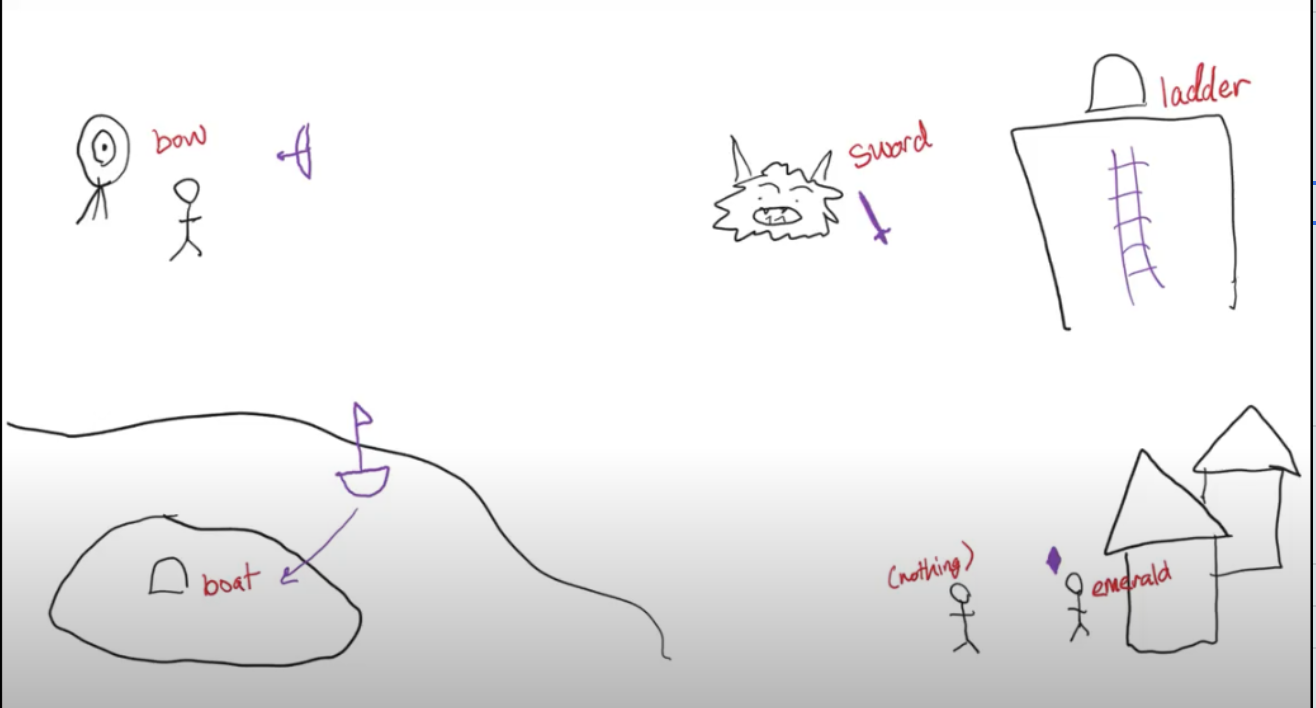}

  \caption{A student-drawn diagram illustrating 
  the layout of their spec. This layout relies on 
  relative positioning, grouped elements, 
  and iconography.}
  \label{f:handdrawn}
\end{figure}

47 (81\%) of these projects involved some kind of custom visualization (most typically relying on Sterling's D3 integration). 
We examined these projects to determine how the principles we derived from the cognitive science literature were reflected in the 
visualizations students created. We also included projects in our analysis that did not include a custom visualization, but where authors were
especially clear about wanting a custom visual component even if they did not
have one, e.g., by
presenting a hand-drawn diagram
(e.g., \cref{f:handdrawn}) or explicitly moving visualizer nodes around to explain their ideas.

In our analysis, we found that:
\begin{enumerate}
  \item 44 projects (75\%) involved custom visualizations that had clear representation salience and minimal visual clutter. Each atom in the model was represented by at most one 
  element in the visualization, and each element in the visualization corresponded to an important aspect of the model.  
  \item 41 visualizations (70\%) communicated semantic relationships between elements via their relative orientation. 
  Examples include the grid layout of cells when visualizing the games Minesweeper and Battleships, and the relative positions of nodes in red black trees.
  \item 19 visualizations (33\%) grouped nodes together inside explicit shapes to imply relatedness. Examples include piles of cards in the card game Solitaire, player hands in the game Chopsticks,
  and sets of elements in group theory.
  \item 4 visualizations (7\%) conveyed spec semantics by laying out diagram elements to imply shapes or directionality.
  Examples include the cyclic order of player turns in the card game Coup and travel paths in a map routing algorithm.
\end{enumerate}

A notable outlier that did not align with our diagramming principles was a music generation specification. It communicated instances via audio rather than purely visual means.

 The Forge team also shared anonymized results of a follow-up survey 
that they administered to students who took the course in 2023, asking them about their experiences creating custom visualizations.
Students explained that they created custom visualizations due to the
large, unstructured instances presented by the \ASVs{}. Concretely:

\quoteparticipant{62}{Too many edges and nodes were displayed so it was too difficult to decipher any meaningful information about what we were modeling}
\quoteparticipant{43}{For large traces, it was impossible to see what was happening simply because there were so many edges and nodes.}
\quoteparticipant{19}{When the state gets complicated, there becomes a lot of nodes and overlapping edges and labels. This gets really confusing and crowded – essentially making it unreadable.}
\quoteparticipant{23}{The graphs are essentially useless for more than 10 or so atoms in the model.}

In contrast, respondents valued that their custom visualizations hid unnecessary information, conveyed key aspects of the source domain, and
could even act as useful debugging tools.

\quoteparticipant{11}{Instead of seeing the names of objects which were randomly numbered and slighly confusing, the custom visualization eliminated the unnecessary parts and just displayed the values of the nodes}
\quoteparticipant{25}{I think it was actively helpful in quickly
  identifying whether or not a pattern was a valid
  pattern}%

\quoteparticipant{78}{... having the physical space laid out and important signals highlighted made it much quicker to verify certain properties being true.}

\subsection{Top Down and Bottom Up Alignment}
\label{s:alignment}

Our analysis of student projects suggests that diagramming principles from the cognitive science literature (\cref{s:top-down}),
such as reducing cognitive load and highlighting important
information,
are very much applicable in the context of lightweight formal methods.
Across a diverse range of domains, student visualizations preserved domain-specific spatial relations, 
grouped together related elements, and encoded spec semantics in terms of spatial layout and directionality.
They also reflect (as hinted by P25 above) a key additional requirement in the context of formal methods:
support for bad-instances. We discuss this in further detail in \Cref{s:when-viz-fail}.

\subsection{The Pain Barrier to Custom Visualization}
\label{s:pain-barrier}

We have motivated the value in having a custom visualization. However,
this comes at a heavy price.  Not only do users need familiarity with
JavaScript to use the library, but D3 itself is notorious for its
steep learning curve, API verbosity, and low-level
operations~\cite{battle2022exploring,nair2016interactive}.  (It is
also worth noting that both JavaScript and D3 are entirely orthogonal
learning objectives to the goal of doing formal methods.)  In addition
to these cognitive difficulties, \SDT{} scripts are often of
comparable size to the specs they are built to visualize.  For small
specifications, or those in development, the ratio is frequently
even worse. For example, the river-crossing puzzle visualization on
the Sterling website demo~\cite{dyer_sterling_js_demo} contains more
than \emph{four times} the lines of code (148) as the spec (merely
35). Furthermore, a typical custom visualization is an
``all-or-nothing'' proposition; lacking incrementality, it critically
lacks the ``pay as you go'' flavor that is characteristic of
lightweight formal methods.

This complexity can make it difficult for users to understand which
parts of their D3 code are responsible for which parts of the diagram.
Not only does this lead to brittle visualizations
(\cref{s:when-viz-fail}), but its heavyweight nature disincentivizes
the use of D3 for exploratory diagramming tasks.  In
\cref{s:bottom-up}, we have already indicated that some students used
paper drawings or manually dragged boxes to simulate what they
wanted. The same survey in which students praised the value of custom
visualizations also identified the complexity of D3 and the need to
use JavaScript as a significant barrier to building custom
visualizations, reporting, for example:

\quoteparticipant{15}{I [...] wasn't confident in my ability to learn Javascript to the extent necessary to write a custom visualization while also working on everything else.}
\quoteparticipant{24}{[Writing the visualization] would have been like 2x the amount of work}
\quoteparticipant{2}{We tried to start, but we had a lot of trouble figuring out how to do it [...] so we decided to focus on adding more properties to our model than spending too much time on it.}
\quoteparticipant{33}{It seemed really complicated and we prefered to spend more time trying to understand our model better and making sure it worked through tests than just visuals.}
\quoteparticipant{9}{ultimately we ran out of time/chose to spend it on other things like testing and debugging, but we did originally want to make our own visualization and got really confused when we tried to understand the visualization scripts we used in other projects/start making our own}
\quoteparticipant{19}{[...] spending a lot of time on the visualizer. [...] the visualization probably took 6-8 hours.}
\quoteparticipant{53}{lack of experience hindered me from getting exactly what I wanted}
\quoteparticipant{12}{We did not have enough time to invest in a visualizer.}

We thus see that the potential for custom visualization is undermined
by the current tooling, which is onerous and often too daunting for
users. (We discuss other drawing systems and languages in
\cref{s:rel-work}.) This observation demands a better, lightweight way
of capturing the \emph{essential} elements of diagrams that draws on
the above principles and examples.

\section{\cnd{}}
\label{s:cnd}

 \cndFull{} (\cnd{}) is a lightweight diagramming language designed for use with Alloy-like languages.
 The tool's name is inspired by the cope-and-drag casting process in metallurgy, reflecting its adaptable approach to diagram creation. 
 \cnd{} implements the diagramming operations identified in \cref{s:design}, and applies them to \ASV{}-like directed graphs.
 These generated diagrams support user interactions while maintaining consistency with the diagram specification.
 Thus, \cnd{} not only facilitates diagram creation, 
 but also supports interactive exploration of instances being visualized.

\subsection{\cnd{} Primitives}
\label{s:primitives}

\begin{figure}[t]

  \(
  \begin{array}[t]{l@{\hspace{1mm}}l}
    \begin{array}[t]{l}
      \mathsf{Program} \Coloneqq
        \textit{Constraint}^* \ \textit{Directive}^*
      \\
      \mathsf{Constraint} \Coloneqq
        \textit{CyclicConstraint} \\
        \quad \mid \textit{OrientationConstraint} \\
        \quad \mid \textit{GroupingConstraint}
      \\
      \mathsf{CyclicConstraint} \Coloneqq
        \textit{FieldName} \ \textit{FlowDirection}
      \\
      \mathsf{OrientationConstraint} \Coloneqq
        \textit{FieldName} \ \textit{Direction}^+ \\
        \quad \mid \textit{SigName} \ \textit{Direction}^+
      \\
      \mathsf{GroupingConstraint} \Coloneqq
        \textit{FieldName} \ \textit{Target}
      \\
      \mathsf{FlowDirection} \Coloneqq
        \textit{clockwise} \mid \textit{counterclockwise}\\
      \mathsf{Direction} \Coloneqq
      \textit{above} \mid \textit{below} \mid \textit{left} \mid \textit{right} \\
      \quad \mid \textit{directlyAbove} \mid \textit{directlyBelow} \\
      \quad \mid \textit{directlyLeft} \mid \textit{directlyRight}
      \\
      \mathsf{Target} \Coloneqq
        \textit{domain} \mid \textit{range}
    \end{array}
    &
    \begin{array}[t]{l}

      \mathsf{Directive} \Coloneqq
        \textit{PictorialDirective} \\
        \quad \mid \textit{ThemingDirective}
      \\
      \mathsf{PictorialDirective} \Coloneqq \\
        \quad \textit{SigName} \ \textit{IconDefinition}
      \\
      \mathsf{ThemingDirective} \Coloneqq \\
      \quad \textit{AttributeDirective} \mid \textit{SigColorDirective} \\
      \quad \mid \textit{ProjectionDirective} \mid \textit{VisibilityFlag}
    \\
    \mathsf{IconDefinition} \Coloneqq 
      \textit{path} \ \textit{height} \ \textit{width}
    \\
    \mathsf{AttributeDirective} \Coloneqq
      \textit{FieldName}
    \\
    \mathsf{SigColorDirective} \Coloneqq
      \textit{SigName} \ \textit{color}
    \\
    \mathsf{ProjectionDirective} \Coloneqq
      \textit{SigName}
    \\
    \mathsf{VisibilityFlag} \Coloneqq \textit{hideDisconnected} \\ 
      \quad \mid \textit{hideDisconnectedBuiltIns}
    \end{array}
  \end{array}
  \)

  \label{f:cnd-abstract-syntax}
  \caption{Abstract syntax of \cndFull{}.}
\end{figure}

\cnd{} primitives operate upon Alloy/Forge sigs (which are
essentially types) and their fields. These primitives fall into two categories: directives and constraints.
Directives are used to specify the visual representation of a sig or field, while constraints are used to specify the relationships between these visual representations.
The resultant diagrams are interactive, allowing users to drag nodes around and explore the spec, while ensuring that the visual representation remains consistent with 
the specified constraints. For instance, if a constraint specifies that \texttt{A} must be to the left of \texttt{B}, 
a user will not be allowed to drag \texttt{B} to the left of
\texttt{A} (\pill{ab}). In cases where constraints cannot be satisfied, no diagram
is produced (\cref{s:unsat}).

Each \cnd{} operation is designed to fulfill a specific requirement identified in \Cref{s:design}:

\paragraph{Relative Positioning: Cyclic and Orientation Constraints}

Cyclic and Orientation constraints are used to specify the \emph{relative positioning} (\cref{s:alignment}) of diagram elements.

\begin{figure}[h]
  \centering
  \begin{subfigure}{0.45\textwidth}
  \centering
  \includegraphics[height=0.4\textheight]{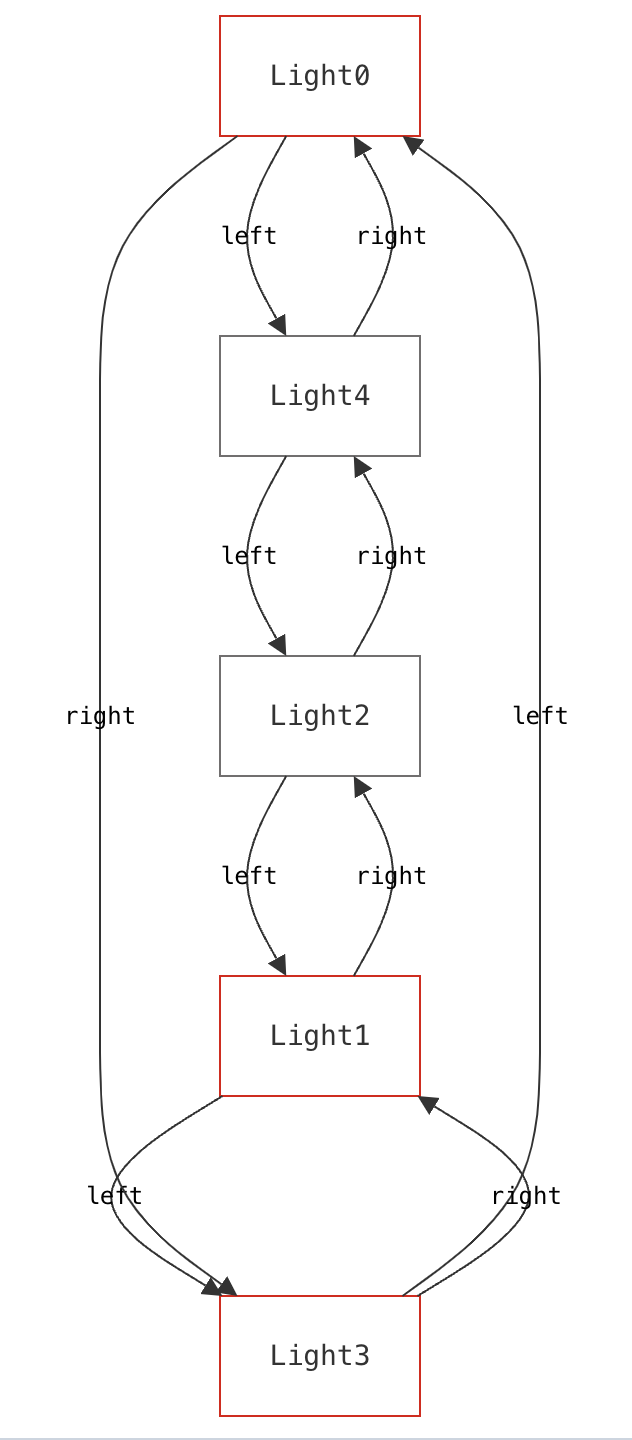}
  \caption{\ASV{}:
  This visualization obscures the spatial relationship between 
  \texttt{Light3} and \texttt{Light0}. The two lights are directly adjacent, with \texttt{Light3} to the right of \texttt{Light0}.}
  \label{f:asv-ring-lights}
  \end{subfigure}
  \hfill
  \begin{subfigure}{0.45\textwidth}
    \centering
    \includegraphics[width=\textwidth]{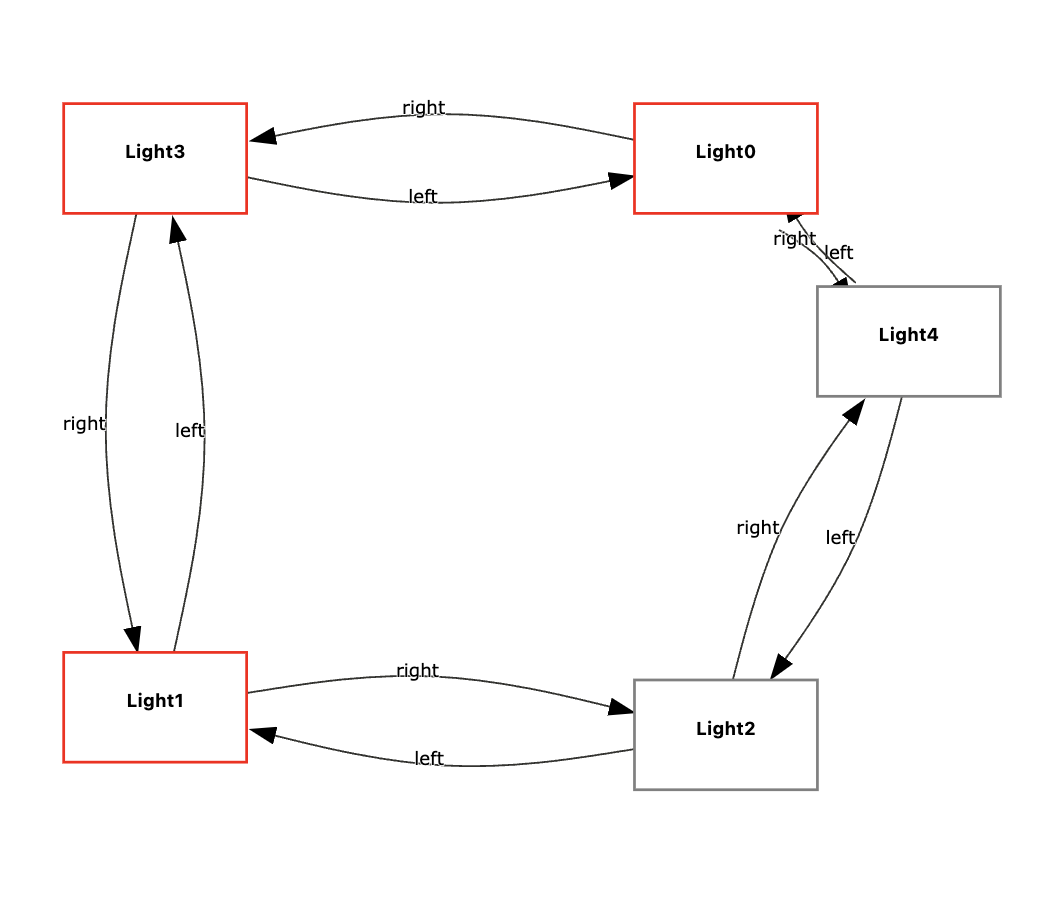}
    \caption{\cnd{} visualization using a cyclic constraint on the \texttt{left} field.}
    \label{f:lights-cyclic-example}
  \end{subfigure}\\

  \begin{subfigure}{0.45\textwidth}
    \centering
\begin{lstlisting}[language=cnd]
constraints:
  - cyclic:
      field: left
      direction: clockwise
directives:
  - flag: hideDisconnected
  - color:
      sig: Lit
      value: "red"
  - color:
      sig: Unlit
      value: "grey"
\end{lstlisting}
    \caption{\cnd{} spec}
  \end{subfigure}

  \caption{\pill{ring-lights} \ASV{} and \cnd{} visualizations of a light-switching puzzle 
  instance involving 5 lights in a ring from the Forge examples repository. 
  Lights that are lit are colored red, while unlit lights are colored grey.}
  \label{f:lights-cyclic}
\end{figure}

Cyclic constraints lay out atoms related by a sig field along the perimeter of a notional circle, suggesting they form a shape. 
If a field doesn't define a complete cycle, each connected subset is arranged on the circle's perimeter based on a depth-first exploration of the field. 
Atoms in the relation can be positioned either clockwise (the default) or counterclockwise.
\Cref{f:lights-cyclic} shows how a cyclic constraint can be used to visualize the spatial 
relationships in a diagram.

\begin{figure}[h]
  \centering
  \begin{subfigure}[b]{0.45\textwidth}
    \centering
    \includegraphics[width=\textwidth]{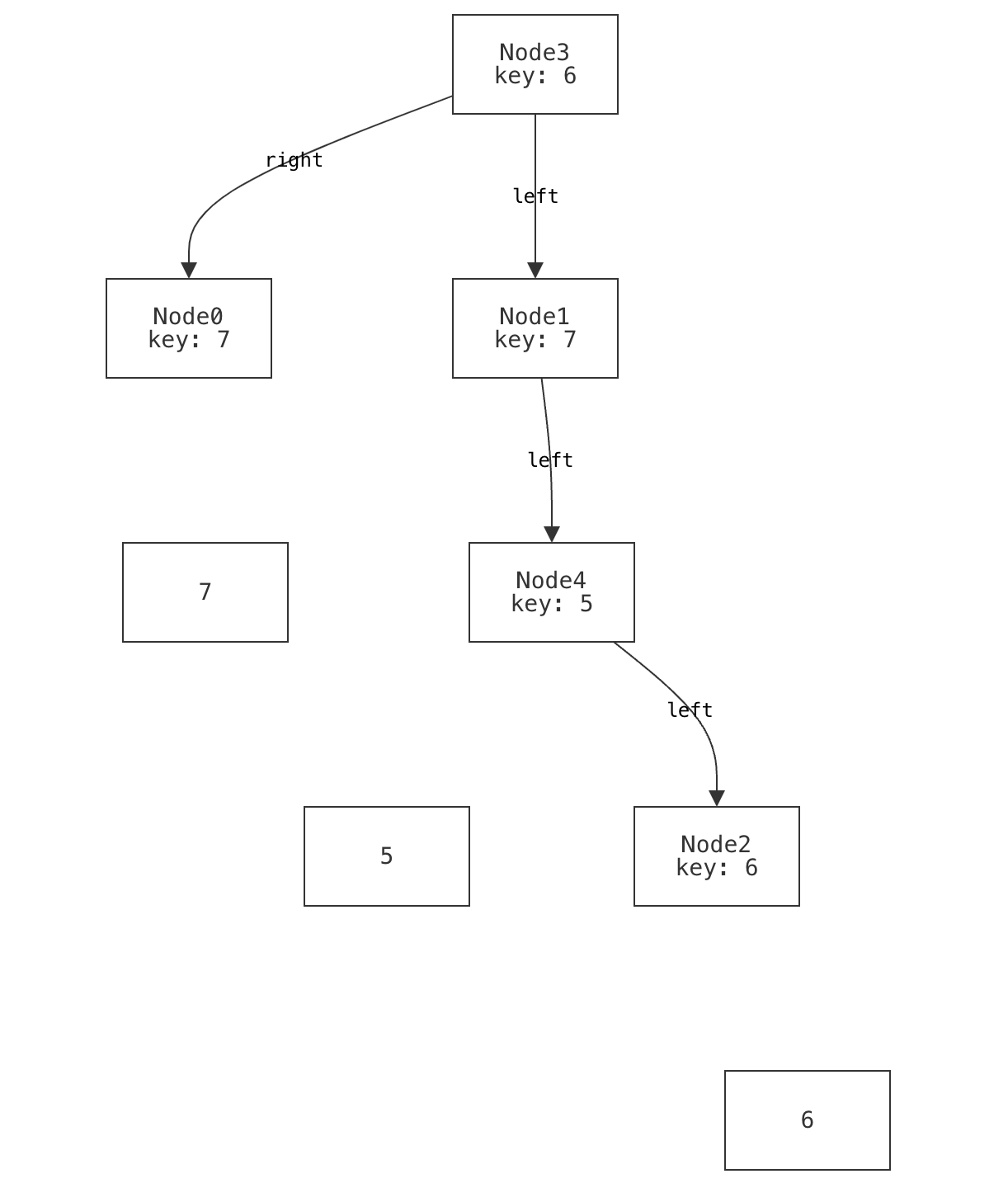}
    \caption{\ASV{}: By laying out \texttt{left} children to 
    the right and \texttt{right} children to the left, this visualization lends itself to spatial Stroop effects.}
    \label{f:sterling-default-bt}
  \end{subfigure}
  \hfill
  \begin{subfigure}[b]{0.45\textwidth}
    \centering
    \includegraphics[width=\textwidth]{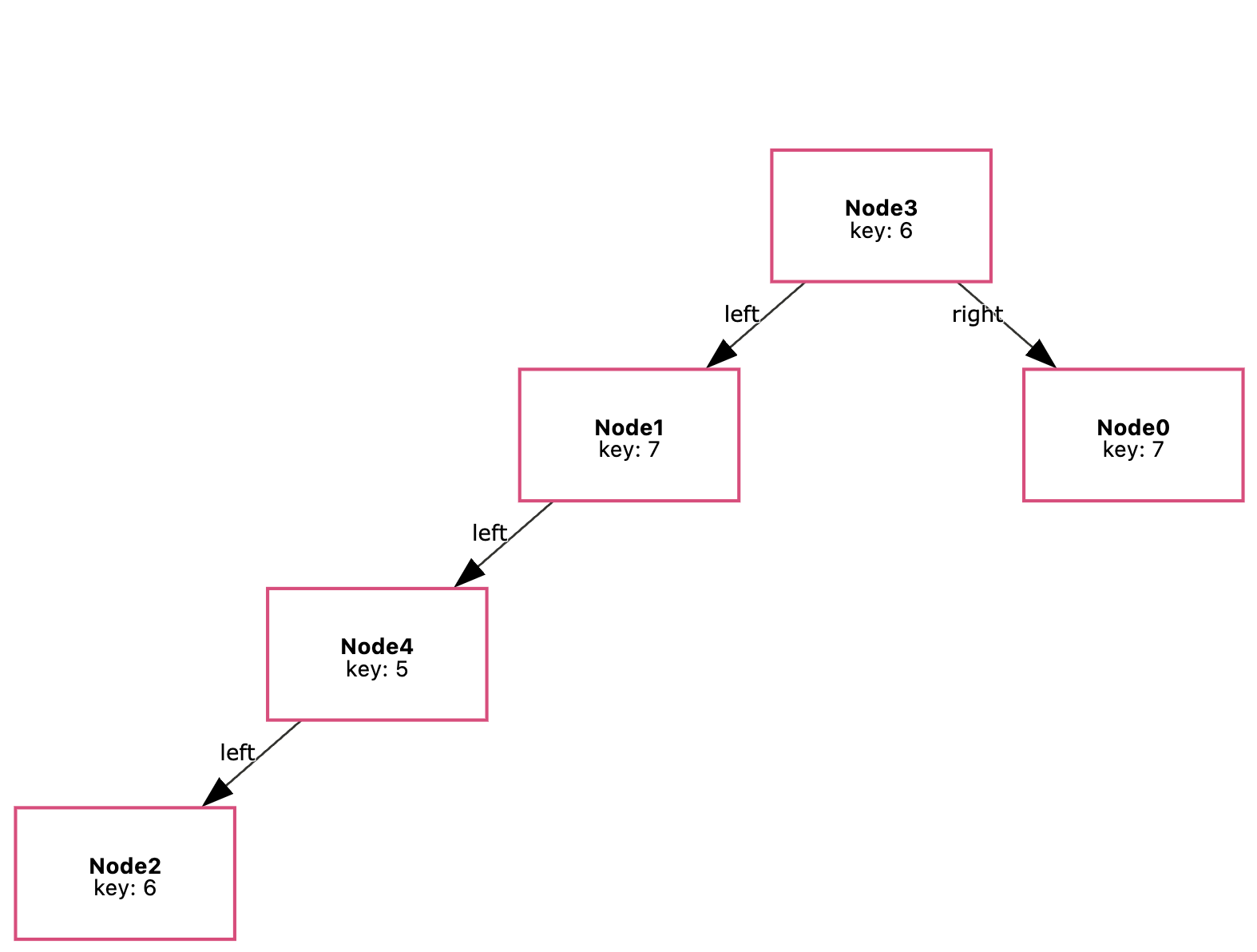}
    \caption{\cnd{} visualization using orientation constraints. All nodes along the \texttt{left} field are
    laid below and to the left of their parent node, while nodes along the \texttt{right}
    field are placed to the right and below their
    parent node.}
    \label{f:bst-orientation-example}
  \end{subfigure} \\
  \begin{subfigure}{0.8\textwidth}
    \centering
\begin{lstlisting}[language=cnd]
constraints:
  - orientation:
      field: left
      directions: [below, left]
  - orientation: 
      field: right
      directions: [below, right]
directives:
  - attribute: { field: key }
  - flag: hideDisconnected
\end{lstlisting}
    \caption{\cnd{} spec}
  \end{subfigure}

  \caption{\pill{bt} \ASV{} and \cnd{} visualizations of a binary tree.
   }
  \label{f:bst}
\end{figure}

Orientation constraints specify the placement of atoms in a diagram instance relative to one another.
These constraints can be applied between sig types (e.g., all atoms of type \texttt{A} must be to the left of those of type \texttt{B})
or along sig fields (e.g., atoms related by the \texttt{child} field must be placed downward).
\Cref{f:bst-orientation-example} uses orientation constraints to visualize
the binary tree instance in \cref{f:sterling-default-bt}, while avoiding the Stroop effect.

\begin{figure}[h]
  \centering

  \begin{subfigure}{0.45\textwidth}
    \centering
    \includegraphics[width=\textwidth]{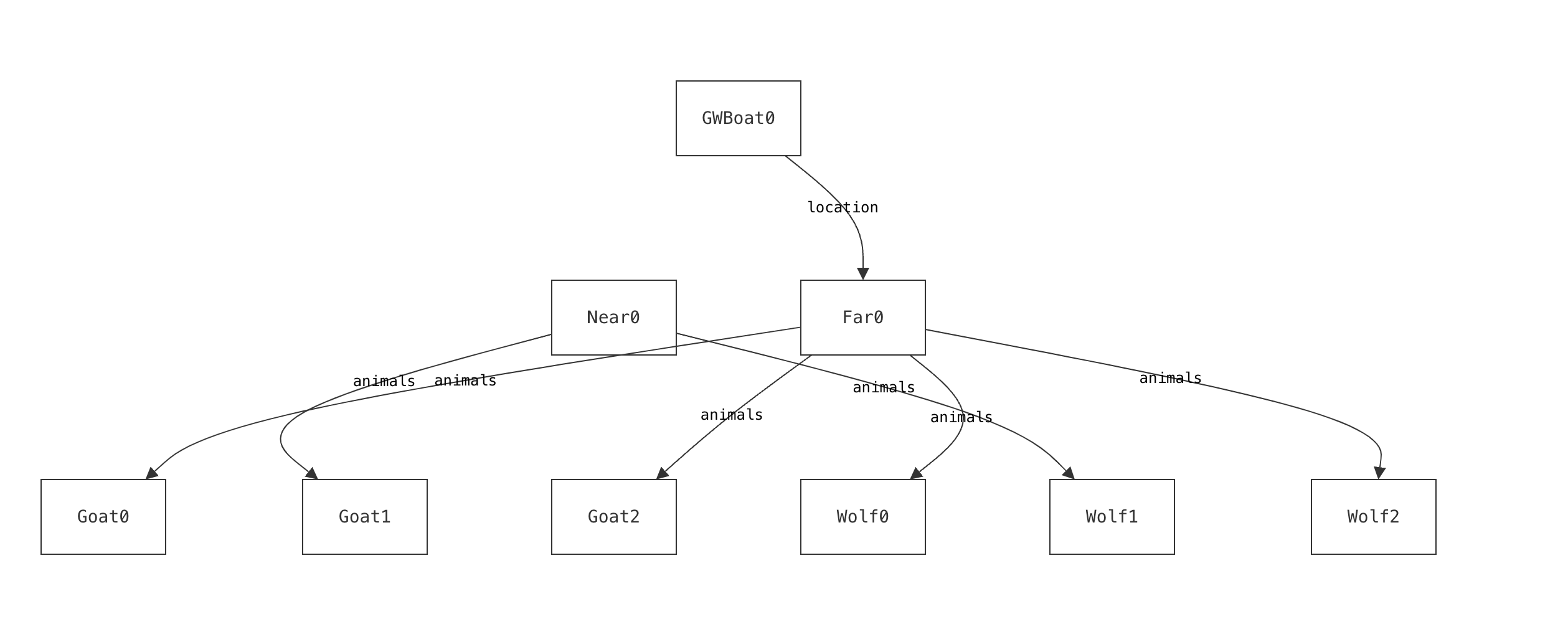}
    \caption{\ASV{}:
    This visualization does not spatially group animals by shore, 
    making it harder to determine if the spec is valid.}
    \label{f:sterling-default-gw}
  \end{subfigure}
  \hfill
  \begin{subfigure}{0.45\textwidth}
    \centering
    \includegraphics[width=\textwidth]{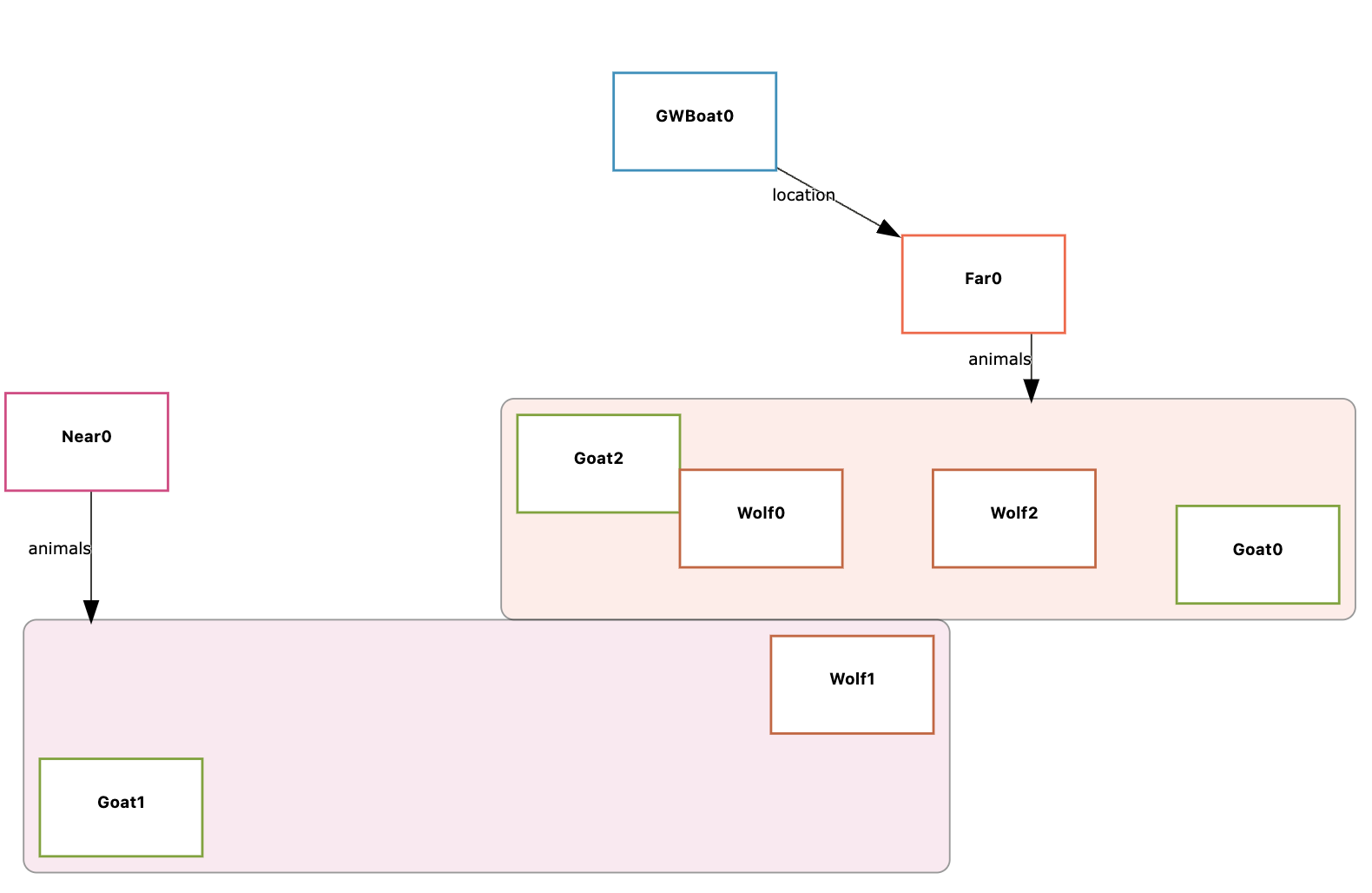}
    \label{f:gw-grouping-example}
    \caption{\cnd{} visualization using a grouping constraint on the \texttt{animals} field.
    All animals are grouped together by the shore they are on.}
  \end{subfigure}\\
  \begin{subfigure}{0.45\textwidth}
    \centering
\begin{lstlisting}[language=cnd]
constraints:
    - group: 
        field: animals
        target: domain

\end{lstlisting}
    \caption{\cnd{} spec}
  \end{subfigure}
  \caption{\pill{rc} \ASV{} and \cnd{} representations of a river-crossing puzzle instance, where
    a valid spec must ensure that wolves never outnumber goats on either shore.}
  \label{f:gw}
\end{figure}

\paragraph{Grouping Constraints}
Grouping constraints specify that atoms related by a sig field should be grouped together.
Groups can be created based on either the range (default) or domain of the sig field.
A grouping constraint replaces the individual edges between the group source and target atoms
with a single edge, and places grouped atoms within a bounding box. 
Groups cannot intersect unless one is entirely subsumed by another. 
\Cref{f:gw-grouping-example} uses a grouping constraint to simplify
the river-crossing instance visualization in \cref{f:sterling-default-gw}.\footnote{
The bounding boxes in \cref{f:gw-grouping-example} are larger than they need to be.
This is a limitation of the WebCola library~\cite{webcola} used by \cnd{}.}

\noindent
This picture reduces the number of edges; it also shows which entities
are on the same shore, thereby avoiding confusion because near and far
shore entities are interleaved in \cref{f:sterling-default-gw}. In
general, grouping constraints allow users to focus on higher-level
structures and relationships without being overwhelmed by individual
connections.

\paragraph{Directives}

Directives allow users to control the visual aspects of how atoms of a particular sig type are rendered.

Pictorial directives specify that atoms of a particular sig type must be represented by a specific icon. 
These icons can be customized to represent domain-specific concepts, making the diagram more intuitive and easier to interpret.
For instance, in a file system spec, folders might be represented by folder icons, while files could be shown as document icons.
This visual distinction helps users quickly grasp the spec's relationships and structure.

\cnd{} also supports a subset of the theming directives made available by Alloy Visualizer and Sterling.
Attribute directives display a field relation as a label on atoms of a particular sig type.
Sig color directives associate specific colors with atoms of a certain sig type.
Projection directives allow diagrams to focus on the atoms of a particular sig and their
directly connected relations, hiding all other elements.
Users can also hide disconnected atoms, 
depending on the provenance of their sig type, by using the \texttt{hideDisconnected} 
and \texttt{hideDisconnectedBuiltIns} visibility flags.

\subsection{When Constraints Cannot be Satisfied}
\label{s:unsat}

All \cnd{} constraints are hard constraints, and thus \emph{must} be satisfied for a diagram to be produced.
Constraints may not be satisfied for one of two reasons: 
\begin{enumerate}
  \item \cnd{} constraints could be internally inconsistent. 
  This represents a bug in the \cnd{} specification, and can be identified statically.
  In this case, \cnd{} produces an error message in terms of the constraints that could 
  not be satisfied. Checking for inconsistencies is a relatively simple process:
  \cnd{} just needs to ensure that the same field is not laid out in conflicting directions.
  For instance, a constraint that requires the same field to be
  laid out both leftwards and rightwards
\begin{lstlisting}[language=cnd]
- orientation:
    field: next
    directions: [right, left]
\end{lstlisting}
  would result in the error
  \begin{verbatim}
  Inconsistent orientation constraint: 
  Field next cannot be laid out with directions: right, left.
  \end{verbatim}

  \item \cnd{} constraints could be unsatisfiable when laying out the specific instance being visualized.
        This is akin to a dynamic error, as it depends on the details of that instance.
        \cnd{} identifies these inconsistencies by treating layout as a linear optimization problem.
        Constraint primitives are incrementally added to the Cassowary linear constraint solver~\cite{badros2001cassowary}
        to ensure that a satisfying layout exists. \cnd{} stops at the first indication that constraints cannot be satisfied,
         and provides an error message in terms of the currently added constraints and instance atoms (\cref{f:silent-failure-bt}).

\end{enumerate}
In both these cases, \cnd{} does \emph{not} produce a diagram. Instead, it provides an error message explaining
that the constraints could not be met. If all constraints can be satisfied,
\cnd{} relies on the WebCola library~\cite{webcola} to produce a diagram.

\subsection{The Lightweightness of \cnd{}}

Having described how \cnd{} is derived from the alignment of cognitive
design principles with the features we find in bottom-up exploration,
we briefly call attention to its lightweight nature.

Critically, \cnd{} \emph{refines} an \ASV{} output. Thus, the user can apply it
incrementally: the empty \cnd{} program still produces output, indeed,
the exact same output as Sterling. Users can thus decide which aspects
of the output most need refinement (just as they already do with
\ASV{} directives) and apply these rules incrementally. Furthermore,
it is easy to turn a rule on or off, seeing whether or not it has the
desired effect.

\cnd{} programs also tend to be very brief. There is only a small
number of primitives (constraints and directives) in the language
(\cref{s:primitives}). Because these primitives apply to sigs and fields, there is
an upper-bound on how many we can write, and this is independent of
the size of the \emph{operational} part of the spec.

\cnd{} output is also ``lightweight'' in another sense: the
language offers few bells-and-whistles, and the output is not
necessarily pretty. We view the path from a \cnd{} program to a
full-fledged custom visualization as akin to that from an Alloy or
Forge spec to one written and rigorously proved in a proof
assistant. The goal is not to produce diagrams that are attractive,
but rather ones that are functionally useful and do not confuse,
mislead, or otherwise abuse cognitive principles.

\subsection{Where do \cnd{} Programs Belong?}

A design question is where to write a \cnd{} program. Is it part of
the spec, or does it live separately from the spec? This partially
depends on whether it even makes sense to have more than one \cnd{}
program for a spec.

We argue that it is not only possible but also sensible to have
multiple views on instances of the same spec. As an example,
suppose a user is modeling a self-stabilizing protocol for a distributed system.
Depending on the goal, they might want to use
a ring to show network topology, a DAG to examine hierarchies, or groups to explain permissions.
No one view is more privileged than the other. They might even switch
between these as they explore and debug different aspects.

In our current implementation, the \cnd{} program resides in
the visualizer. This is for three more reasons beyond the one given
above. First, it is easy to explore
commands incrementally even for a given instance. Second, the same
visualizer works with both Alloy and Forge (and any other tools that
use the same protocol), providing portability of these ideas. Finally,
it saves us the effort of modifying the implementations of Alloy and Forge.

That said, \cnd{} programs are not completely separate from the spec;
they are really a \emph{spatial refinement} of the spec. Therefore, it
would be meaningful to extend the spec languages to include (and
name) each of these views. The ``static'' checking that \cnd{} provides
(\cref{s:unsat}) is then arguably part of the act of checking the spec
itself. The user interface could then provide a menu of these
pre-defined views, which were presumably chosen by the domain expert,
while enabling the user of the spec to write their own views as well
(as they currently can).

\subsection{Visual Details}

Whenever possible, \cnd{} diagrams also encode a set of visual principles designed to reduce cognitive load and enhance user comprehension.
Graph edges are laid out in a way that minimizes crossings and with as few bends as possible~\cite{purchase1997aesthetic}.
To reduce edge ambiguity, \cnd{} also ensures that arrowheads are not incident on the same point of a node.
Atoms of each sig are assigned a unique color by default, making salient the type of each node in the diagram.

\section{Bad-Instances: When Custom Visualizations Fail}
\label{s:when-viz-fail}

Model-finding tools like Alloy and Forge have an important modality
not found in other formal methods tools like verifiers (\cref{s:intro}).
A verifier is inert in the absence of a property: it cannot operate on a spec
alone. Furthermore, determining the properties a system should enjoy
is difficult and subtle.

In contrast, the lightweight philosophy employs model-finders because
they can consume \emph{just} a spec and show concrete instances of it 
(and some authors~\cite{montaghami2017bordeaux, posvalueofnegativeinfo} 
have shown there is value to modifying them to even show 
\emph{non}-instances). These instances
help bootstrap the formalization process, helping authors both improve
and correct their spec and realize what aspects they would like
to capture as properties. Thus, the modeling process involves a series
of incremental efforts that in practice includes producing many
incorrect specs.

These incorrect specs lead to what we dub \emph{bad-instances}. It is
critical for a visualizer to faithfully render these instances, since
the user needs to see them and realize that the spec is incorrect.

This is not an issue when using the \ASVs{}: for all their weaknesses
described earlier in this paper, to their credit, instances and
bad-instances alike are displayed as directed graphs, with no
information lost or hidden (except as stated in an explicit user
directive). In contrast, a custom visualization needs to be sensitive
to bad-instances and not misrepresent them or, worse, accidentally
suppress the way in which they are bad (so the user may never discover
that the spec is flawed). As a participant from the survey in \cref{s:bottom-up} says,

\quoteparticipant{70}{...buggy visualization code could lead a student down the wrong path while debugging}

\noindent
Nelson et al.~\cite{nelson2024forge} discuss this in their study of
custom \SDT{} visualizations. They find that as users modify their
spec, domain-specific visualizations would behave unexpectedly,
leading to confusion and frustration for users.

\begin{figure}[t]
  \centering
  \begin{subfigure}[b]{0.3\textwidth}
    \centering
    \includegraphics[width=\textwidth]{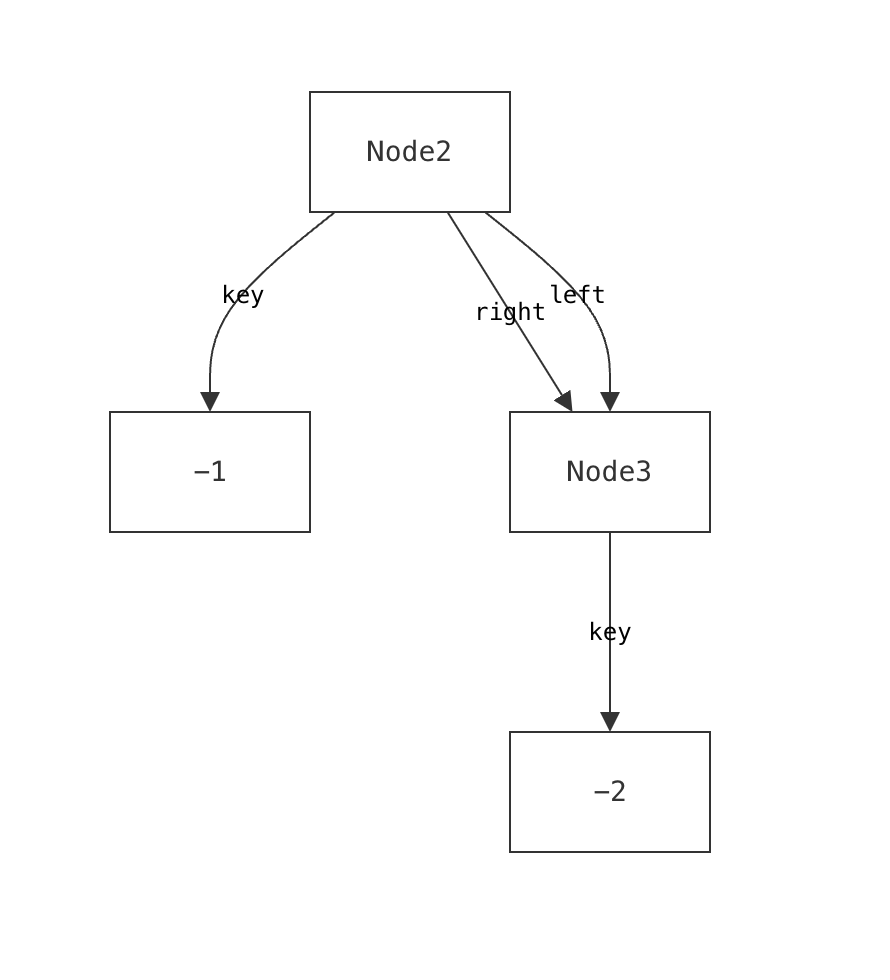}
    \caption{The \ASV{} visualization of the binary search ``tree'' bad-instance.}
  \end{subfigure}
  \hfill
  \begin{subfigure}[b]{0.3\textwidth}
    \centering
    \includegraphics[width=\textwidth]{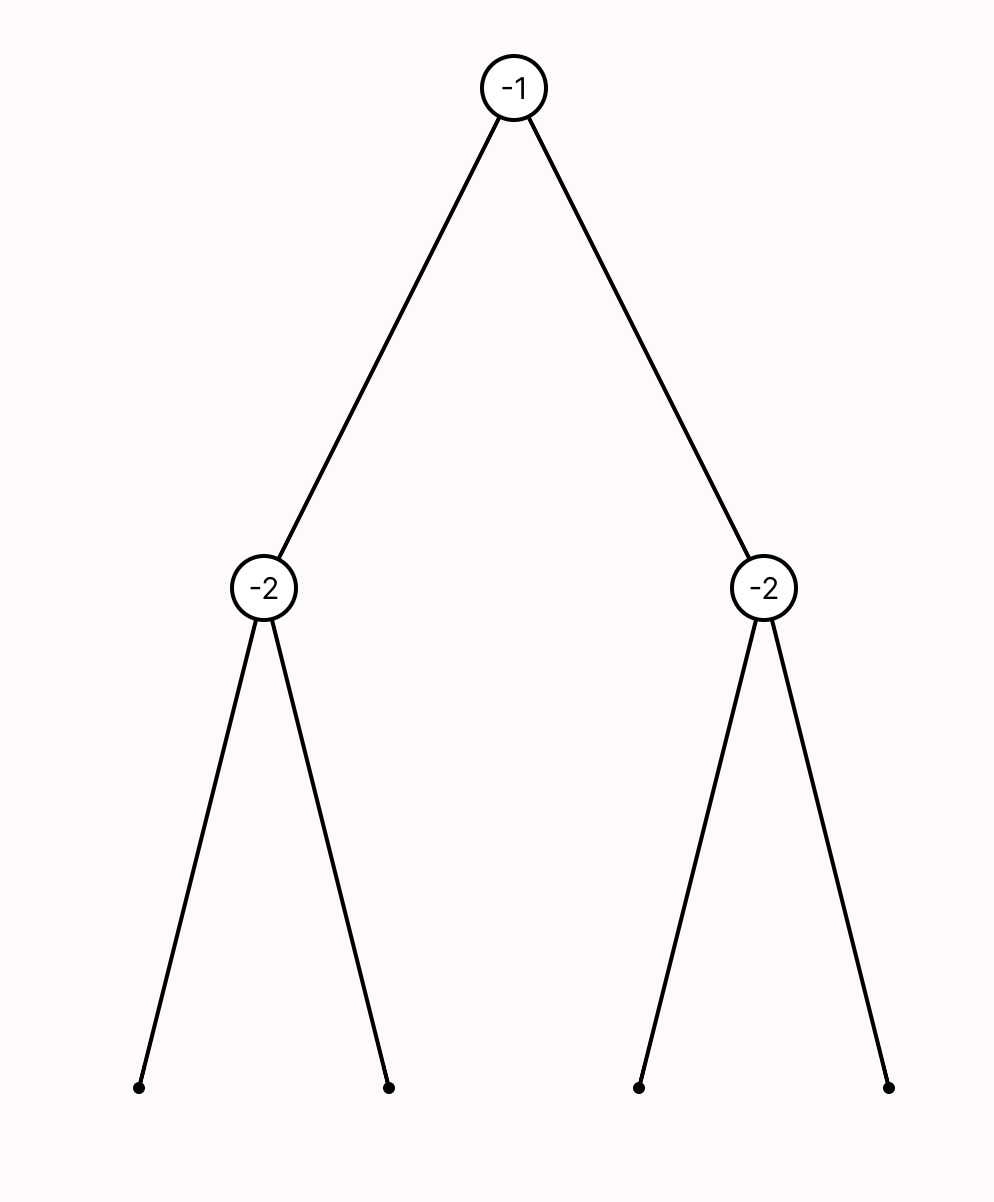}
    \caption{An expert-authored visualization of the same binary search ``tree'' bad-instance.}
  \end{subfigure}
  \hfill
  \begin{subfigure}[b]{0.3\textwidth}
    \centering
    \includegraphics[width=\textwidth]{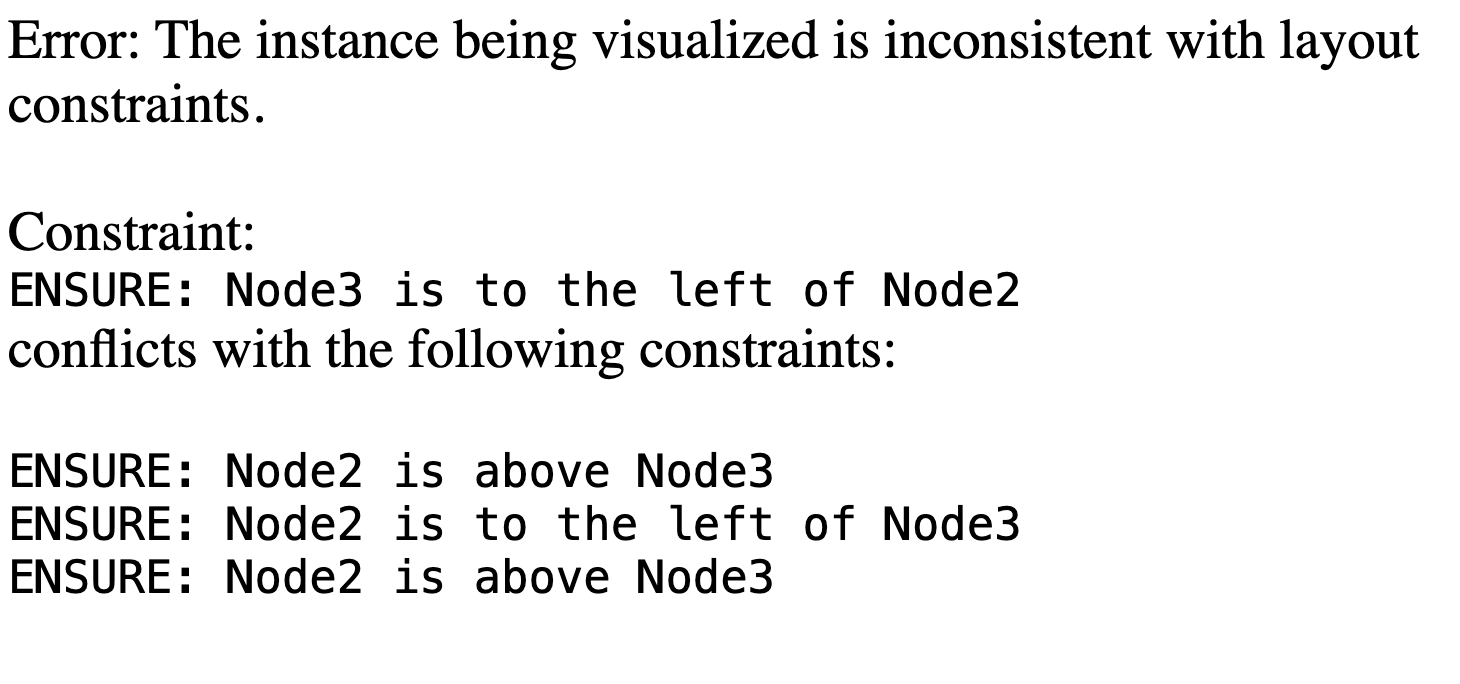}
    \caption{\pill{bt-dag} The \cnd{} specification in \Cref{f:bst-orientation-example} produces
     an error message instead of a diagram.}
  \end{subfigure}

  \caption{
    Visualizations of a binary search tree bad-instance that is a DAG.}
  \label{f:silent-failure-bt}
\end{figure}

\begin{figure}[t]
  \centering
  \begin{subfigure}[b]{0.3\textwidth}
    \centering
    \includegraphics[width=\textwidth]{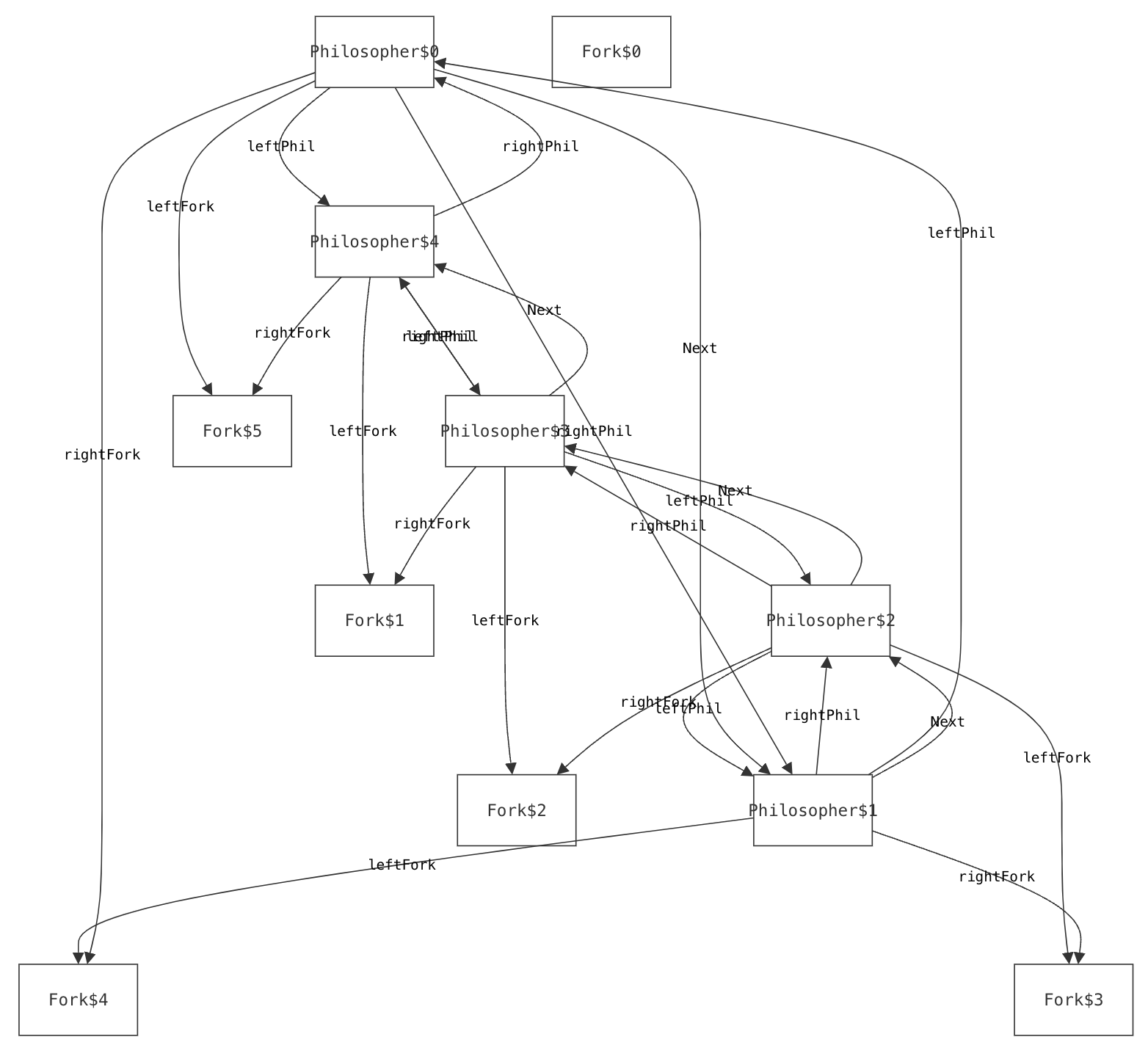}
    \caption{\ASV{} of the dining philosophers bad-instance. The extra fork is visualized, but is not particularly salient.}
  \end{subfigure}
  \hfill
  \begin{subfigure}[b]{0.3\textwidth}
    \centering
    \includegraphics[width=\textwidth]{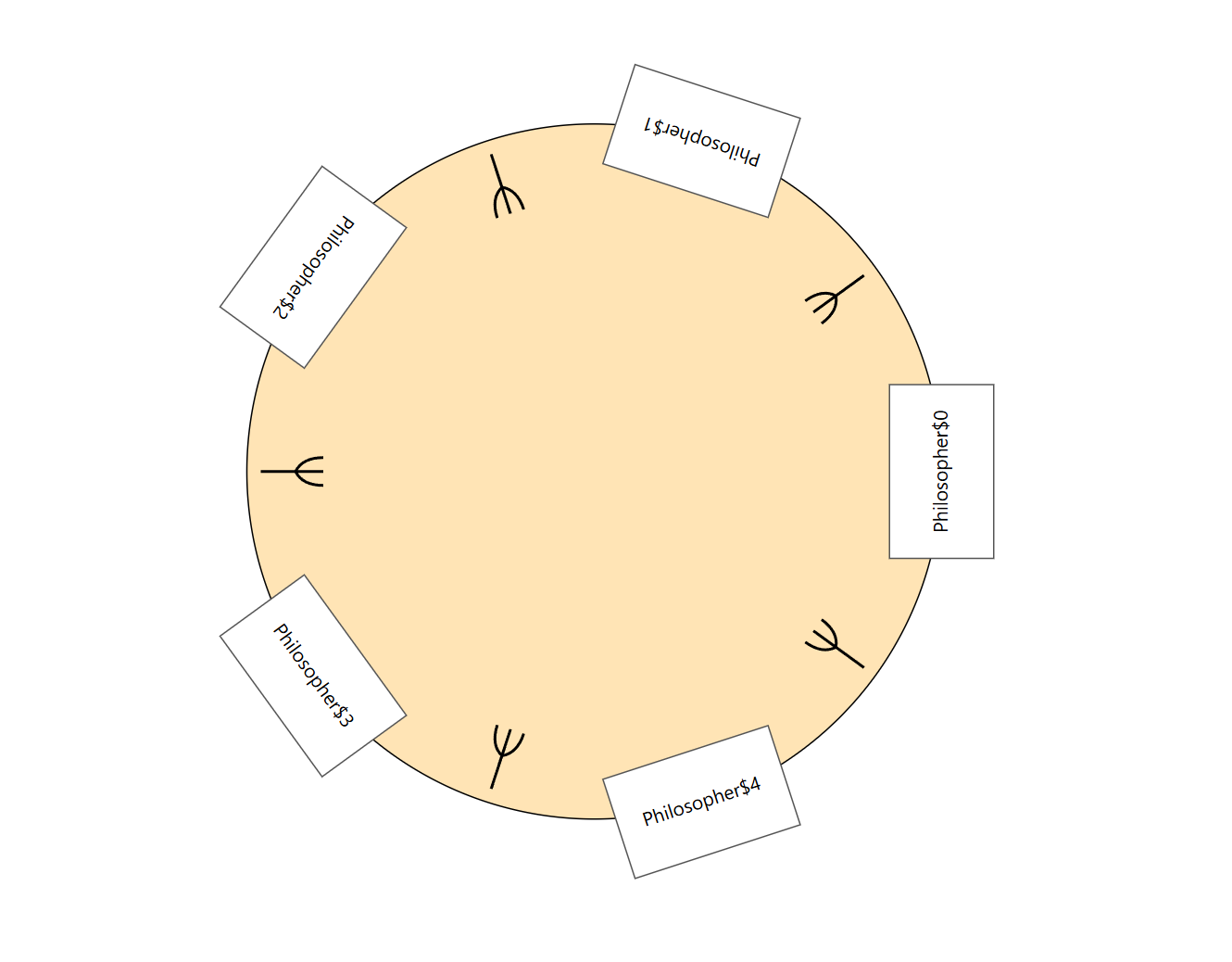}
    \caption{Custom visualization of the invalid dining philosophers instance. This 
    visualization fails silently, hiding the extra fork.}
  \end{subfigure}
  \hfill
  \begin{subfigure}[b]{0.3\textwidth}
    \centering
    \includegraphics[width=\textwidth]{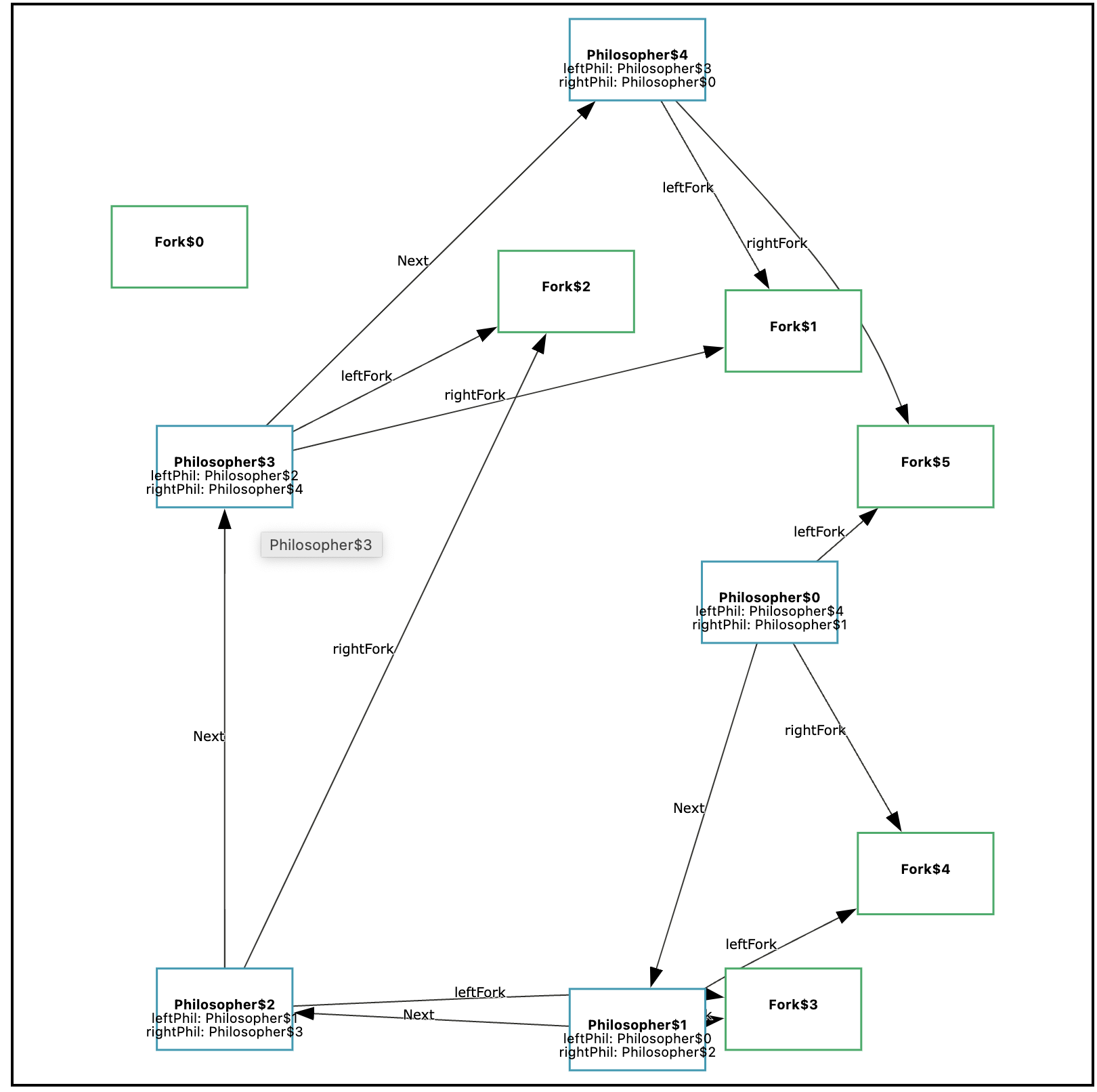}
    \caption{\pill{phil-inv} A \cnd{} diagram for the invalid dining philosophers instance.
    The diagram shows the circular seating layout while also showing all 6 forks.}
  \end{subfigure}

  \caption{
   Visualizations of a dining-philosophers bad-instance with 6 forks
   and 5 philosophers.
   }

  \label{f:silent-failure-phil}
\end{figure}

This is not a problem limited to students. Consider these two examples
from experts:
\begin{itemize}

\item A common underconstraint when modeling a binary search tree is
  to forget to enforce that a node's left and right children are
  distinct.  \Cref{f:silent-failure-bt} shows the result of applying
  an expert-authored visualization to a bad instance: a DAG.  The
  custom visualizer is not robust to this possibility and thus masks
  an important failure. The \ASV{} reveals the underlying structure,
  while \cnd{} does even better, displaying an error message
    about the inconsistency between the bad-instance and diagramming
    constraints.

\item An expert-authored visualization of the dining philosophers problem
from the Sterling website~\cite{dyer_sterling_js_demo} also fails silently when presented with bad-instances.
A common underconstraint when modeling the dining philosophers problem is not
ensuring that there are more forks than philosophers. 
As shown in \cref{f:silent-failure-phil}, the expert-written visualization
presents the bad-instance as an instance, hiding the presence of the 6th fork.
The \ASV{} hides no information, but the lack of semantically meaningful layout 
means that the sixth fork is not particularly salient.
The \cnd{} diagram, however, arranges the domain in a meaningful way, which
makes salient the presence of the extra fork.

\end{itemize}

Of course, an attentive visualization author can protect against some
of these issues with checks, preconditions, and assertions. Besides
being onerous and easy to overlook, these suffer from a more
fundamental problem: they are limited by the author's assumptions of
what can go wrong. That is, they can only handle \emph{known}
unknowns: faulty specs that the author thinks of. However, these would
probably be turned into properties in the first place. Problems that the
author does not think about (\emph{unknown} unknowns)---the very kind
for which lightweight formal methods are ideal---are, by
definition, impossible to account for, and may well result in
misleading or deceptive visualizations. This problem is especially
salient in educational contexts, where it is well-documented that
experts suffer from blind spots~\cite{nathan2001expert, nathan2003expert} about what mistakes students
might make.

\cnd{} again strikes a happy medium between generic and custom
visualization. Because it only refines the generic output, it does not hide
information. Because the refinements encode domain information, bad
instances will either fail noisily (\cref{f:silent-failure-bt})
or produce a structured diagram that does not mask the incorrectness
(\cref{f:silent-failure-phil}).
We study this phenomenon further in \cref{s:prolific-non-model}.

\section{Evaluation}
\label{s:eval}

We now evaluate \cnd{}. Our evaluation has two parts. In
\cref{s:examples}, we examine its ability to generate visualizations
for new domains. In \cref{s:studies}, we provide empirical data on the
effectiveness of the generated diagrams.

\subsection{Examples}
\label{s:examples}

It is unsurprising that \cnd{} can capture well the demands of the
examples in \cref{s:bottom-up}, given that these were effectively the
``training set''. We instead need a distinct ``test set''. For that,
we use examples from the Alloy Models repository, a publicly available
collection of specifications maintained by the Alloy
community.\footnote{https://github.com/AlloyTools/models} It is
unclear how to \emph{comprehensively} study these; instead, we chose a
sample of specs for domains we understood well (and hence for which could
meaningfully write \cnd{} programs) and assess their effect.

\subsubsection{Filesystem}

\begin{figure}[H]
  \centering
  \begin{subfigure}[b]{0.45\textwidth}
    \centering
    \includegraphics[width=\textwidth]{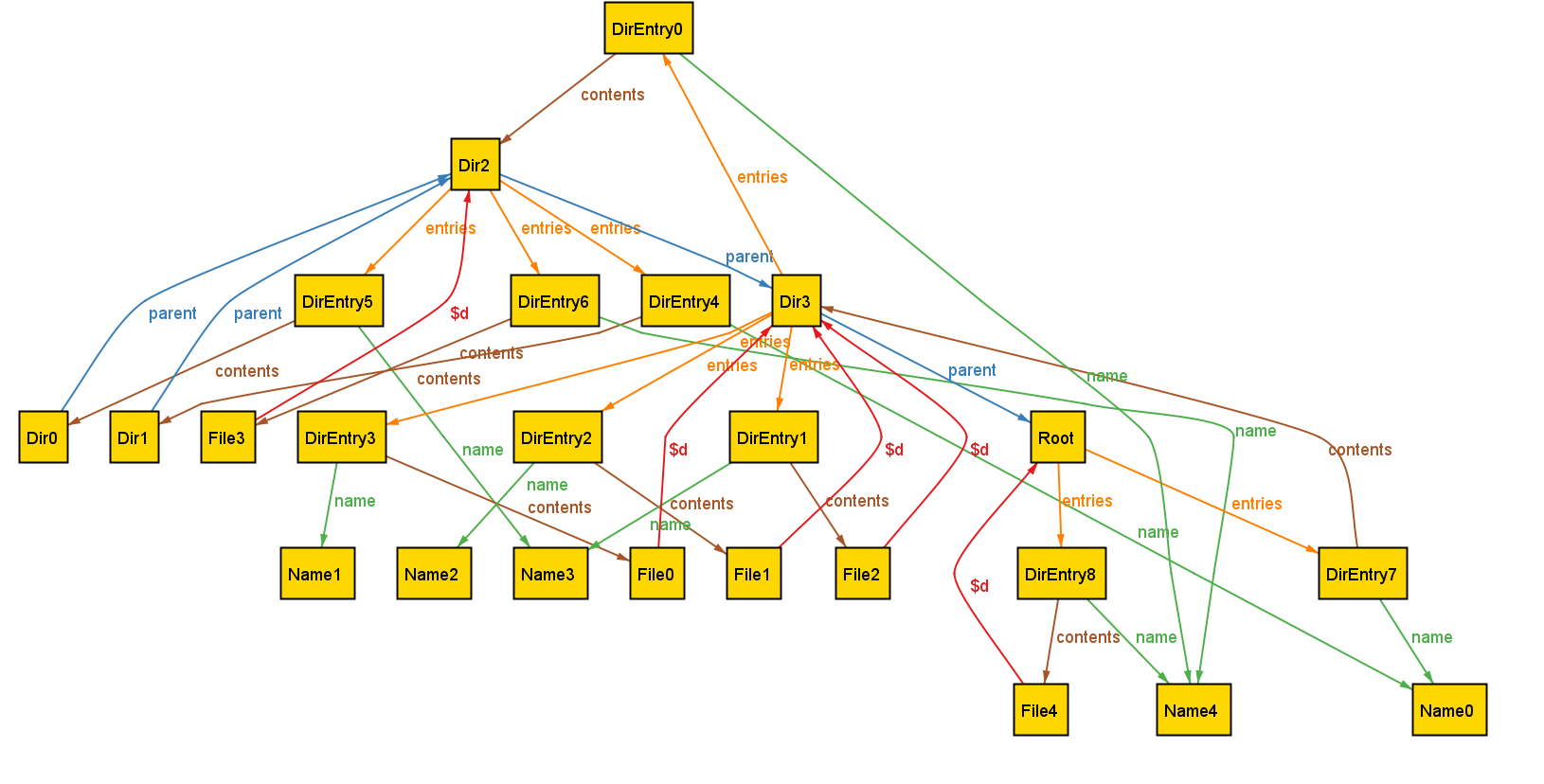}
    \caption{\ASV{}}
    \label{f:filesystem-asv}
  \end{subfigure}
  \hfill
  \begin{subfigure}[b]{0.45\textwidth}
    \centering
    \includegraphics[width=\textwidth]{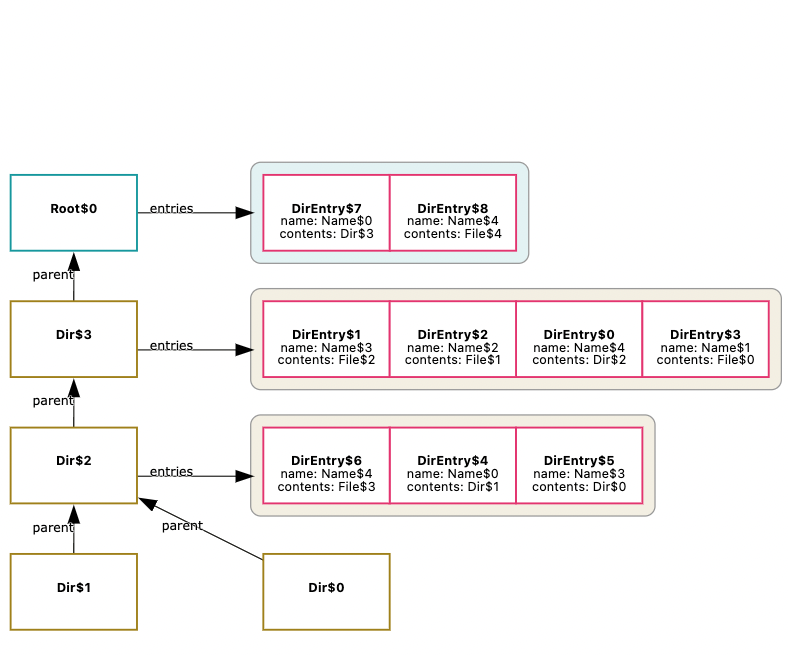}
    \caption{\pill{filesystem} \cnd{} diagram.}
  \end{subfigure}
  \\
  \begin{subfigure}[b]{0.5\textwidth}
    \centering
\begin{lstlisting}[language=cnd]
constraints:
  - orientation:
      field: parent
      directions: [above]
  - group:
      field: entries
      target: domain
  - orientation:
      field: entries
      directions: [directlyRight]
directives:
  - attribute:
      field: name
  - attribute:
      field: contents
  - flag: hideDisconnected
\end{lstlisting}
    \caption{\cnd{} spec}
  \end{subfigure}

  \caption{An instance of the Filesystem specification visualized by the \ASV{} and using \cnd{}.}  
  \label{f:filesystem}
\end{figure}

\Cref{f:filesystem} demonstrates how \cnd{} can be used to visualize the structure of 
a generic filesystem. This diagram uses
an orientation constraint to lay out directories hierarchically, and a combination of grouping 
and orientation constraints to lay out entries in the same directory.
Entry contents and names are shown as attributes.

\subsubsection{Ring Orientation}

\begin{figure}[H]
  \centering
  \begin{subfigure}[b]{0.4\textwidth}
    \centering
    \includegraphics[width=\textwidth]{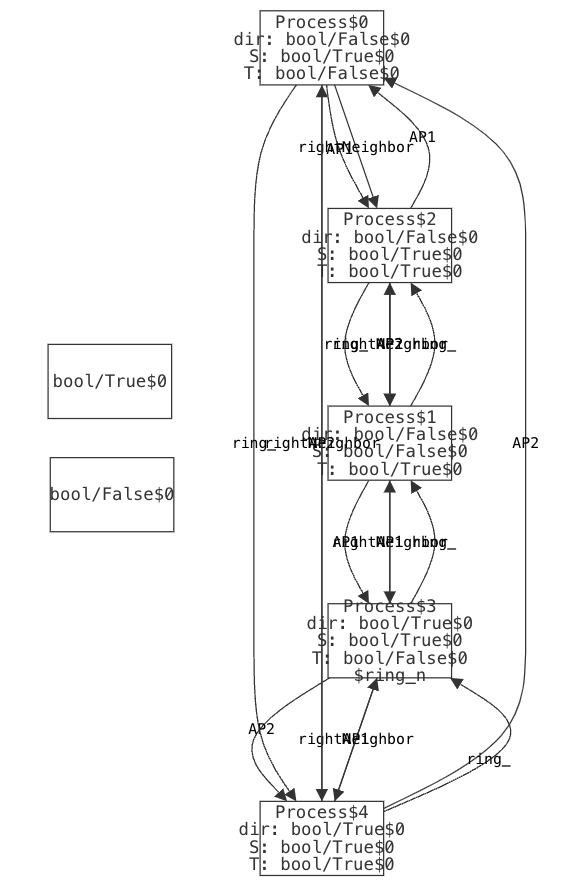}
    \caption{\ASV{}}
  \end{subfigure}
  \hfill
  \begin{subfigure}[b]{0.5\textwidth}
    \centering
    \includegraphics[width=\textwidth]{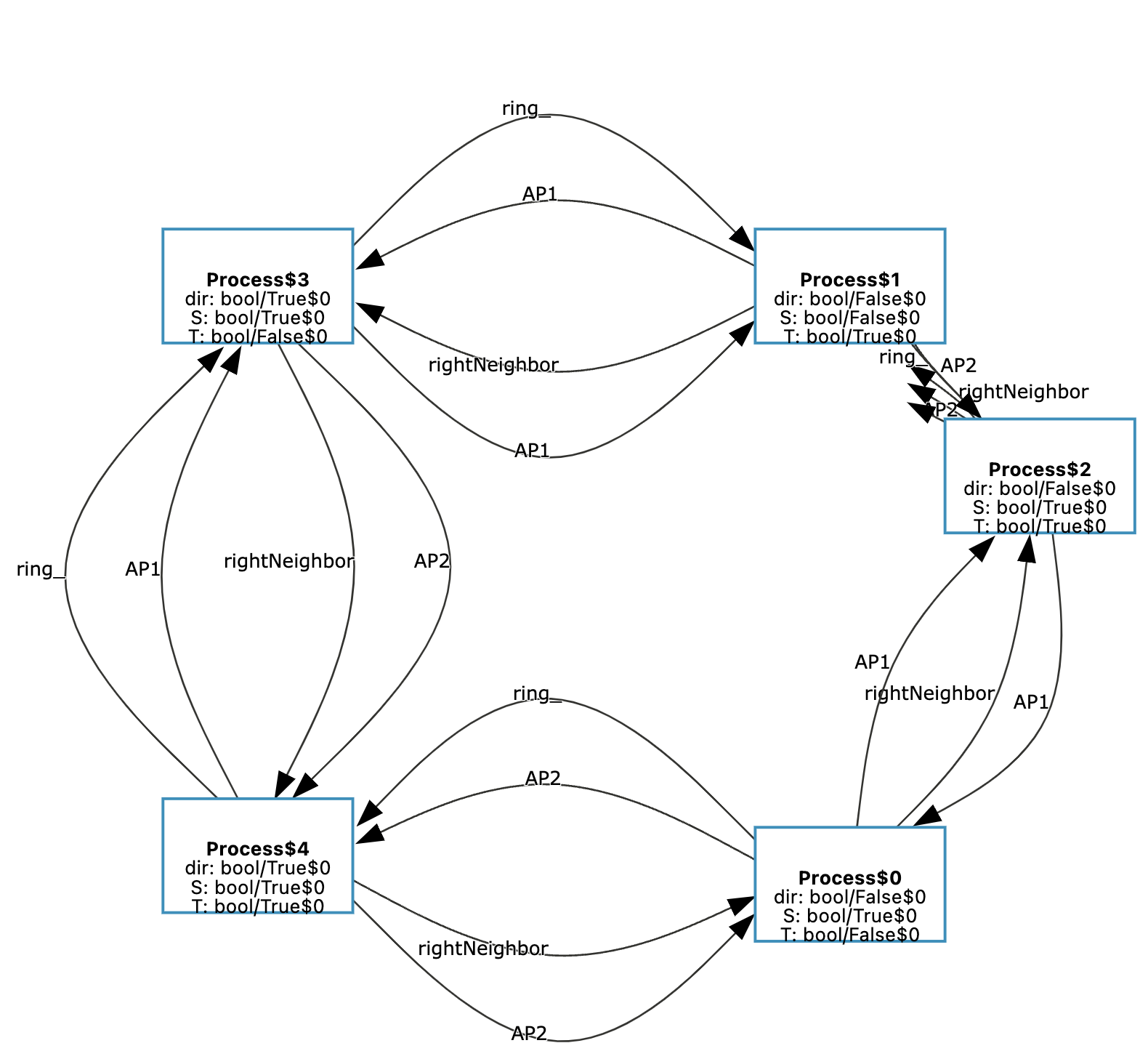}
    \caption{\pill{ring-orientation} \cnd{} diagram.}
  \end{subfigure}
  \\
  \begin{subfigure}[b]{0.5\textwidth}
    \centering
\begin{lstlisting}[language=cnd]
constraints:
  - cyclic:
      field: rightNeighbor
      direction: counterclockwise
  - cyclic:
      field: ring
      direction: clockwise
directives:
  - flag: hideDisconnected
  - attribute:
      field: dir
  - attribute:
      field: S
  - attribute:
      field: T
  - projection:
      sig: this/Tick
\end{lstlisting}
    \caption{\cnd{} spec}
  \end{subfigure}

  \caption{An instance of the Ring Orientation specification visualized by the \ASV{} and using \cnd{}.}  
  \label{f:ring-example}
\end{figure}

\Cref{f:ring-example} demonstrates how \cnd{} can be used to visualize instances
of an Alloy specification for the self-stabilization problem in distributed systems~\cite{dijkstra1974self}
for a ring of processes. A cyclic constraint is used to convey the circular topology of 
processes, while boolean flags are shown as attributes.

\subsubsection{Chord}

\begin{figure}[H]
  \centering
  \begin{subfigure}[b]{0.45\textwidth}
    \centering
    \includegraphics[width=\textwidth]{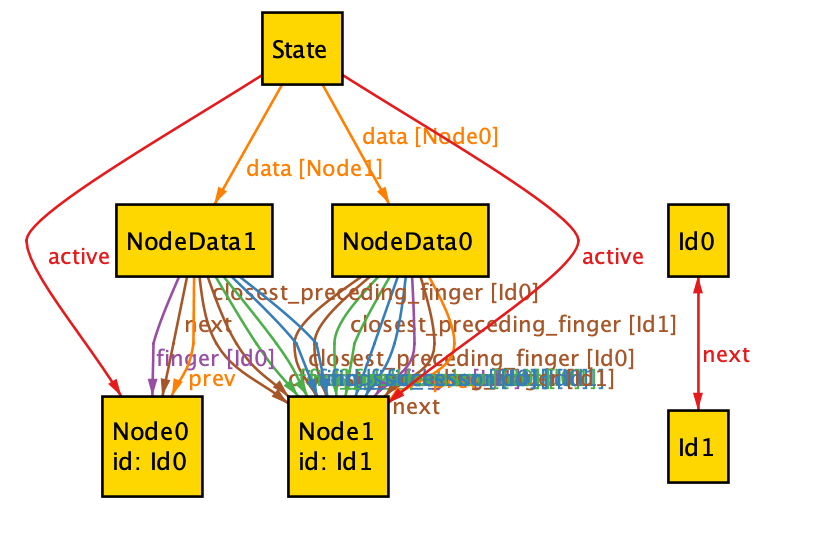}
    \caption{\ASV{}}
  \end{subfigure}
  \hfill
  \begin{subfigure}[b]{0.5\textwidth}
    \centering
    \includegraphics[width=\textwidth]{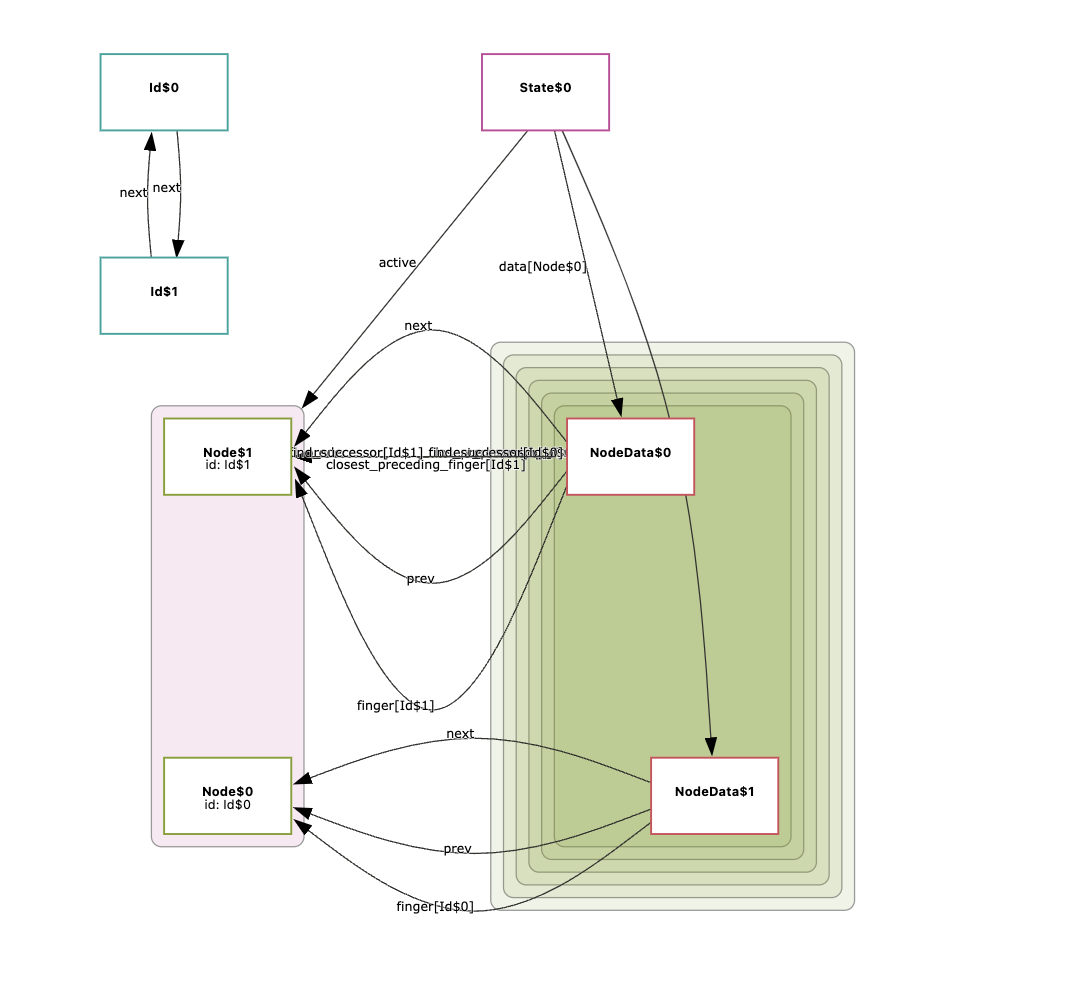}
    \caption{\pill{chord} \cnd{} diagram.}
  \end{subfigure}
  \\ 
  \begin{subfigure}[b]{0.5\textwidth}
    \centering
\begin{lstlisting}[language=cnd]
constraints:
  - group:
      field: active
      target: domain
  - group:
      field: closest_preceding_finger
  - group:
      field: find_predecessor
  - group:
      field: find_successor
  - orientation:
      field: prev
      directions:
        - directlyLeft
directives:
  - flag: hideDisconnected
  - attribute:
      field: id
\end{lstlisting}
    \caption{\cnd{} spec}
  \end{subfigure}

  \caption{An instance of the Chord specification visualized by the \ASV{} and using \cnd{}.}  
  \label{f:chord-example}
\end{figure}

\Cref{f:chord-example} demonstrates how \cnd{} can be used to visualize instances of the Chord~\cite{stoica2001chord} 
distributed hash table protocol. This diagram groups active nodes together, and 
lays out the previous node seen by a node to its direct left.
Finally, it uses grouping constraints to show the results of the
find\_successor and find\_predecessor operations.\footnote{
The multiple subsuming rectangles in this diagram are a limitation of the 
WebCola library used by \cnd{}, and are used to indicate subsumed groups.
One could imagine alternate implementations that collapse these groups into a single rectangle.}

\subsection{User Studies}
\label{s:studies}

We now examine the effectiveness of the diagrams generated by
\cnd{}. We use the \ASV{} as a baseline for comparison, for three
reasons. First, \ASVs{} are a fixed entity, whereas the set of possible
custom visualizations is limitless and our choices might be
biased. Second, as noted in \cref{s:when-viz-fail}, \ASVs{} does not
suppress problems in bad-instances. Finally, the \ASV{} represents
what a user gets by default, and indeed only a user aware of
custom visualization tools (e.g., Sterling), installed them, and has
written a custom visualization can do better.

In this section, we describe three studies that evaluate the
effectiveness of \cnd{} diagrams in conveying specification
relationships:
\begin{enumerate}
  \item Do \cnd{} specs help users understand the instances being explored (\cref{s:prolific-model})?
  \item Can pictorial
    directives further enhance spec understanding (\cref{s:prolific-icons} )?
  \item Do \cnd{} diagrams make salient key aspects of bad-instances (\cref{s:prolific-non-model})?
\end{enumerate}
For all three studies, we recruited participants on the Prolific
platform. Participants had prior experience in computer science, to
better reflect a typical formal methods user.  Each participant was
compensated USD 5 for about 15 minutes of time, and no individual took
part in more than one study. Our study was deemed to not be human
subjects research by our review board office; we nevertheless took
reasonable safeguards to protect the participants.
The questions and tasks for each study are available as supplementary material.

\subsubsection{Instance understanding}
\label{s:prolific-model}

In order to better understand how \cnd{} diagrams offer insight into
spec instances, we designed three specs, each of which lends itself
well to one kind of \cnd{} constraint. 

\begin{itemize}
  \item Cards: This scenario tested the efficacy of cyclic constraints by asking participants questions about the movement of player roles across rounds of a card game.
  \item Subway: This scenario tested the efficacy of orientation constraints by asking participants questions about the relative geographical positions of stations in a subway system.
  \item Fruit: This scenario tested the efficacy of grouping constraints by asking participants questions about the relative distribution of fruit across baskets.
\end{itemize}

Users were shown one instance
of the spec. For each spec, we developed 2--3 questions that would
exercise the user's understanding of the instance they were shown. The
specs avoided details that would enable a user to answer the questions
based on prior knowledge rather than the instance itself.

\begin{figure}[t]
  \centering
  \begin{subfigure}[b]{0.45\textwidth}
    \includegraphics[width=\textwidth]{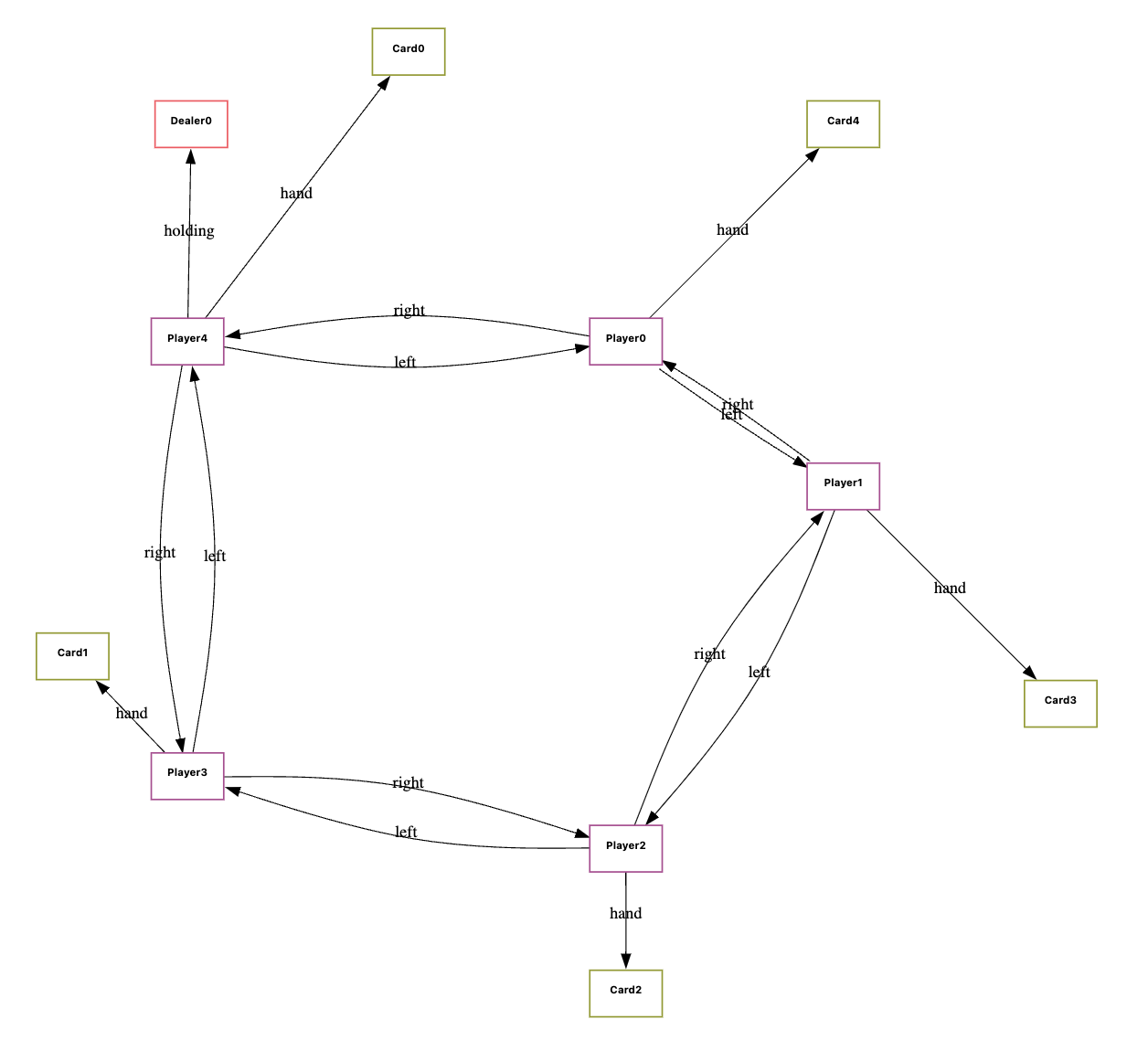}
    \caption{The \pill{cards} visualization.}
    \label{f:cards}
  \end{subfigure}
  \hfill
  \begin{subfigure}[b]{0.45\textwidth}
    \centering
    \includegraphics[width=\textwidth]{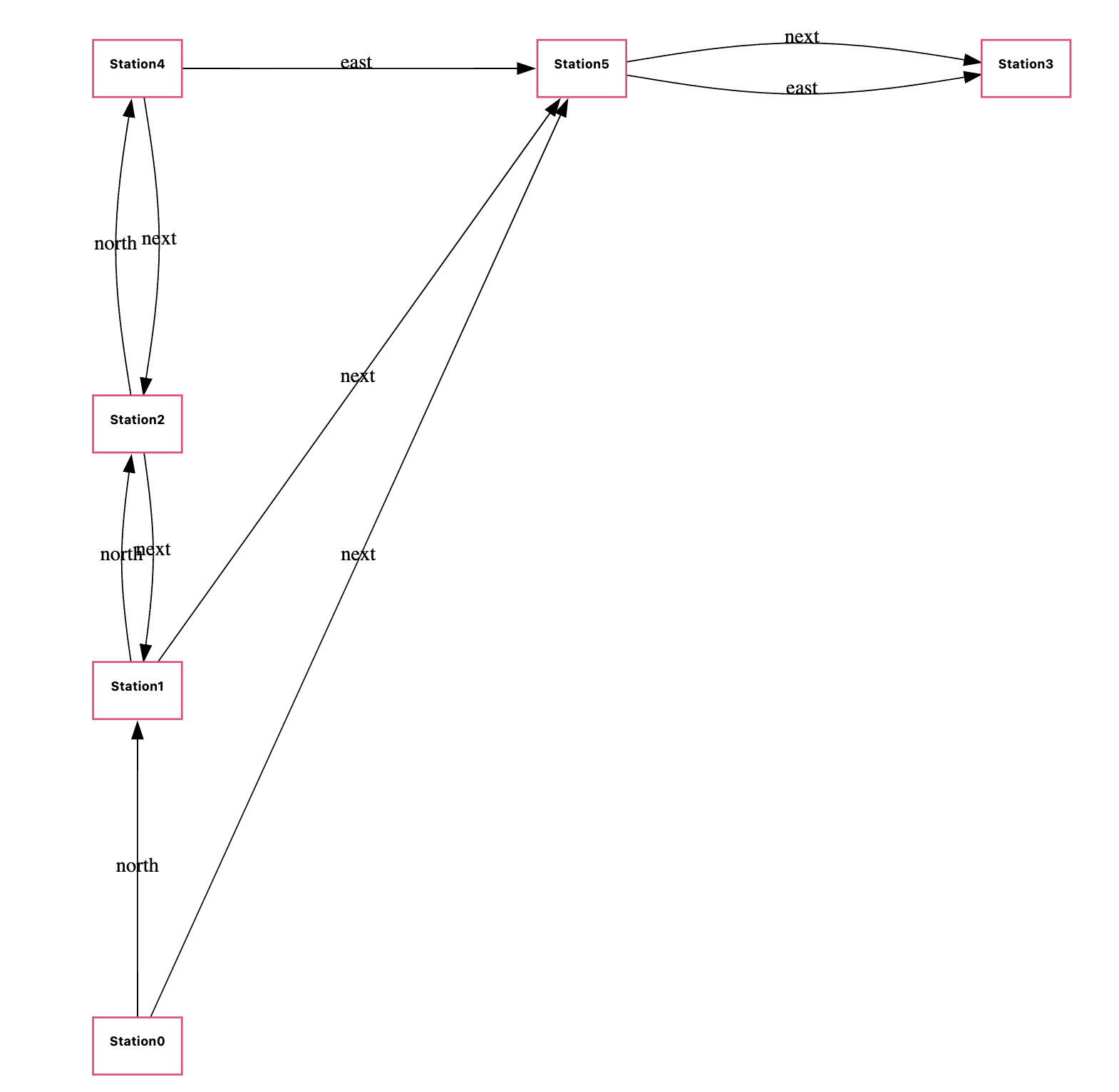}
    \caption{The \pill{subway} visualization.}
    \label{f:subway}
  \end{subfigure}
  \caption{
    \cnd{} visualizations of the scenarios described in \cref{t:study1-correctness}.}
  \label{f:p1-figs}
\end{figure}

Each participant (n = 38) was randomly assigned 2 scenarios, one visualized
using the \ASV{} and the other using \cnd{} with \emph{only} the associated constraint.

Participants shown \cnd{} diagrams were significantly more likely to
get questions correct than those shown Sterling visualizations 
($62.75\%$ vs $48.04\%$ correct, $Z = 2.30$, $p < 0.05$), 
with a small-to-medium effect size ($d = 0.26$).

Participants took approximately the same amount of time
to complete the tasks 
(mean time on task: $105.3$s for \cnd{}, $100.46$s for \ASV{}, $t=-0.81$, $p = 0.22$).

\begin{table}[t]
  \caption{Scenarios used to study the effectiveness of \cnd{} constraints on spec understanding.}
  \label{t:study1-correctness}
  \centering
  \begin{tabularx}{\textwidth}{l|l|rr|rr}

    Scenario & Constraint & \multicolumn{2}{c}{Correct Answer Percent} & \multicolumn{2}{c}{Mean Time on Task} \\
    \cline{3-6}
    &                         & \makecell{\ASV{}} & \makecell{\cnd{}} & \makecell{\ASV{}} & \makecell{\cnd{}} \\
    \hline
    \pill{cards} & Cyclic & 28\% & 36\%    & 161.96s & 128.81s \\ 

    \pill{subway} & Orientation & 31\% & 56\%   & 84.02s & 121.73s \\ 

    \pill{fruit} & Grouping & 69\% & 74\%      & 60.15s & 70.32s \\ 

  \end{tabularx}

\end{table}

\subsubsection{Pictorial directives}
\label{s:prolific-icons}

\begin{figure}[t]
  \centering
  \begin{subfigure}[b]{0.9\textwidth}
    \centering
    \includegraphics[width=\textwidth]{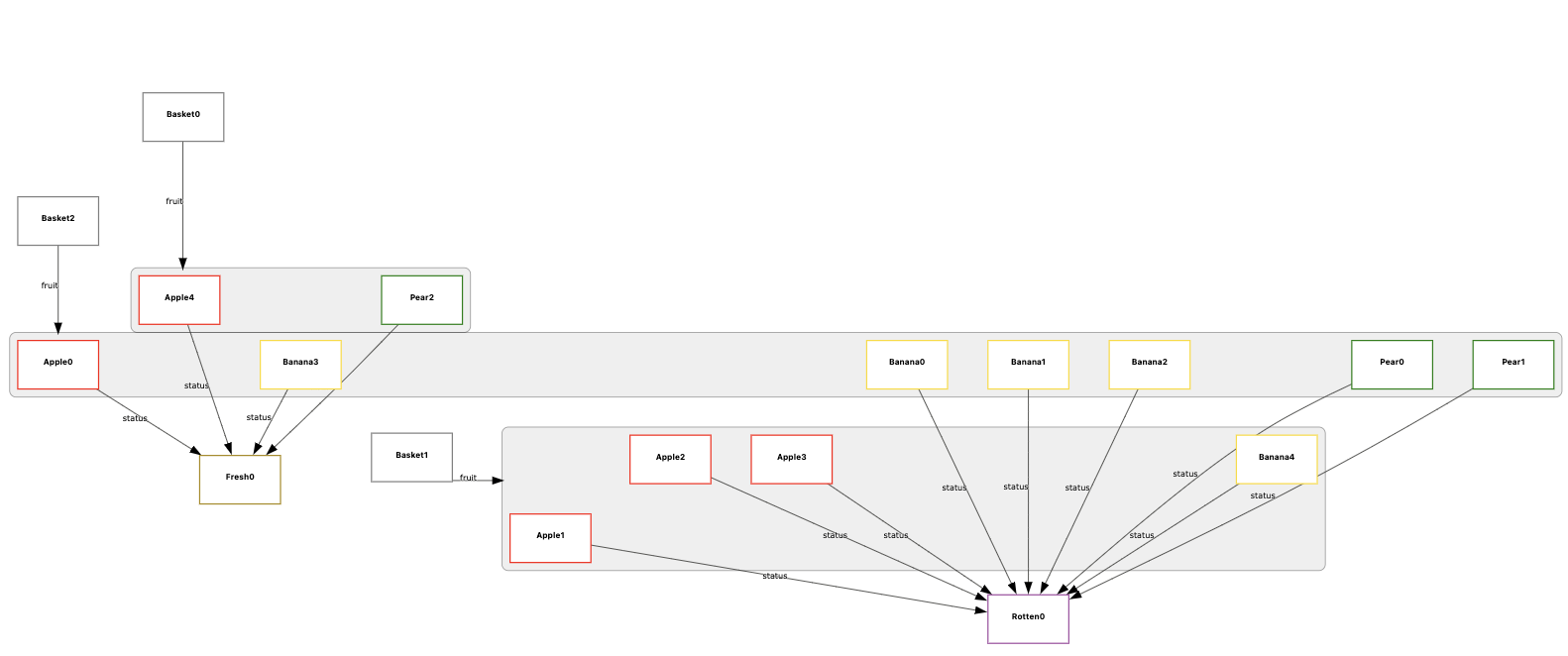}
    \caption{\pill{fruit} Fruit with default nodes.}
    \label{f:fruit}
  \end{subfigure}
  \\
  \begin{subfigure}[b]{0.9\textwidth}
    \centering
    \includegraphics[width=\textwidth]{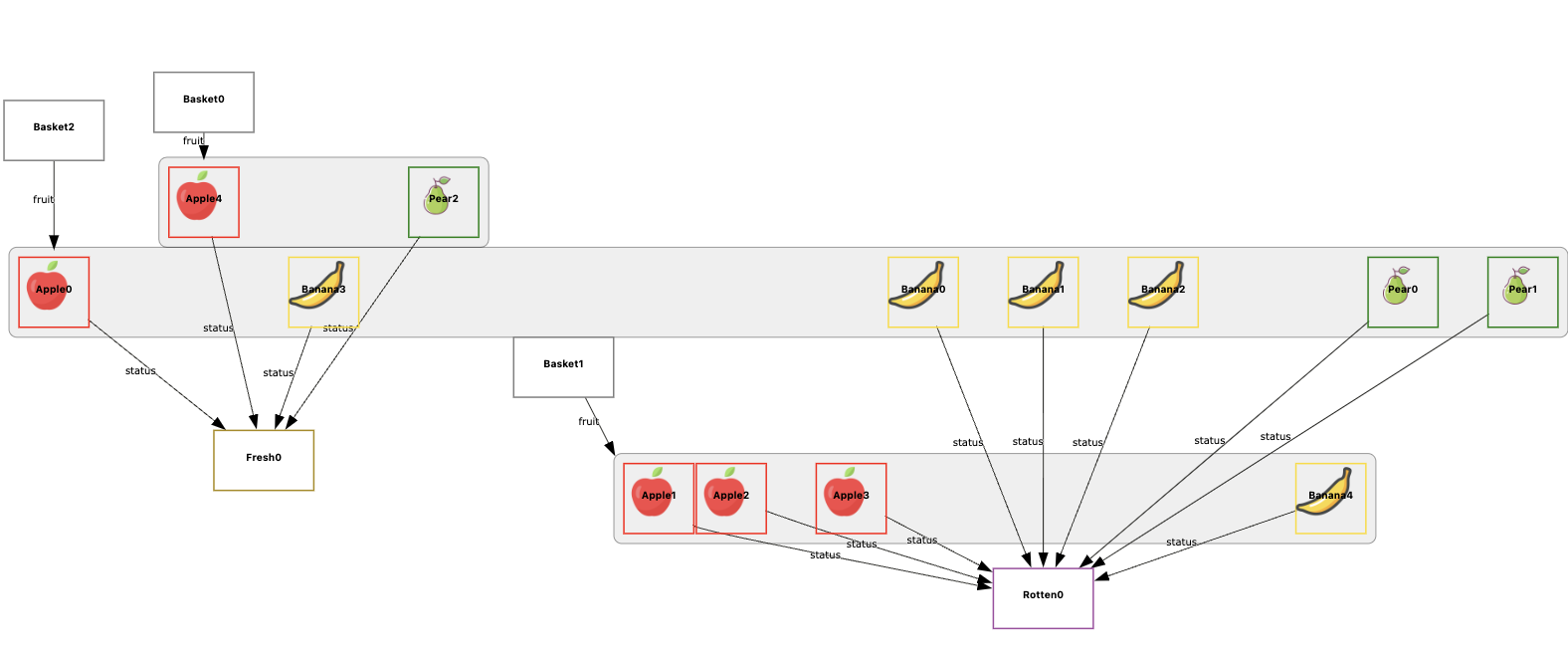}
    \caption{\pill{fruit-icons} Fruit with icons.}
    \label{f:fruit-icons}
  \end{subfigure}
  \caption{
    \cnd{} visualizations of the fruit scenario described in \cref{t:study1-correctness}.}
  \label{f:p2-figs}
\end{figure}

In order to test the efficacy of pictorial directives, we conducted a 
second study extending the Fruit scenario described in \Cref{s:prolific-model}.
While the scenario remained unchanged, participants (n = 30) were
assigned one of two variations of the fruit diagram: one with \cnd{}'s
default node representation (\cref{f:fruit}),
and one with images representing different fruit types (\cref{f:fruit-icons}).

Participants shown the diagram with pictorial directives answered questions correctly more often, and faster
than those shown the diagram without them ($86.67\%$ vs $73.33\%$ correct, mean time-on-task: $56.76$s vs $68.55$s). 
However, these were not statistically significant at a 95\% confidence level 
(correctness: $Z = 1.60$, $p = 0.11$, time-on task: $t=-0.96$, $p=0.34$).

Participants were then shown the fruit scenario with three different diagram styles:
 \ASV{}, default \cnd{}, and \cnd{} with pictorial directives, and asked to rank them in order of preference.
Participants tended to rank the \cnd{} with icons diagram highest (mean rank $1.45$), followed by the \cnd{} 
with default nodes (mean rank $2.00$), and then the \ASV{} (mean rank $2.57$).
This difference in ranking was statistically significant (Friedman $Q = 19.27$, $p = 6.55e-5$).
A pairwise comparison showed that the ranking for \cnd{} with icons was significantly higher than that 
of the \ASV{} ($p = 3.4e-5$).
Participants did identify that the \cnd{} with icons diagram conveyed the same information as
\cnd{} with default nodes, but still showed a slight preference for the former ($p = 0.07$)
in their rankings.

Participant feedback suggested that the icons helped them quickly identify
the types of fruit in the diagram, making it easier to answer questions.

\quoteparticipant{26}{I ranked [\cnd{} + pictorial directives] the highest because its clarity is enhanced by the structured layout,
 and the fruit photos make the relationships more visible and easy to understand. [Default \cnd{}] is similar but slightly less effective because it 
 lacks the fruit photos, making it harder to visualize the elements quickly. [The \ASV{}] is the least effective, as its layout is too complex, making it difficult to understand the connections.}
\quoteparticipant{29}{[The \ASV{}] is the least clear compared to [\cnd{} default] and [\cnd{} + pictorial directives]. [\cnd{} + pictorial directives] is better because it includes a graphical representation of the information.}

\quoteparticipant{18}{Because it is much simpler to find what you are looking for using [\cnd{} + pictorial directives]
 (seeing the name and the fruit) and [\cnd{} default], as for [the \ASV{}] many may find it difficult to find what they are looking for.}

\subsubsection{Bad-instances}
\label{s:prolific-non-model}

\begin{table}[h]
  \centering
    \caption{Description of scenarios used to study the effectiveness of \cnd{} constraints on understanding bad-instances.}
    \label{t:study3-scenarios}
  \begin{tabularx}{\textwidth}{lXX}
    Scenario         & Good Instance                                                                  & Bad Instance                          \\
    \hline
    Tic-Tac-Toe      & 3x3 grid of Nodes with Xs and Os representing a Tic-Tac-Toe board   & 9 node graph marked with Xs and Os that do not form a grid \\
    Face             & Facial features arranged to represent a valid human face.           & Facial features arranged so that an eyebrow is aligned with the eyes.       \\
  \end{tabularx}

\end{table}

To identify \emph{bad} instances, the participant must have some
intuitive sense of what constitutes a \emph{good} instance. To avoid
the confound of training (which some participants may do only
half-heartedly), we picked two situations that users are likely to be
able to implicitly judge: the game Tic-Tac-Toe and a human face.  For
each scenario, we constructed a valid and invalid instance and
visualized them using the \ASV{} and \cnd{}.

\begin{figure}[t]
  \centering
  \begin{subfigure}[b]{0.5\textwidth}
    \centering
    \includegraphics[width=\textwidth]{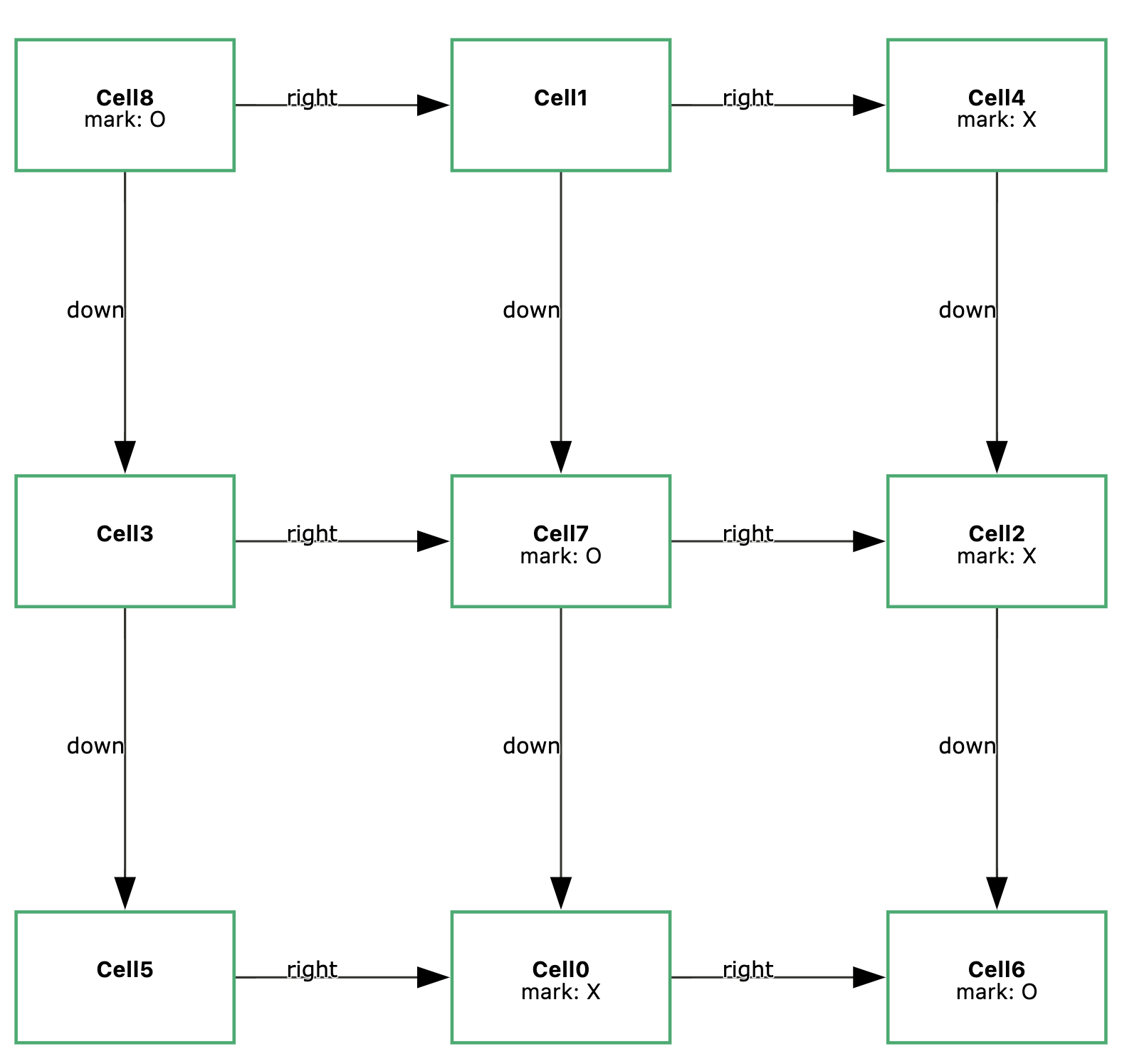}
    \caption{\pill{ttt} \cnd{} diagram of the valid Tic-Tac-Toe board.}
  \end{subfigure}
  \hfill
  \begin{subfigure}[b]{0.4\textwidth}
    \centering
    \includegraphics[width=\textwidth]{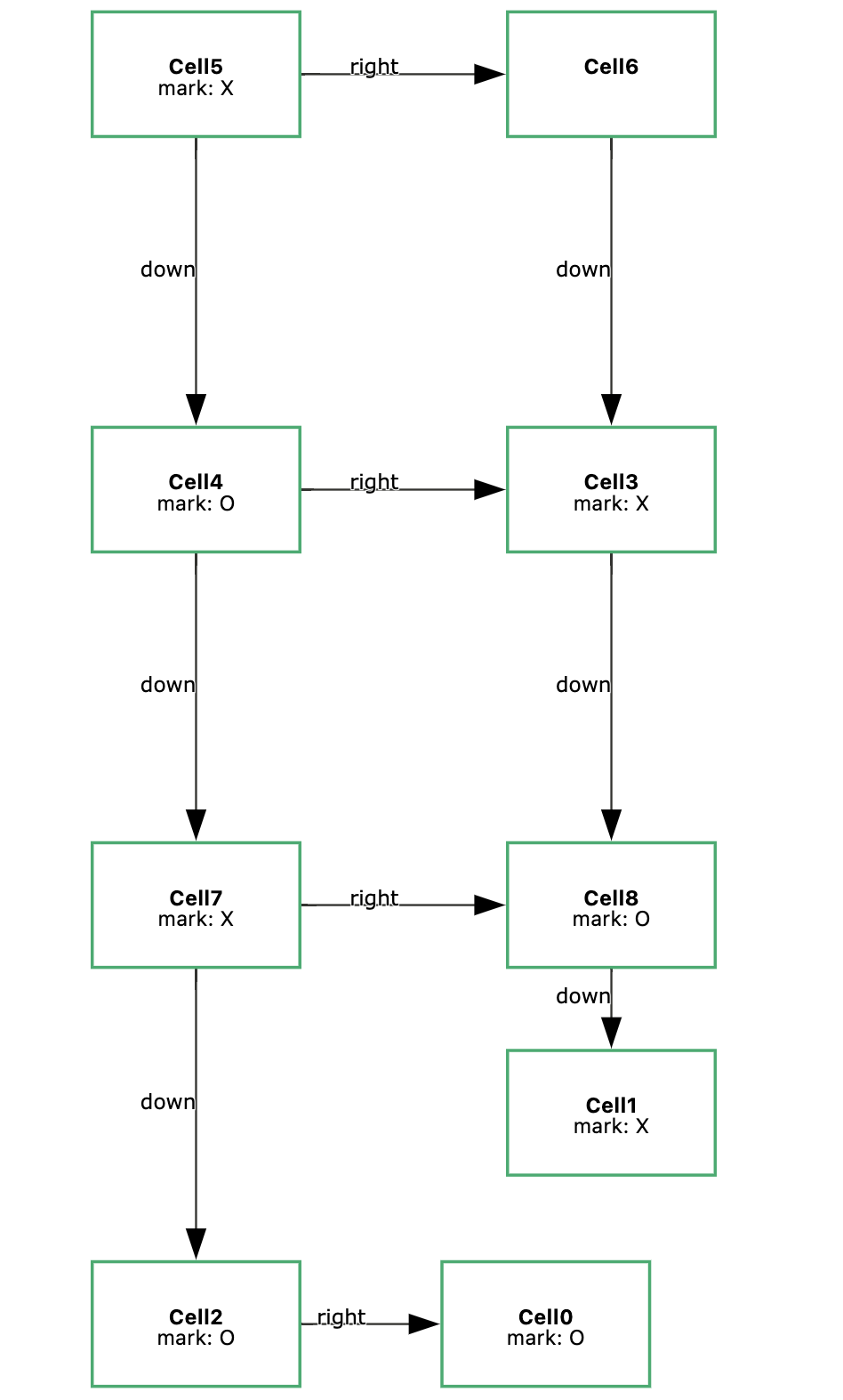}
    \caption{\pill{ttt-inv} \cnd{} diagram of the invalid Tic-Tac-Toe board.}
  \end{subfigure} \\
  \begin{subfigure}[b]{0.4\textwidth}
    \centering
    \includegraphics[width=\textwidth]{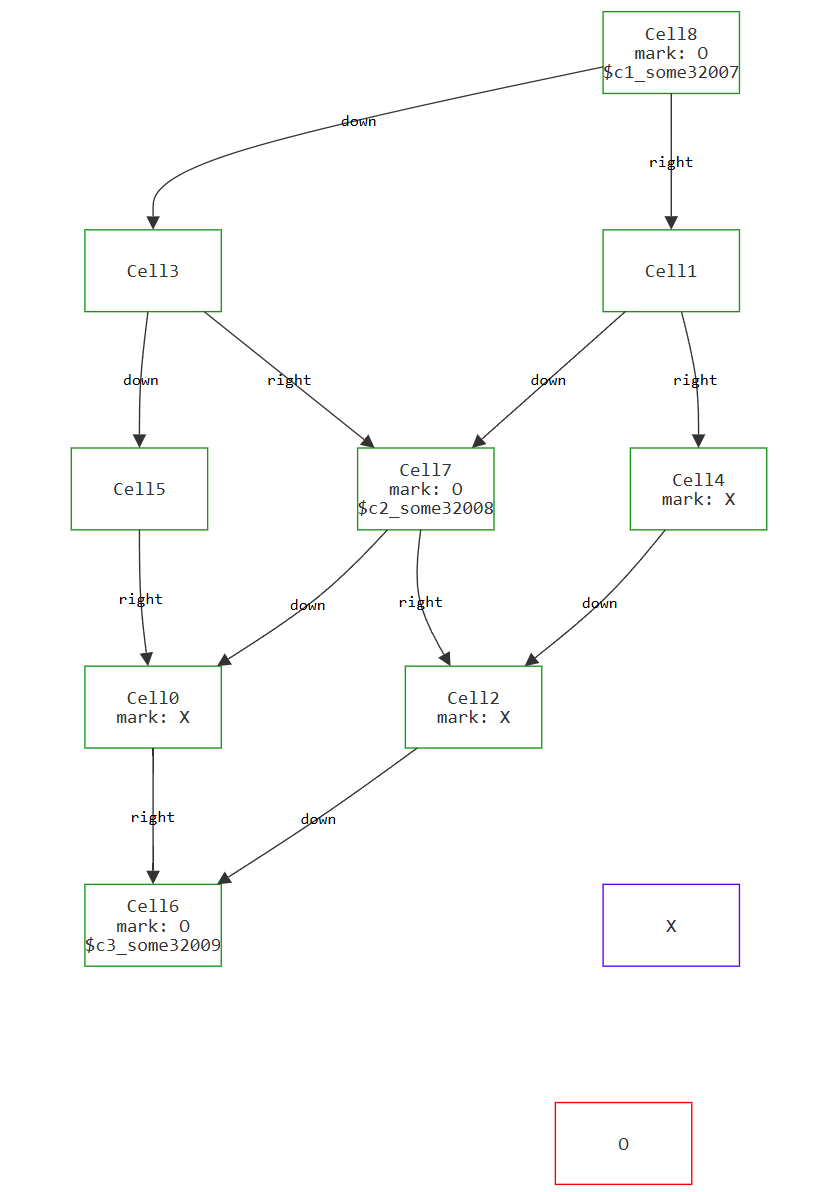}
    \caption{\ASV{} diagram of the valid Tic-Tac-Toe board.}
  \end{subfigure}
  \hfill
  \begin{subfigure}[b]{0.4\textwidth}
    \centering
    \includegraphics[width=\textwidth]{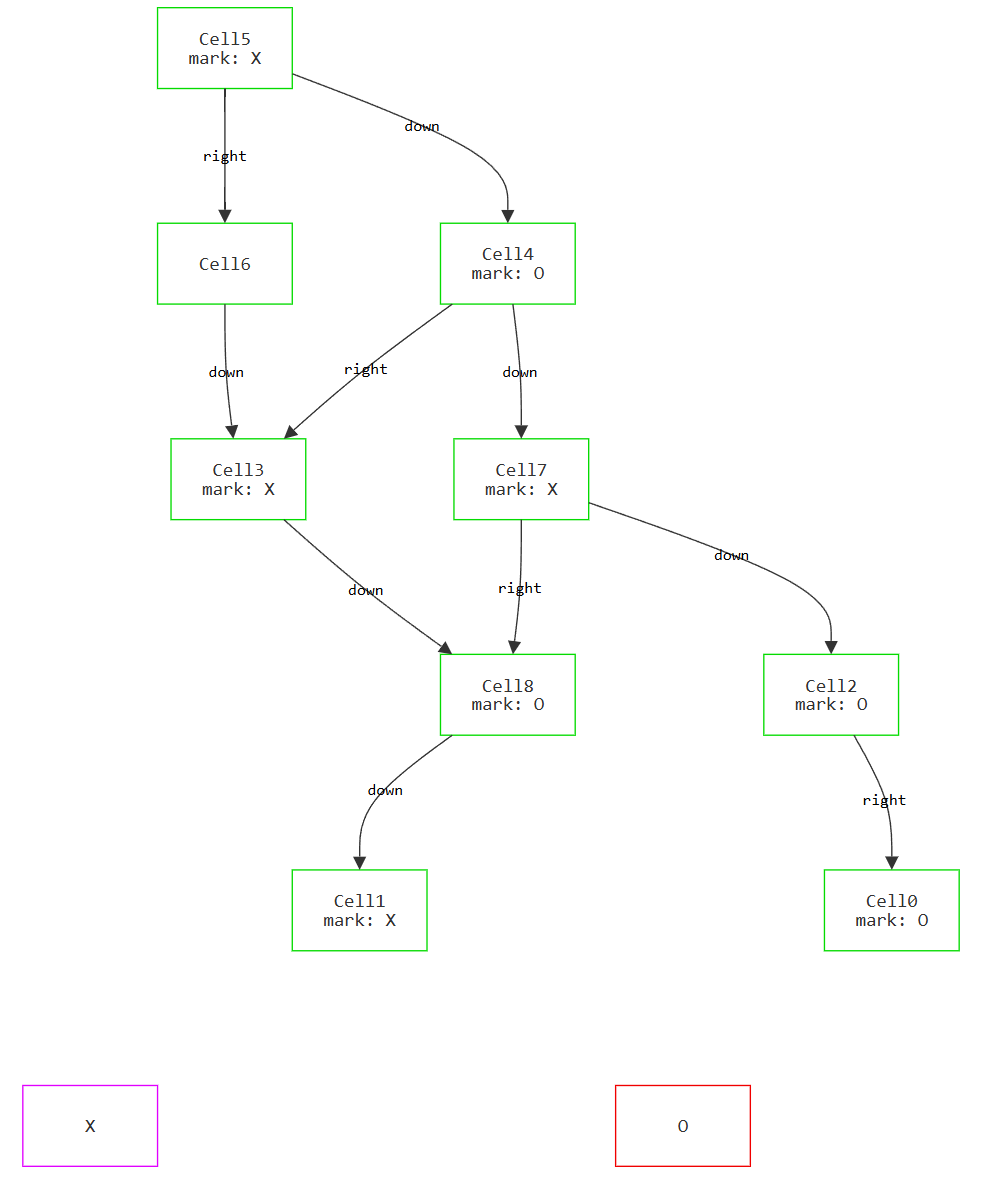}
    \caption{\ASV{} visualization of the invalid Tic-Tac-Toe board.}
  \end{subfigure}

  \caption{
    The diagrams of valid and invalid Tic-Tac-Toe boards, as described in \cref{t:study3-scenarios}.}
  \label{f:ttt}
\end{figure}

\begin{figure}[t]
  \centering
  \begin{subfigure}[b]{0.45\textwidth}
    \centering
    \includegraphics[width=\textwidth]{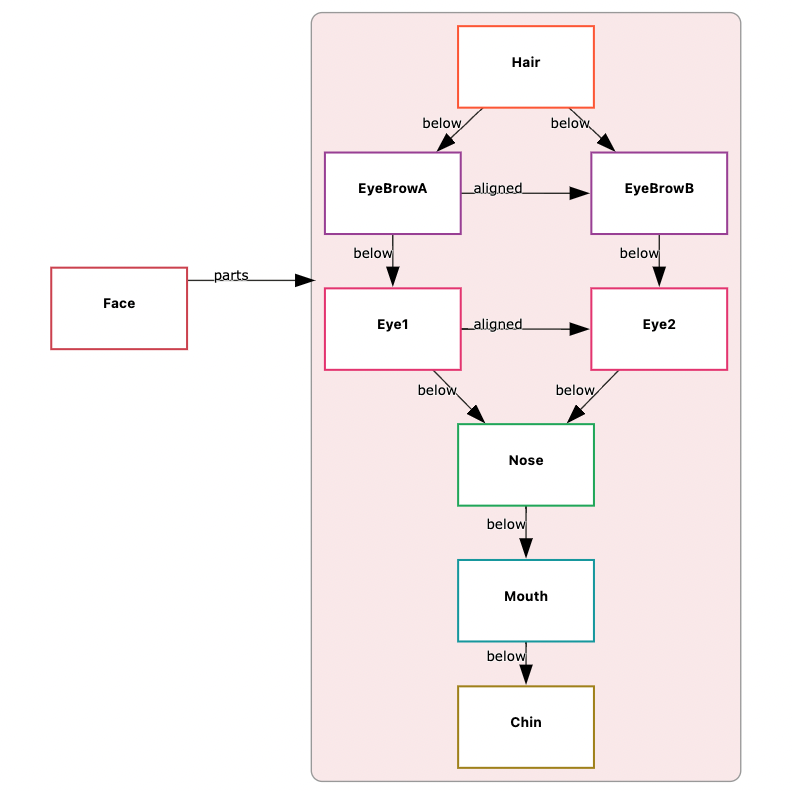}
    \caption{\pill{face} \cnd{} diagram of the valid face instance.}
  \end{subfigure}
  \hfill
  \begin{subfigure}[b]{0.45\textwidth}
    \centering
    \includegraphics[width=\textwidth]{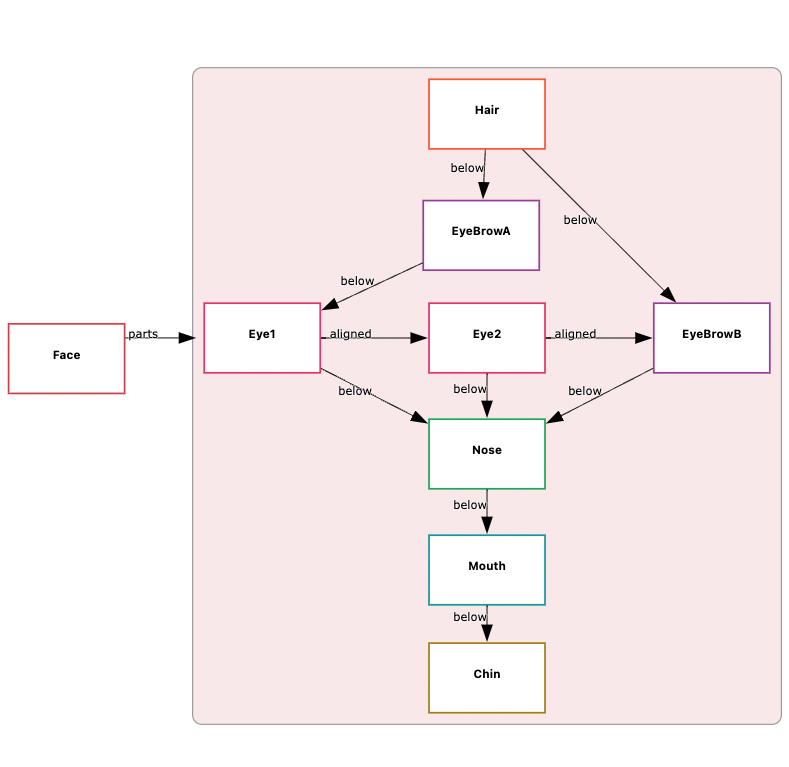}
    \caption{\pill{face-inv} \cnd{} diagram of the invalid face instance.}
  \end{subfigure} \\

  \begin{subfigure}[b]{0.45\textwidth}
    \centering
    \includegraphics[width=\textwidth]{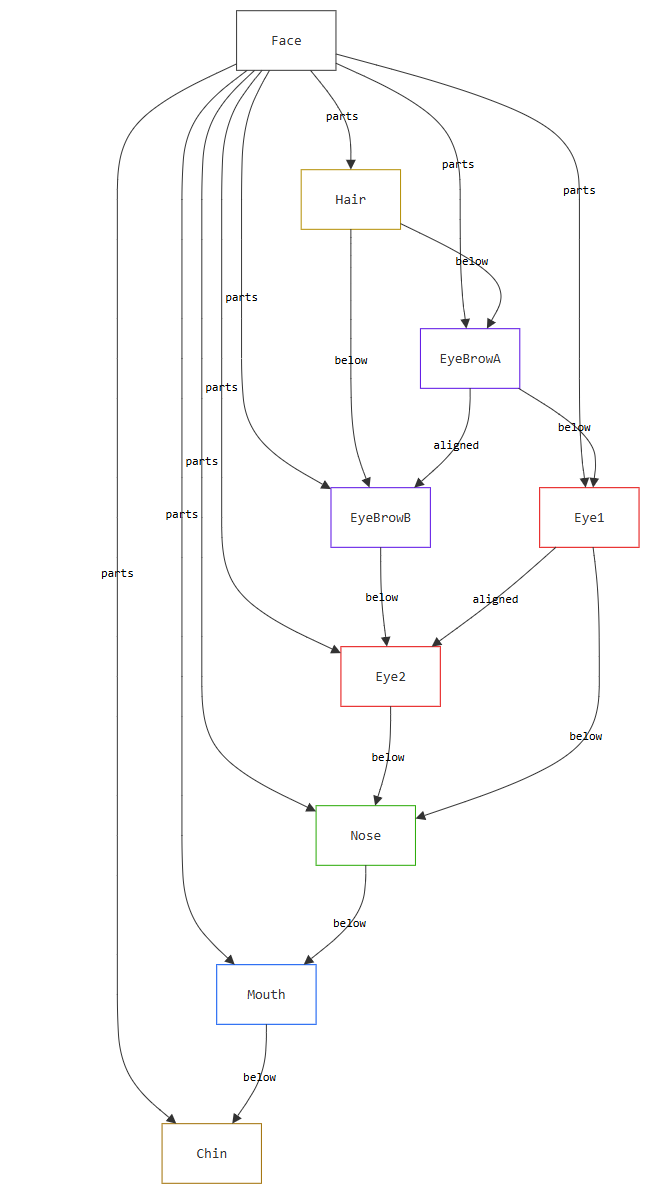}
    \caption{\ASV{} diagram of the valid face instance.}
  \end{subfigure}
  \hfill
  \begin{subfigure}[b]{0.45\textwidth}
    \centering
    \includegraphics[width=\textwidth]{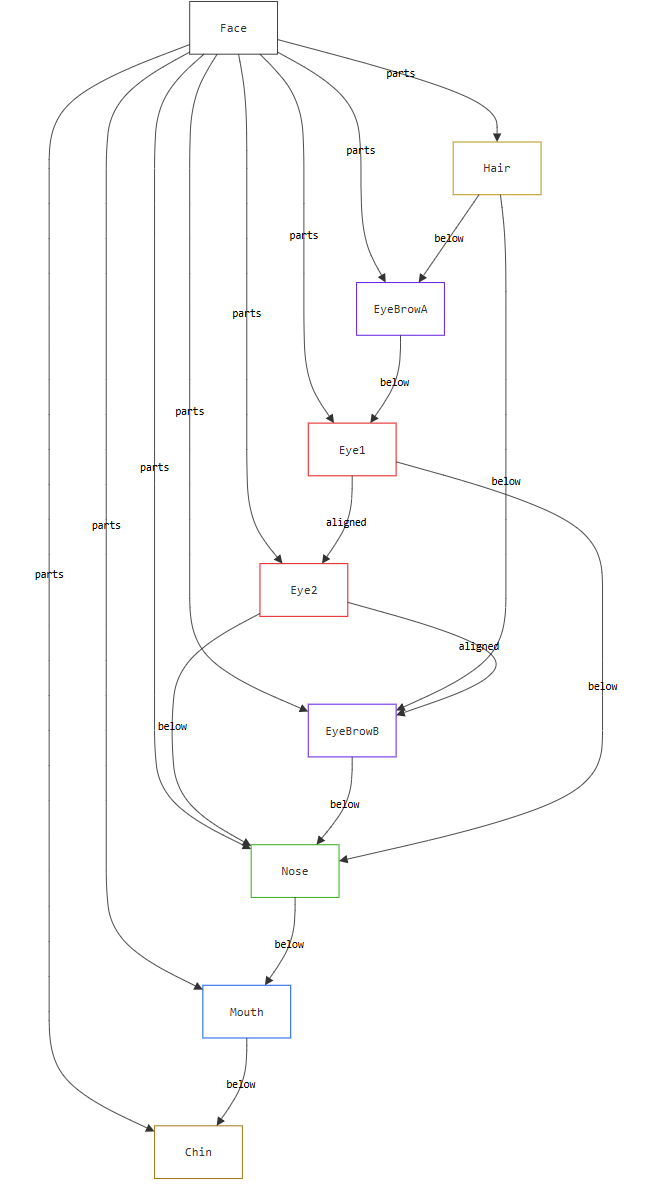}
    \caption{\ASV{} diagram of the invalid face instance.}
  \end{subfigure} 
  \caption{
    Diagrams of valid and invalid face instances, as described in \cref{t:study3-scenarios}.}
  \label{f:face}
\end{figure}

Out of n = 40 participants, half of them saw \cnd{} diagrams for
Tic-Tac-Toe and Sterling diagrams for the Face scenario, while the
other half saw the reverse. For each scenario, participants were first
shown a diagram of a valid instance, asked to judge its validity, and
then asked a follow-up question about the scenario 
that required them to parse the diagram's structure.
This helped build
familiarity with the scenario, the diagramming style, and key aspects
of the model. Participants were then shown a diagram of an invalid
instance, and asked to judge its validity. (Of course, participants did
not \emph{know} they were being shown instances in this order: they
were simply passing judgment on two potential instances.)

We divide responses into two groups:
\begin{itemize}
    \item Validity judgements: The participant asserted that the diagram was either valid or invalid. 
    Participants who said a diagram was invalid were asked to explain why.
    \item Uncertain responses: The participant was unsure about the
      diagram's validity. These participants were asked why they were uncertain.
\end{itemize}

\paragraph{Uncertain Responses}

It is tricky to interpret uncertainty.
If a user is spot-checking a spec, uncertainty may arouse their suspicions, leading them to investigate further.
On the other hand, if a user is browsing a spec, they may simply move
on to the next instance.
However, we did not examine this in depth because relatively few participants expressed uncertainty about validity.
Only 5 participants ($12.5\%$, mean time = $37.84$s) were unsure about the validity of \cnd{} diagrams, 
while 3 participants ($7.5\%$, mean time = $29.52$s) were not sure about the validity of Sterling diagrams.

\paragraph{Validity Judgements}

Participants were significantly more likely to identify bad-instances and
correctly explain \emph{why} they were bad 
when shown \cnd{} diagrams than when shown the \ASV{} ($71.43\%$ vs $43.24\%$, $p = 0.03$, $\chi^2 = 4.73$).

Participants also made these validity judgements faster with \cnd{}
(mean of $58.06$s vs.\ $69.82$s),
but this difference was not statistically significant ($p = 0.37$, $U = 568.00$).

\subsubsection{Threats to Validity}
\label{s:threats}

Several factors may limit the generalizability and interpretation of our findings.
While study participants had prior programming experience, this does not imply familiarity with
formal methods, Alloy, or the \ASV{}. Effects on understanding could differ for users with more 
experience in these areas~\cite{mansoor2023empirical}.

While we attempted to control for domain expertise by selecting scenarios that were either novel 
(\cref{s:prolific-model,s:prolific-icons}) or universally familiar 
(\cref{s:prolific-non-model}), different kinds of expertise could affect
understanding. For instance, the ability to interpret a diagram might be different if the interpreter
is also the author of the specification.

Finally, as with any online study, there are inherent threats related to participant engagement and motivation, 
which could influence the outcomes. We believe that these limitations do not invalidate our findings,
but they do warrant caution in interpreting the results as directly applicable to
experienced Alloy users in real-world settings.

\clearpage

\section{Related Work}
\label{s:rel-work}

\cndFull{} occupies a middle ground between default visualizations,
which often lack domain-specific semantics, and programmatic tools,
which require users to define every aspect of the diagram from scratch.

To position \cnd{} within the landscape of existing tools, we explore three major categories of related work:
\begin{enumerate}
  \item Generic visualizers that provide default visualizations for Alloy instances (\cref{s:asv}).
                While they are flexible and broadly applicable, their lack of domain-specific insight
                can lead to confusing visualizations (\cref{s:bottom-up}).

  \item Drawing and diagramming tools that offer users fine-grained control over visualizations,
         enabling them to craft detailed domain-specific diagrams (\cref{s:comp-d3,s:penrose,s:bluefish,s:gupu,s:prob}).
        This control demands significant effort. Users have to manage the nuts-and-bolts
        of diagram creation, dealing not only with domain-level constructs,
        but also actual ``drawing'' primitives (e.g., lines, shapes, points).

  \item Layout generation tools that offer users automated placement of diagram elements (\cref{s:webcola,s:dagre}).
        These are not diagramming tools in themselves but serve as essential building blocks for other systems.
\end{enumerate}

\subsection{\ASVs{}}
\label{s:asv}

The Alloy Visualizer~\cite{j-alloy-2012} is the default visualizer for Alloy (and Alloy-like) specifications.
The visualizer generates directed graph visualizations of the spec instance,
 with nodes representing atoms and edges representing relations between them.
Users are provided a range of theming options, such as being able to change the color or shape of graph nodes 
or hide nodes via projection. Users, however, struggle to make sense of these as specifications grow in size 
and complexity~\cite{mansoor2023empirical}.

Sterling~\cite{dyer2021sterling} is a web-based variant of the Alloy Visualizer that is designed to make
large instances more comprehensible. Nodes and edges are laid out hierarchically, reducing cognitive load, 
while layouts are consistent across instances, making the differences between them more salient.

We group both these tools together as \ASVs{} -- default visualizers that do not encode any domain-specific 
knowledge about the spec being visualized. Although both Alloy and Sterling provide numerous theming options, 
users cannot exercise control over \emph{layout} decisions,
leading to a lack of exploration support and live engagement. On the other hand, the simple node-edge representation
offers a high degree of representation salience, vocabulary correspondence, and robustness to bad-instances.

\subsection{\SDT{}}
\label{s:comp-d3}

In addition to its default layout, Sterling also allows users to define custom visualizations
using the D3 visualization library~\cite{bostock2011d3}.
D3 is a powerful, expressive JavaScript library that gives users a great deal of control over the 
appearance of their diagrams. This means that users can create diagrams highly tailored to the
domain being modeled, and use a wide range of visual encodings to represent relationships between entities.

This complexity, however, comes at a cost. We discuss the barrier to adoption of D3 in \cref{s:pain-barrier}.

\subsection{Penrose}
\label{s:penrose}
Penrose~\cite{ye2020penrose} is a domain-specific language designed to 
build diagrams of mathematical objects.
The language's design is informed by both
the Gestalt principles (\cref{s:visual-principles}) and the diagramming tool
requirements described in \cref{s:diagramming-tools}.

Penrose introduces a structured approach to the diagramming process by 
requiring users to separate the content of their diagram from how it is drawn.
Diagrams, therefore, are constructed as three sub-programs: domain, substance, and style.
The domain specifies the rules of the mathematical object being diagrammed,
the substance specifies the instance of the object being visualized, 
and the style specifies how the object should be visualized.
This separation of concerns makes Penrose a very powerful diagramming language, with 
the ability to generate diagrams across a wide range of domains.

Penrose substances and domains can be thought of as analogous to 
specifications in formal modeling languages. 
A Penrose domain is similar to an Alloy specification,
while a Penrose substance corresponds to an instance of that specification.
In contrast, ``style files'' are more akin to programs.
Users must describe how diagram elements
should be constructed from rendering primitives
(e.g., lines, shapes, points) before they can 
write higher-level constraints on how these elements
should be laid out. Given the choice of rendering primitives,
a user may even have to specify the order in which higher-level
constraints are applied.

This represents a heavy up-front cost.
A user must have already built the diagram from the ground up before 
they can begin to add layout constraints
relevant to their domain.

Penrose's heavy up-front cost disincentivizes its use as an exploratory tool.
The Penrose directed graphs demo,\footnote{https://github.com/penrose/penrose/tree/main/packages/examples/src/box-arrow-diagram}
for instance, involves nearly 150 lines of style code, including 13 constraints, 4 constraint application phases,
and complex visualization calculations (e.g., for the angle of incidence of arrowheads on nodes).
Much of this would have to be re-written if the shape of nodes were to change from circles to, say, squares.

\begin{figure}[t]
  \centering
  \begin{subfigure}[b]{0.45\textwidth}
    \centering
    \includegraphics[width=\textwidth]{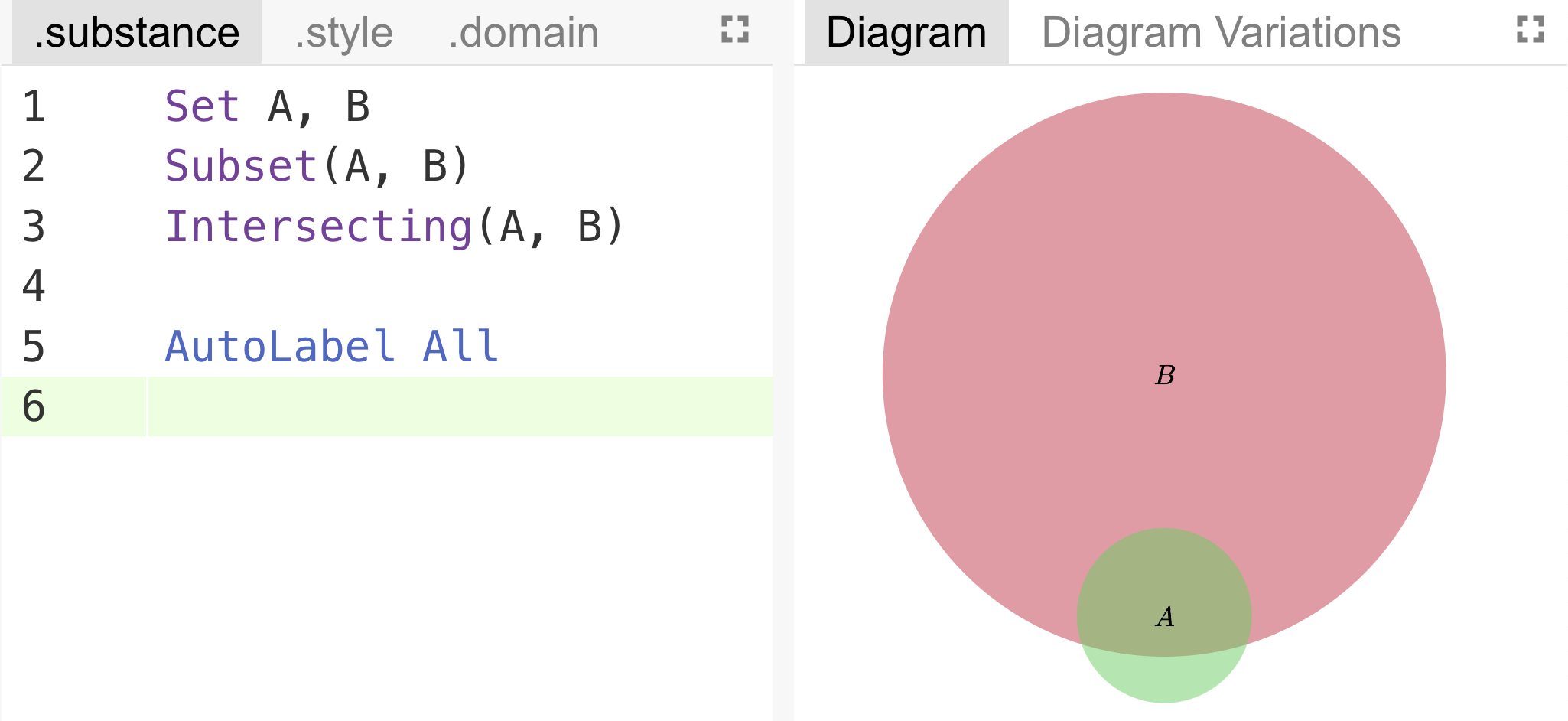}
    \label{f:penrose-bad-inst}
    \caption{Misleading Penrose visualization of a bad-instance.
    The style file implicitly assumes that intersecting
    sets cannot subsume one another. As a result,
    the circle A is not fully subsumed by the circle B in visualization,
    despite A being a subset of B.}
  \end{subfigure}
  \hfill
  \begin{subfigure}[b]{0.45\textwidth}
    \centering
    \includegraphics[width=\textwidth]{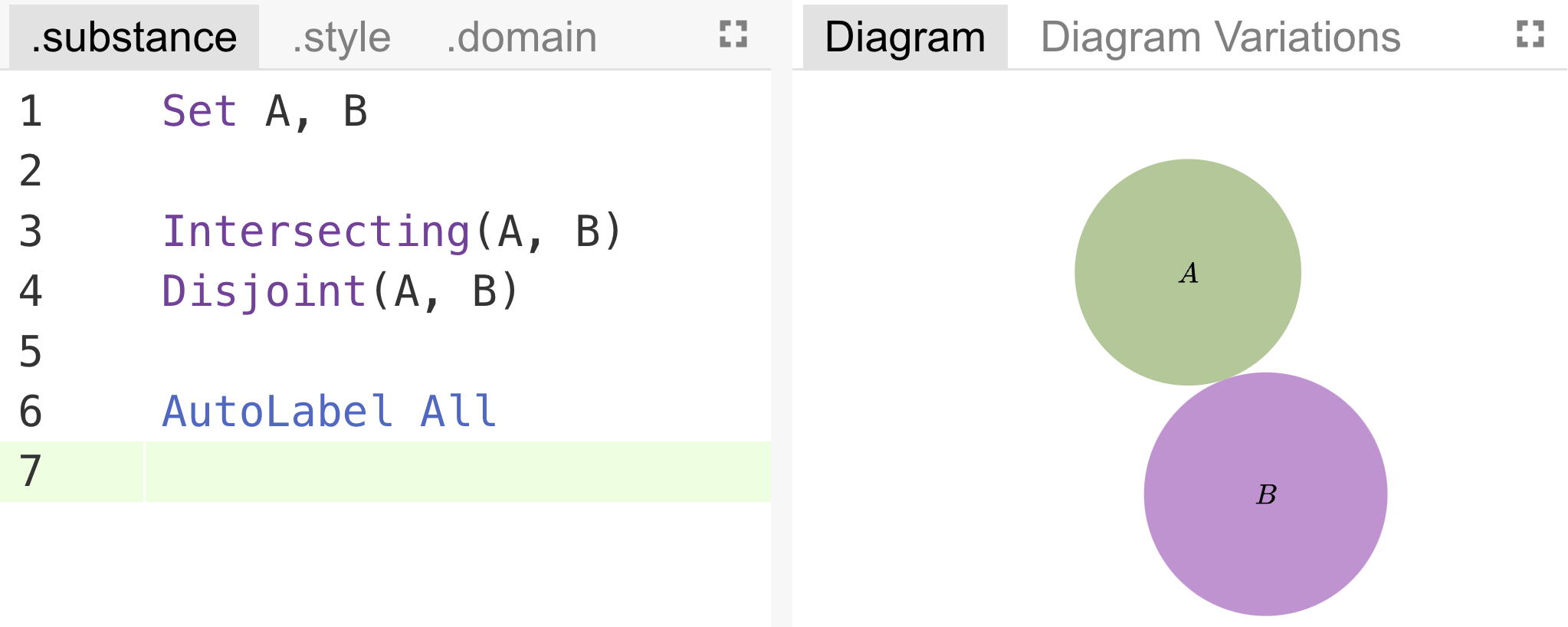}
    \caption{Penrose visualization of an impossible instance.
            Sets A and B are both both intersecting and disjoint.
            While layout is unsatisfiable, 
            Penrose still generates a (misleading) diagram.}
    \label{f:penrose-unsat}
  \end{subfigure}
  \label{f:penrose-failures}
  \caption{Unexpected Penrose behaviors for a bad-instance and an
    instance where layout constraints are unsatisfiable.}
\end{figure}

After spending significant time on the nuts-and-bolts of 
rendering, a Penrose user might plausibly still find themselves
looking at a misleading visualization. This could be for one of two reasons:
\begin{enumerate}
  \item Since users have to build visualizations from scratch, Penrose diagrams
        may be brittle to bad-instances (\cref{s:when-viz-fail}). 
        We show an example of this in \cref{f:penrose-bad-inst}, based on the Penrose Euler Diagrams demo.\footnote{https://penrose.cs.cmu.edu/try/?examples=set-theory-domain\%2Ftree-euler}
  \item Penrose generates diagrams even if constraints are not satisfied:
        Penrose's numerical optimization problem is designed to always find a solution.
        This means that that the system will generate diagrams even when 
        some user-specified constraints are not satisfied (\cref{f:penrose-unsat}).
\end{enumerate}

\subsection{BlueFish}
\label{s:bluefish}

Bluefish~\cite{pollock2024bluefish} is a declarative diagramming framework inspired by
the principles of component-based UI frameworks. 
The core Bluefish primitive is the ``relation'', which is 
used to capture semantic associations between diagram elements.
Bluefish is particularly well suited to diagramming because these relations are composable
and extensible. Relations do not need to fully specify how elements are laid out,
and so can share children elements. 

The relaxed nature of relation composability, however, comes at a cost.
BlueFish relations act on diagram elements, and not their types.
This means that every Bluefish program needs to explicitly encode
each diagram element and all its relationships. 
For example, an author must have knowledge of the structure and relative positions
of every \emph{node} in a binary tree to write a relation that ensures that all left descendants
are laid out to the left of their ancestors. This is especially problematic in
the context of formal methods, where the exact elements of an instance 
(or bad-instance) are not known in advance.

\subsection{Pro~B}
\label{s:prob}
The Pro~B~\cite{leuschel+butler:fme03:prob} suite provides multiple tools 
(e.g., BMotionWeb~\cite{ladenberger+leuschel:sefm16:bmotionweb} 
and VisB~\cite{werth+leuschel:abz2020:visb})
for visualizing model traces. Moreover the
Alloy2B~\cite{krings++:abz18:alloy-to-b} project can translate a rich 
subset of Alloy to B, meaning that Pro~B could potentially be used for 
some Alloy visualizations. These tools are more powerful, but also more 
heavyweight than \cnd{}. 
For example, BMotionWeb~\cite{ladenberger+leuschel:sefm16:bmotionweb} 
enables interactive, domain-specific visualizations, but creating these visualizations 
requires some knowledge of web programming. In VisB~\cite{werth+leuschel:abz2020:visb},
users modify attributes of a base SVG image by linking elements of the SVG to aspects 
of the underlying model. In contrast, \cnd{} uses lightweight, declarative constraints 
to control the layout and presentation of Alloy's existing visual idioms.

\subsection{GUPU}
\label{s:gupu}

GUPU~\cite{neumerkel2002declarative} is a pedagogic tool that allows users visualize 
Prolog substitution via domain-specific ``viewers''. 
Students can use these graphical tools alongside tailored feedback 
and testing capabilities to better understand Prolog solutions.
GUPU's viewers are written in Prolog~\cite{neumerkel1997visualizing}, and 
are capable of generating sophisticated diagrams.
Unlike \cnd{} specifications, however, GUPU viewers are heavyweight.
Authors must invest significant time to get started, specifying
 each diagram construct, the minutiae of layout, and 
reconstructing the relationships between diagram elements.
This is a significant barrier to entry:
custom viewers presented by GUPU authors~\cite{neumerkel1997visualizing}
take the form of Prolog programs that are much more complex than
the Prolog programs they visualize.

\subsection{WebCola}
\label{s:webcola}

WebCola~\cite{webcola}  is a JavaScript library designed to support constraint-based graph layout.
It is used as an optional layout engine across a variety of domain-specific layout tools, including D3, SetCola~\cite{setcola}, and CytoScape~\cite{shannon2003cytoscape}.
WebCola constraints are specified in terms of a graph's nodes, and can be used to control node alignment, separation, and grouping.
These constraints, however, are always ``soft''. This means that WebCola will always generate a layout, even if the constraints are not satisfied.
This silent failure can lead to misleading diagrams that do not accurately reflect the spec being visualized.
Nevertheless, WebCola is still useful for layout, so \cnd{} uses
WebCola \emph{after} \cnd{} constraints are deemed consistent and 
satisfiable (see \cref{s:unsat}).

\subsection{DAGRE}
\label{s:dagre}

DAGRE is a JavaScript library for generating directed graph layouts,
organizing nodes into hierarchical layers to minimize edge crossings and improve readability~\cite{dagre}.
While DAGRE allows users to exercise some control over graph layout (e.g., specifying the direction of graph flow and node separation),
users are unable to control the layout of individual nodes or edges. 
DAGRE's focus on hierarchical layout means that it is not well suited to more general graph layouts (e.g., cyclic graphs).
Sterling uses DAGRE to generate layouts for its default visualizations, so
\cnd{} uses DAGRE to generate layouts if no layout constraints are specified.

\section{Limitations}
\label{s:limitations}

\begin{figure}[t]
  \centering
  \begin{subfigure}[b]{0.45\textwidth}
    \centering
    \includegraphics[width=\textwidth]{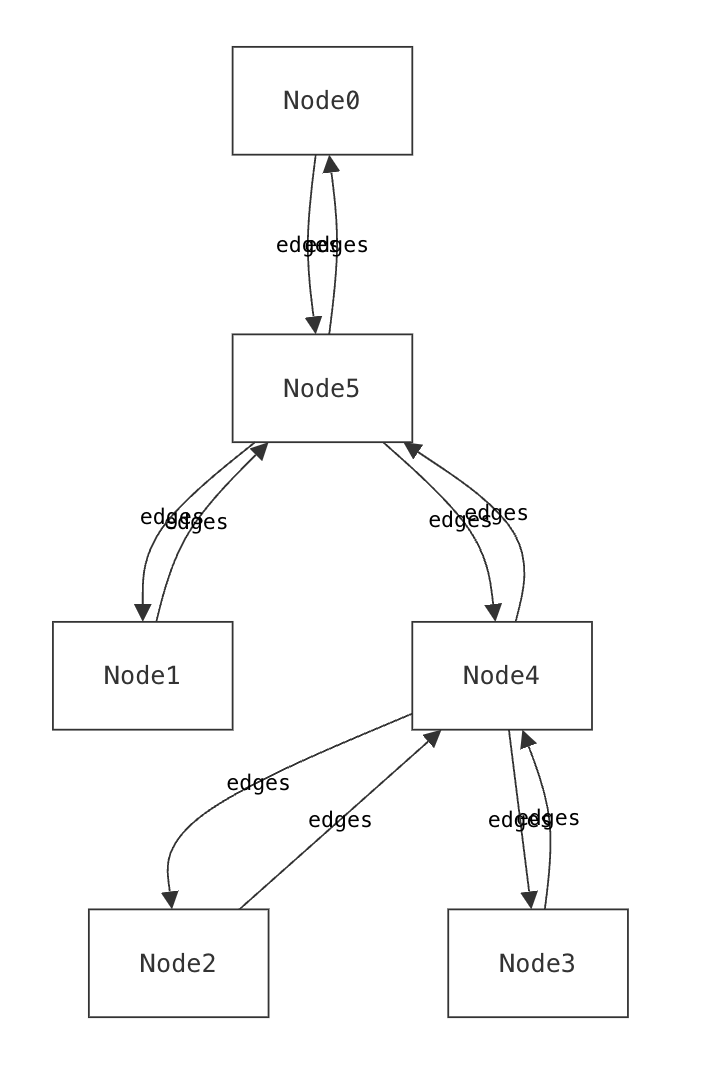}
    \caption{\ASV{}}
    \label{f:undirected-tree-asv}
  \end{subfigure}
  \hfill
  \begin{subfigure}[b]{0.45\textwidth}
    \centering
    \includegraphics[width=\textwidth]{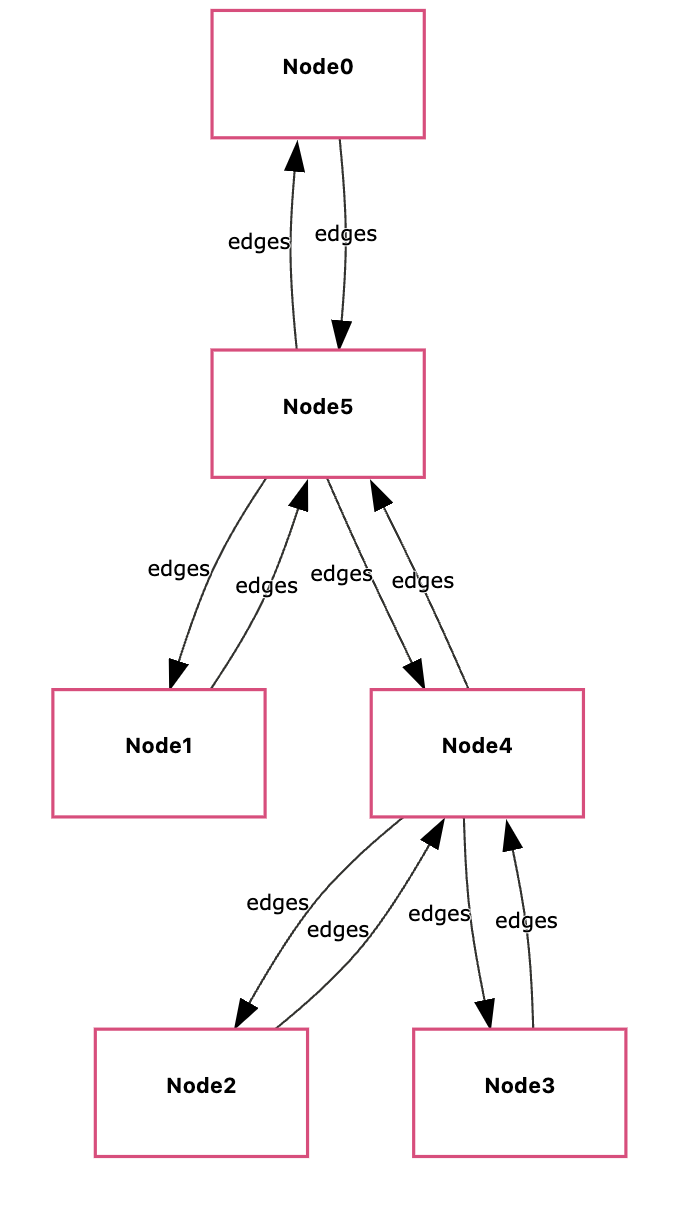}
    \caption{\pill{undirected-tree} \cnd{}}
    \label{f:undirected-tree-cnd}
  \end{subfigure}

  \caption{An undirected tree specification visualized by the \ASV{} and using \cnd{}.
    The \cnd{} diagram is generated with no constraints and/or directives.}  
  \label{f:undirected-tree}
\end{figure}

Since \cnd{} acts as a refinement on the \ASV{} graphs, it inherits many of the \ASV{}'s limitations.
For instance, layout constraints may not be applicable to non-functional, non-hierarchical relations.
Consider the example of an undirected tree, where each edge between nodes takes 
the form of a symmetric relation (\cref{f:undirected-tree}).
Since any orientation constraint on symmetric relations is necessarily inconsistent,
\cnd{} cannot enforce a layout on the tree based on its edges.

Furthermore, \cnd{} constraints cannot add \emph{new} information not 
reflected in the \ASV{} diagram.\footnote{Pictorial directives, however, can communicate new information.}
The instance in \cref{f:undirected-tree} does not contain any information 
on which node is the parent of another.  Thus, a diagrammer cannot purely rely on \cnd{}
to exercise control over which node is visualized as the root of the tree.
In order to do so, they have recourse to one of two options:
\begin{enumerate}
\item Indicate the root of the tree by means of a field or sig to their spec.
\item Build a custom visualization that encodes the root of the tree.
      In this case, the custom visualization functions as a second repository 
      of implicit spec information.
      As discussed in \cref{s:when-viz-fail},
      this opens up the possibility of misleading visualizations of bad-instances.
\end{enumerate}

A silver lining here, however, is that the worst-case scenario for \cnd{} is the \ASV{} visualization. 
Indeed, the visual principles involved in the \cnd{} design process (\cref{s:visual-principles})
typically lead to less cluttered, more informative diagrams than the \ASV{}, even in the absence of constraints.
The \cnd{} diagram in \cref{f:undirected-tree-cnd}, for instance, does not exhibit overlapping edge
labels like the \ASV{} diagram in \cref{f:undirected-tree-asv}.
In general,
\cnd{} diagrams are always at least as informative as \ASV{} diagrams, and never less so.

\section{Discussion}

In personal communication, Daniel N.~Jackson (the lead architect of
Alloy) once explained to the last author the difference between a
program and a spec: the empty program exhibits no behaviors, while the
empty spec admits all behaviors. Roughly speaking, growing a program
typically leads to more behaviors, whereas growing a specification
typically adds constraints that reduce the set of admitted behaviors.

We believe a similar divide applies to diagramming. \ASV{} is like a
``spec'': the default visualizer shows everything. It admits a small
degree of customization (specified through menus and other graphical
elements) in the form of theming; adding theming alters or reduces
some of the output. As this paper has noted, this genericity and
domain-independence causes problems (\cref{s:bottom-up}) but also has virtues (\cref{s:when-viz-fail}).

At the other extreme are drawing systems like D3. They provide full
control over the output, and can address some of the weaknesses we
have noted for both \ASV{} (\cref{s:asv}) and \cnd{} (\cref{s:limitations}). 
However, the user gets nothing ``for free'': every piece of output requires
programmer effort. In that sense, these visualizations resemble a ``program''
as opposed to a spec. Systems like Penrose fall in this end, too, though they provide
much more structure to the program and enable some separation of concerns.

In this context, \cnd{} is ``spec'', not ``program''. It consciously
alters the \ASV{} output, with the empty \cnd{} program leaving the
output unchanged. Adding constraints and directives shapes and
sometimes limits the amount of output and the number of specs or
instances that can be displayed.

Another useful typology is that introduced by the Scratch programming
language~\cite{maloney2010scratch}, of ``floors'' and ``ceilings''. Scratch presents
itself as having a low floor and high ceiling: it's very easy to start
programming but there is little limit to how sophisticated a program
one can write. Without dwelling on Scratch's claims, we can view the
systems we have discussed through the same lens. \ASV{} has the lowest
floor of all---one needs to do nothing to obtain output---but
it also has a very low ceiling. Systems like D3 and
Penrose have an arbitrarily high ceiling, but also an elevated
floor (which in D3 is known to cause difficulties for some users;
there is very little evidence either way about Penrose). We would
classify \cnd{} as having a very low floor (since, in the limiting
case, it defaults to the \ASV{}), but also a moderate ceiling: it does
some things very well, but other things (\cref{s:limitations}) poorly relative to a
custom visualization.

Both these analyses demonstrate that \cnd{} is not meant to be a last
word, but rather an interesting point in a large space. It is unclear
how to make the ceiling for \cnd{} much higher, at least not without
``lifting'' the floor. This is very difficult in the current design
(which consciously modifies \ASV{} output) and may cause problems
such as masking bad-instances. Instead, we suspect there is value to
having other languages that can provide higher ceilings.

It is natural to ask if an LLM could be used to generate \cnd{} specs
automatically. While this is an interesting idea, we are skeptical that it would
be productive. The overhead of adding \cnd{} constraints and directives is small,
and introducing an LLM could risk introducing a new source of error or inaccuracy.
This is particularly problematic in the context of formal methods,
where correctness and precision are critical. 
While we haven't pursued this avenue in the current paper, 
it's an intriguing possibility that could warrant further
investigation in future work.

\section{Supplementary Material}
\label{s:supplement}

The paper has two supplements: the interactive visualizations and the
study instruments.

To run the visualizations, install
\href{https://docs.docker.com/engine/install/}{Docker}.
You can then access the supplement by running \code{docker pull sidprasad/cnd:latest} 
and then \code{docker run --rm -it -p 3000:3000 sidprasad/cnd:latest}.
This makes \cnd{} available at \url{https://localhost:3000}. All paper examples
are available at
\begin{center}
  \url{https://localhost:3000/example}
\end{center}
and specific examples can be accessed 
at
\begin{center}
  \url{https://localhost:3000/example/<example\_name>}
\end{center}
For instance, the \pill{ab} visualization is available at \url{https://localhost:3000/example/ab}.

The surveys from \cref{s:studies} are provided in the Qualtrics
\texttt{.qsf} format. A QSF file can be imported into Qualtrics as
detailed here:
\begin{center}
  \url{https://www.qualtrics.com/support/survey-platform/survey-module/survey-tools/import-and-export-surveys/}
\end{center}

\section*{Acknowledgments}

We are deeply grateful to Rob Goldstone for gently educating us about
cognitive science. We also thank Skyler Austen and Ji Won Chung for
their feedback on study design and supplementary material.
This work is partially supported by US NSF grant DGE-2208731.

\clearpage
\bibliographystyle{ACM-Reference-Format}
\bibliography{bib.bib,../../bibs/tn}


\begin{thebibliography}{50}


\ifx \showCODEN    \undefined \def \showCODEN     #1{\unskip}     \fi
\ifx \showDOI      \undefined \def \showDOI       #1{#1}\fi
\ifx \showISBNx    \undefined \def \showISBNx     #1{\unskip}     \fi
\ifx \showISBNxiii \undefined \def \showISBNxiii  #1{\unskip}     \fi
\ifx \showISSN     \undefined \def \showISSN      #1{\unskip}     \fi
\ifx \showLCCN     \undefined \def \showLCCN      #1{\unskip}     \fi
\ifx \shownote     \undefined \def \shownote      #1{#1}          \fi
\ifx \showarticletitle \undefined \def \showarticletitle #1{#1}   \fi
\ifx \showURL      \undefined \def \showURL       {\relax}        \fi
\providecommand\bibfield[2]{#2}
\providecommand\bibinfo[2]{#2}
\providecommand\natexlab[1]{#1}
\providecommand\showeprint[2][]{arXiv:#2}

\bibitem[Badros et~al\mbox{.}(2001)]%
        {badros2001cassowary}
\bibfield{author}{\bibinfo{person}{Greg~J Badros}, \bibinfo{person}{Alan
  Borning}, {and} \bibinfo{person}{Peter~J Stuckey}.}
  \bibinfo{year}{2001}\natexlab{}.
\newblock \showarticletitle{The Cassowary linear arithmetic constraint solving
  algorithm}.
\newblock \bibinfo{journal}{\emph{ACM Transactions on Computer-Human
  Interaction (TOCHI)}} \bibinfo{volume}{8}, \bibinfo{number}{4}
  (\bibinfo{year}{2001}), \bibinfo{pages}{267--306}.
\newblock


\bibitem[Battle et~al\mbox{.}(2022)]%
        {battle2022exploring}
\bibfield{author}{\bibinfo{person}{Leilani Battle}, \bibinfo{person}{Danni
  Feng}, {and} \bibinfo{person}{Kelli Webber}.}
  \bibinfo{year}{2022}\natexlab{}.
\newblock \showarticletitle{Exploring D3 implementation challenges on {Stack
  Overflow}}. In \bibinfo{booktitle}{\emph{2022 IEEE visualization and visual
  analytics (VIS)}}. IEEE, \bibinfo{pages}{1--5}.
\newblock


\bibitem[Bostock et~al\mbox{.}(2011)]%
        {bostock2011d3}
\bibfield{author}{\bibinfo{person}{Michael Bostock}, \bibinfo{person}{Vadim
  Ogievetsky}, {and} \bibinfo{person}{Jeffrey Heer}.}
  \bibinfo{year}{2011}\natexlab{}.
\newblock \showarticletitle{D$^3$ data-driven documents}.
\newblock \bibinfo{journal}{\emph{IEEE transactions on visualization and
  computer graphics}} \bibinfo{volume}{17}, \bibinfo{number}{12}
  (\bibinfo{year}{2011}), \bibinfo{pages}{2301--2309}.
\newblock


\bibitem[Chen(2017)]%
        {chen2017information}
\bibfield{author}{\bibinfo{person}{Hsuanwei~Michelle Chen}.}
  \bibinfo{year}{2017}\natexlab{}.
\newblock \showarticletitle{Information visualization principles, techniques,
  and software}.
\newblock \bibinfo{journal}{\emph{Library technology reports}}
  \bibinfo{volume}{53}, \bibinfo{number}{3} (\bibinfo{year}{2017}),
  \bibinfo{pages}{8--16}.
\newblock


\bibitem[Chesi(1973)]%
        {MBTA1973Map}
\bibfield{author}{\bibinfo{person}{Piergiuliano Chesi}.}
  \bibinfo{year}{1973}\natexlab{}.
\newblock \bibinfo{title}{1973 MBTA Rapid Transit Map Card}.
\newblock
\newblock
\urldef\tempurl%
\url{https://commons.wikimedia.org/wiki/File:1973_MBTA_rapid_transit_map_card.jpg}
\showURL{%
\tempurl}
\newblock
\shownote{Public domain image}.


\bibitem[Dijkstra(1974)]%
        {dijkstra1974self}
\bibfield{author}{\bibinfo{person}{Edsger~W Dijkstra}.}
  \bibinfo{year}{1974}\natexlab{}.
\newblock \showarticletitle{Self-stabilizing systems in spite of distributed
  control}.
\newblock \bibinfo{journal}{\emph{Commun. ACM}} \bibinfo{volume}{17},
  \bibinfo{number}{11} (\bibinfo{year}{1974}), \bibinfo{pages}{643--644}.
\newblock


\bibitem[Dwyer(2017)]%
        {webcola}
\bibfield{author}{\bibinfo{person}{Tim Dwyer}.}
  \bibinfo{year}{2017}\natexlab{}.
\newblock \bibinfo{title}{{cola.js}: Constraint-Based Layout in the Browser}.
\newblock
\newblock
\urldef\tempurl%
\url{https://ialab.it.monash.edu/webcola/}
\showURL{%
\tempurl}
\newblock
\shownote{Accessed: 2024-12-02}.


\bibitem[Dyer(2024)]%
        {dyer_sterling_js_demo}
\bibfield{author}{\bibinfo{person}{Tristan Dyer}.}
  \bibinfo{year}{2024}\natexlab{}.
\newblock \bibinfo{title}{Sterling JS Demo}.
\newblock
\newblock
\urldef\tempurl%
\url{https://sterling-js.github.io/demo/}
\showURL{%
\tempurl}
\newblock
\shownote{Accessed: 2024-11-12}.


\bibitem[Dyer and Baugh(2021)]%
        {dyer2021sterling}
\bibfield{author}{\bibinfo{person}{Tristan Dyer} {and} \bibinfo{person}{John
  Baugh}.} \bibinfo{year}{2021}\natexlab{}.
\newblock \showarticletitle{Sterling: A web-based visualizer for relational
  modeling languages}. In \bibinfo{booktitle}{\emph{International Conference on
  Rigorous State-Based Methods}}. Springer, \bibinfo{pages}{99--104}.
\newblock


\bibitem[Dyer et~al\mbox{.}(2022)]%
        {posvalueofnegativeinfo}
\bibfield{author}{\bibinfo{person}{Tristan Dyer}, \bibinfo{person}{Tim Nelson},
  \bibinfo{person}{Kathi Fisler}, {and} \bibinfo{person}{Shriram
  Krishnamurthi}.} \bibinfo{year}{2022}\natexlab{}.
\newblock \showarticletitle{Applying cognitive principles to model-finding
  output: the positive value of negative information}.
\newblock \bibinfo{journal}{\emph{Proceedings of the ACM on Programming
  Languages}} \bibinfo{volume}{6}, \bibinfo{number}{OOPSLA1}
  (\bibinfo{year}{2022}), \bibinfo{pages}{1--29}.
\newblock


\bibitem[Goldstone(1994)]%
        {goldstone1994efficient}
\bibfield{author}{\bibinfo{person}{Robert Goldstone}.}
  \bibinfo{year}{1994}\natexlab{}.
\newblock \showarticletitle{An efficient method for obtaining similarity data}.
\newblock \bibinfo{journal}{\emph{Behavior Research Methods, Instruments, \&
  Computers}}  \bibinfo{volume}{26} (\bibinfo{year}{1994}),
  \bibinfo{pages}{381--386}.
\newblock


\bibitem[Hoffswell et~al\mbox{.}(2018)]%
        {setcola}
\bibfield{author}{\bibinfo{person}{Jane Hoffswell}, \bibinfo{person}{Alan
  Borning}, {and} \bibinfo{person}{Jeffrey Heer}.}
  \bibinfo{year}{2018}\natexlab{}.
\newblock \showarticletitle{Setcola: High-level constraints for graph layout}.
  In \bibinfo{booktitle}{\emph{Computer Graphics Forum}},
  Vol.~\bibinfo{volume}{37}. Wiley Online Library, \bibinfo{pages}{537--548}.
\newblock


\bibitem[Jackson(2012)]%
        {j-alloy-2012}
\bibfield{author}{\bibinfo{person}{Daniel Jackson}.}
  \bibinfo{year}{2012}\natexlab{}.
\newblock \bibinfo{booktitle}{\emph{Software Abstractions: Logic, Language, and
  Analysis} (\bibinfo{edition}{2} ed.)}.
\newblock \bibinfo{publisher}{{MIT} Press}.
\newblock
\showISBNx{978-0-262-10114-1}


\bibitem[Jackson and Wing(1996)]%
        {jackson1996lightweight}
\bibfield{author}{\bibinfo{person}{D. Jackson} {and} \bibinfo{person}{J.
  Wing}.} \bibinfo{year}{1996}\natexlab{}.
\newblock \showarticletitle{Lightweight Formal Methods}.
\newblock \bibinfo{journal}{\emph{IEEE Computer}} (\bibinfo{date}{April}
  \bibinfo{year}{1996}), \bibinfo{pages}{21--22}.
\newblock


\bibitem[Koffka(1922)]%
        {koffka1922perception}
\bibfield{author}{\bibinfo{person}{Kurt Koffka}.}
  \bibinfo{year}{1922}\natexlab{}.
\newblock \showarticletitle{Perception: an introduction to the
  Gestalt-Theorie.}
\newblock \bibinfo{journal}{\emph{Psychological bulletin}}
  \bibinfo{volume}{19}, \bibinfo{number}{10} (\bibinfo{year}{1922}),
  \bibinfo{pages}{531}.
\newblock


\bibitem[Krings et~al\mbox{.}(2018)]%
        {krings++:abz18:alloy-to-b}
\bibfield{author}{\bibinfo{person}{Sebastian Krings}, \bibinfo{person}{Joshua
  Schmidt}, \bibinfo{person}{Carola Brings}, \bibinfo{person}{Marc Frappier},
  {and} \bibinfo{person}{Michael Leuschel}.} \bibinfo{year}{2018}\natexlab{}.
\newblock \showarticletitle{A Translation from {Alloy} to {B}}. In
  \bibinfo{booktitle}{\emph{Conference on Abstract State Machines, {Alloy},
  {B}, and {Z}}}. \bibinfo{pages}{71--86}.
\newblock
\showISBNx{978-3-319-91271-4}
\urldef\tempurl%
\url{https://doi.org/10.1007/978-3-319-91271-4_6}
\showDOI{\tempurl}


\bibitem[Ladenberger and Leuschel(2016)]%
        {ladenberger+leuschel:sefm16:bmotionweb}
\bibfield{author}{\bibinfo{person}{Lukas Ladenberger} {and}
  \bibinfo{person}{Michael Leuschel}.} \bibinfo{year}{2016}\natexlab{}.
\newblock \showarticletitle{{BMotionWeb}: A Tool for Rapid Creation of Formal
  Prototypes}. In \bibinfo{booktitle}{\emph{Software Engineering and Formal
  Methods}}. \bibinfo{pages}{403--417}.
\newblock
\showISBNx{978-3-319-41591-8}
\urldef\tempurl%
\url{https://doi.org/10.1007/978-3-319-41591-8_27}
\showDOI{\tempurl}


\bibitem[Larkin and Simon(1987)]%
        {larkin1987diagram}
\bibfield{author}{\bibinfo{person}{Jill~H Larkin} {and}
  \bibinfo{person}{Herbert~A Simon}.} \bibinfo{year}{1987}\natexlab{}.
\newblock \showarticletitle{Why a diagram is (sometimes) worth ten thousand
  words}.
\newblock \bibinfo{journal}{\emph{Cognitive science}} \bibinfo{volume}{11},
  \bibinfo{number}{1} (\bibinfo{year}{1987}), \bibinfo{pages}{65--100}.
\newblock


\bibitem[Leuschel and Butler(2003)]%
        {leuschel+butler:fme03:prob}
\bibfield{author}{\bibinfo{person}{Michael Leuschel} {and}
  \bibinfo{person}{Michael Butler}.} \bibinfo{year}{2003}\natexlab{}.
\newblock \showarticletitle{ProB: A Model Checker for B}. In
  \bibinfo{booktitle}{\emph{International Symposium on Formal Methods ({FM})}},
  \bibfield{editor}{\bibinfo{person}{Keijiro Araki}, \bibinfo{person}{Stefania
  Gnesi}, {and} \bibinfo{person}{Dino Mandrioli}} (Eds.).
\newblock
\showISBNx{978-3-540-45236-2}
\urldef\tempurl%
\url{https://doi.org/10.1007/978-3-540-45236-2_46}
\showDOI{\tempurl}


\bibitem[Ma'ayan et~al\mbox{.}(2020)]%
        {naturaldiagramming}
\bibfield{author}{\bibinfo{person}{Dor Ma'ayan}, \bibinfo{person}{Wode Ni},
  \bibinfo{person}{Katherine Ye}, \bibinfo{person}{Chinmay Kulkarni}, {and}
  \bibinfo{person}{Joshua Sunshine}.} \bibinfo{year}{2020}\natexlab{}.
\newblock \showarticletitle{How domain experts create conceptual diagrams and
  implications for tool design}. In \bibinfo{booktitle}{\emph{Proceedings of
  the 2020 CHI Conference on Human Factors in Computing Systems}}.
  \bibinfo{pages}{1--14}.
\newblock


\bibitem[Maloney et~al\mbox{.}(2010)]%
        {maloney2010scratch}
\bibfield{author}{\bibinfo{person}{John Maloney}, \bibinfo{person}{Mitchel
  Resnick}, \bibinfo{person}{Natalie Rusk}, \bibinfo{person}{Brian Silverman},
  {and} \bibinfo{person}{Evelyn Eastmond}.} \bibinfo{year}{2010}\natexlab{}.
\newblock \showarticletitle{The scratch programming language and environment}.
\newblock \bibinfo{journal}{\emph{ACM Transactions on Computing Education
  (TOCE)}} \bibinfo{volume}{10}, \bibinfo{number}{4} (\bibinfo{year}{2010}),
  \bibinfo{pages}{1--15}.
\newblock


\bibitem[Mansoor et~al\mbox{.}(2023)]%
        {mansoor2023empirical}
\bibfield{author}{\bibinfo{person}{Niloofar Mansoor}, \bibinfo{person}{Hamid
  Bagheri}, \bibinfo{person}{Eunsuk Kang}, {and} \bibinfo{person}{Bonita
  Sharif}.} \bibinfo{year}{2023}\natexlab{}.
\newblock \showarticletitle{An empirical study assessing software modeling in
  {Alloy}}. In \bibinfo{booktitle}{\emph{2023 IEEE/ACM 11th International
  Conference on Formal Methods in Software Engineering (FormaliSE)}}. IEEE,
  \bibinfo{pages}{44--54}.
\newblock


\bibitem[Manual(2024)]%
        {lean:user-widgets}
\bibfield{author}{\bibinfo{person}{Lean Manual}.}
  \bibinfo{year}{2024}\natexlab{}.
\newblock \bibinfo{title}{The user-widgets system}.
\newblock
  \bibinfo{howpublished}{\url{https://lean-lang.org/lean4/doc/examples/widgets.lean.html}}.
\newblock
\newblock
\shownote{[Accessed Nov 19, 2024]}.


\bibitem[Montaghami and Rayside(2017)]%
        {montaghami2017bordeaux}
\bibfield{author}{\bibinfo{person}{Vajih Montaghami} {and}
  \bibinfo{person}{Derek Rayside}.} \bibinfo{year}{2017}\natexlab{}.
\newblock \showarticletitle{Bordeaux: A tool for thinking outside the box}. In
  \bibinfo{booktitle}{\emph{Fundamental Approaches to Software Engineering:
  20th International Conference, FASE 2017, Held as Part of the European Joint
  Conferences on Theory and Practice of Software, ETAPS 2017, Uppsala, Sweden,
  April 22-29, 2017, Proceedings 20}}. Springer, \bibinfo{pages}{22--39}.
\newblock


\bibitem[Moura and Ullrich(2021)]%
        {demoura+ullrich:cade21:lean4}
\bibfield{author}{\bibinfo{person}{Leonardo~de Moura} {and}
  \bibinfo{person}{Sebastian Ullrich}.} \bibinfo{year}{2021}\natexlab{}.
\newblock \showarticletitle{The {Lean 4} Theorem Prover and Programming
  Language}. In \bibinfo{booktitle}{\emph{International Conference on Automated
  Deduction}}. \bibinfo{publisher}{Springer International Publishing},
  \bibinfo{pages}{625--635}.
\newblock
\showISBNx{978-3-030-79876-5}
\urldef\tempurl%
\url{https://doi.org/10.1007/978-3-030-79876-5_37}
\showDOI{\tempurl}


\bibitem[Nair et~al\mbox{.}(2016)]%
        {nair2016interactive}
\bibfield{author}{\bibinfo{person}{Lekha Nair}, \bibinfo{person}{Sujala
  Shetty}, {and} \bibinfo{person}{Siddhanth Shetty}.}
  \bibinfo{year}{2016}\natexlab{}.
\newblock \showarticletitle{Interactive visual analytics on Big Data: Tableau
  vs D3. js}.
\newblock \bibinfo{journal}{\emph{Journal of e-Learning and Knowledge Society}}
  \bibinfo{volume}{12}, \bibinfo{number}{4} (\bibinfo{year}{2016}).
\newblock


\bibitem[Nathan et~al\mbox{.}(2001)]%
        {nathan2001expert}
\bibfield{author}{\bibinfo{person}{Mitchell~J Nathan},
  \bibinfo{person}{Kenneth~R Koedinger}, \bibinfo{person}{Martha~W Alibali},
  {et~al\mbox{.}}} \bibinfo{year}{2001}\natexlab{}.
\newblock \showarticletitle{Expert blind spot: When content knowledge eclipses
  pedagogical content knowledge}. In \bibinfo{booktitle}{\emph{Proceedings of
  the third international conference on cognitive science}},
  Vol.~\bibinfo{volume}{644648}. \bibinfo{pages}{644--648}.
\newblock


\bibitem[Nathan and Petrosino(2003)]%
        {nathan2003expert}
\bibfield{author}{\bibinfo{person}{Mitchell~J Nathan} {and}
  \bibinfo{person}{Anthony Petrosino}.} \bibinfo{year}{2003}\natexlab{}.
\newblock \showarticletitle{Expert blind spot among preservice teachers}.
\newblock \bibinfo{journal}{\emph{American educational research journal}}
  \bibinfo{volume}{40}, \bibinfo{number}{4} (\bibinfo{year}{2003}),
  \bibinfo{pages}{905--928}.
\newblock


\bibitem[Nawrocki et~al\mbox{.}(2023)]%
        {nawrocki++:itp23:proofwidgets}
\bibfield{author}{\bibinfo{person}{Wojciech Nawrocki},
  \bibinfo{person}{Edward~W. Ayers}, {and} \bibinfo{person}{Gabriel Ebner}.}
  \bibinfo{year}{2023}\natexlab{}.
\newblock \showarticletitle{An Extensible User Interface for {Lean 4}}. In
  \bibinfo{booktitle}{\emph{Interactive Theorem Proving}}
  \emph{(\bibinfo{series}{Leibniz International Proceedings in Informatics
  (LIPIcs)}, Vol.~\bibinfo{volume}{268})},
  \bibfield{editor}{\bibinfo{person}{Adam Naumowicz} {and}
  \bibinfo{person}{Ren\'{e} Thiemann}} (Eds.). \bibinfo{publisher}{Schloss
  Dagstuhl -- Leibniz-Zentrum f{\"u}r Informatik}, \bibinfo{address}{Dagstuhl,
  Germany}, \bibinfo{pages}{24:1--24:20}.
\newblock
\showISBNx{978-3-95977-284-6}
\showISSN{1868-8969}
\urldef\tempurl%
\url{https://doi.org/10.4230/LIPIcs.ITP.2023.24}
\showDOI{\tempurl}


\bibitem[Nelson et~al\mbox{.}(2024)]%
        {nelson2024forge}
\bibfield{author}{\bibinfo{person}{Tim Nelson}, \bibinfo{person}{Ben Greenman},
  \bibinfo{person}{Siddhartha Prasad}, \bibinfo{person}{Tristan Dyer},
  \bibinfo{person}{Ethan Bove}, \bibinfo{person}{Qianfan Chen},
  \bibinfo{person}{Charles Cutting}, \bibinfo{person}{Thomas~Del Vecchio},
  \bibinfo{person}{Sidney LeVine}, \bibinfo{person}{Julianne Rudner},
  \bibinfo{person}{Ben Ryjikov}, \bibinfo{person}{Alexander Varga},
  \bibinfo{person}{Andrew Wagner}, \bibinfo{person}{Luke West}, {and}
  \bibinfo{person}{Shriram Krishnamurthi}.} \bibinfo{year}{2024}\natexlab{}.
\newblock \showarticletitle{Forge: A Tool and Language for Teaching Formal
  Methods}.
\newblock \bibinfo{journal}{\emph{Proceedings of the ACM on Programming
  Languages}} \bibinfo{volume}{8}, \bibinfo{number}{OOPSLA1}
  (\bibinfo{year}{2024}), \bibinfo{pages}{613--641}.
\newblock


\bibitem[Neumerkel and Kral(2002)]%
        {neumerkel2002declarative}
\bibfield{author}{\bibinfo{person}{Ulrich Neumerkel} {and}
  \bibinfo{person}{Stefan Kral}.} \bibinfo{year}{2002}\natexlab{}.
\newblock \showarticletitle{Declarative program development in Prolog with
  GUPU}.
\newblock \bibinfo{journal}{\emph{arXiv preprint cs/0207044}}
  (\bibinfo{year}{2002}).
\newblock


\bibitem[Neumerkel et~al\mbox{.}(1997)]%
        {neumerkel1997visualizing}
\bibfield{author}{\bibinfo{person}{Ulrich Neumerkel},
  \bibinfo{person}{Christoph Rettig}, {and} \bibinfo{person}{Christian
  Schallart}.} \bibinfo{year}{1997}\natexlab{}.
\newblock \showarticletitle{Visualizing Solutions with Viewers.}. In
  \bibinfo{booktitle}{\emph{LPE}}. \bibinfo{pages}{43--50}.
\newblock


\bibitem[Pettitt and contributors(2014)]%
        {dagre}
\bibfield{author}{\bibinfo{person}{Chris Pettitt} {and}
  \bibinfo{person}{contributors}.} \bibinfo{year}{2014}\natexlab{}.
\newblock \bibinfo{title}{Dagre: A JavaScript library for directed graph
  layouts}.
\newblock \bibinfo{howpublished}{\url{https://github.com/dagrejs/dagre}}.
\newblock
\newblock
\shownote{Accessed: 2024-11-21}.


\bibitem[Pollock et~al\mbox{.}(2024)]%
        {pollock2024bluefish}
\bibfield{author}{\bibinfo{person}{Josh Pollock}, \bibinfo{person}{Catherine
  Mei}, \bibinfo{person}{Grace Huang}, \bibinfo{person}{Elliot Evans},
  \bibinfo{person}{Daniel Jackson}, {and} \bibinfo{person}{Arvind
  Satyanarayan}.} \bibinfo{year}{2024}\natexlab{}.
\newblock \showarticletitle{Bluefish: Composing Diagrams with Declarative
  Relations}. In \bibinfo{booktitle}{\emph{Proceedings of the 37th Annual ACM
  Symposium on User Interface Software and Technology}}.
  \bibinfo{pages}{1--21}.
\newblock


\bibitem[Purchase(1997)]%
        {purchase1997aesthetic}
\bibfield{author}{\bibinfo{person}{Helen Purchase}.}
  \bibinfo{year}{1997}\natexlab{}.
\newblock \showarticletitle{Which aesthetic has the greatest effect on human
  understanding?}. In \bibinfo{booktitle}{\emph{International Symposium on
  Graph Drawing}}. Springer, \bibinfo{pages}{248--261}.
\newblock


\bibitem[Shannon et~al\mbox{.}(2003)]%
        {shannon2003cytoscape}
\bibfield{author}{\bibinfo{person}{Paul Shannon}, \bibinfo{person}{Andrew
  Markiel}, \bibinfo{person}{Owen Ozier}, \bibinfo{person}{Nitin~S Baliga},
  \bibinfo{person}{Jonathan~T Wang}, \bibinfo{person}{Daniel Ramage},
  \bibinfo{person}{Nada Amin}, \bibinfo{person}{Benno Schwikowski}, {and}
  \bibinfo{person}{Trey Ideker}.} \bibinfo{year}{2003}\natexlab{}.
\newblock \showarticletitle{Cytoscape: a software environment for integrated
  models of biomolecular interaction networks}.
\newblock \bibinfo{journal}{\emph{Genome research}} \bibinfo{volume}{13},
  \bibinfo{number}{11} (\bibinfo{year}{2003}), \bibinfo{pages}{2498--2504}.
\newblock


\bibitem[Sime et~al\mbox{.}(1977)]%
        {sime1977scope}
\bibfield{author}{\bibinfo{person}{Max~E. Sime}, \bibinfo{person}{Thomas R.~G.
  Green}, {and} \bibinfo{person}{DJ Guest}.} \bibinfo{year}{1977}\natexlab{}.
\newblock \showarticletitle{Scope marking in computer conditionals—a
  psychological evaluation}.
\newblock \bibinfo{journal}{\emph{International Journal of Man-Machine
  Studies}} \bibinfo{volume}{9}, \bibinfo{number}{1} (\bibinfo{year}{1977}),
  \bibinfo{pages}{107--118}.
\newblock


\bibitem[Stefik and Ladner(2017)]%
        {quorumlanguage}
\bibfield{author}{\bibinfo{person}{Andreas Stefik} {and}
  \bibinfo{person}{Richard Ladner}.} \bibinfo{year}{2017}\natexlab{}.
\newblock \showarticletitle{The Quorum Programming Language (Abstract Only)}.
  In \bibinfo{booktitle}{\emph{Proceedings of the 2017 ACM SIGCSE Technical
  Symposium on Computer Science Education}} (Seattle, Washington, USA)
  \emph{(\bibinfo{series}{SIGCSE '17})}. \bibinfo{publisher}{Association for
  Computing Machinery}, \bibinfo{address}{New York, NY, USA},
  \bibinfo{pages}{641}.
\newblock
\showISBNx{9781450346986}
\urldef\tempurl%
\url{https://doi.org/10.1145/3017680.3022377}
\showDOI{\tempurl}


\bibitem[Stefik and Siebert(2013)]%
        {stefik2013empirical}
\bibfield{author}{\bibinfo{person}{Andreas Stefik} {and}
  \bibinfo{person}{Susanna Siebert}.} \bibinfo{year}{2013}\natexlab{}.
\newblock \showarticletitle{An empirical investigation into programming
  language syntax}.
\newblock \bibinfo{journal}{\emph{ACM Transactions on Computing Education
  (TOCE)}} \bibinfo{volume}{13}, \bibinfo{number}{4} (\bibinfo{year}{2013}),
  \bibinfo{pages}{1--40}.
\newblock


\bibitem[Stoica et~al\mbox{.}(2001)]%
        {stoica2001chord}
\bibfield{author}{\bibinfo{person}{Ion Stoica}, \bibinfo{person}{Robert
  Morris}, \bibinfo{person}{David Karger}, \bibinfo{person}{M~Frans Kaashoek},
  {and} \bibinfo{person}{Hari Balakrishnan}.} \bibinfo{year}{2001}\natexlab{}.
\newblock \showarticletitle{Chord: A scalable peer-to-peer lookup service for
  internet applications}.
\newblock \bibinfo{journal}{\emph{ACM SIGCOMM computer communication review}}
  \bibinfo{volume}{31}, \bibinfo{number}{4} (\bibinfo{year}{2001}),
  \bibinfo{pages}{149--160}.
\newblock


\bibitem[Stroop(1935)]%
        {stroop1935studies}
\bibfield{author}{\bibinfo{person}{J~Ridley Stroop}.}
  \bibinfo{year}{1935}\natexlab{}.
\newblock \showarticletitle{Studies of interference in serial verbal
  reactions.}
\newblock \bibinfo{journal}{\emph{Journal of experimental psychology}}
  \bibinfo{volume}{18}, \bibinfo{number}{6} (\bibinfo{year}{1935}),
  \bibinfo{pages}{643}.
\newblock


\bibitem[Sweller and Chandler(1991)]%
        {sweller1991evidence}
\bibfield{author}{\bibinfo{person}{John Sweller} {and} \bibinfo{person}{Paul
  Chandler}.} \bibinfo{year}{1991}\natexlab{}.
\newblock \showarticletitle{Evidence for cognitive load theory}.
\newblock \bibinfo{journal}{\emph{Cognition and instruction}}
  \bibinfo{volume}{8}, \bibinfo{number}{4} (\bibinfo{year}{1991}),
  \bibinfo{pages}{351--362}.
\newblock


\bibitem[Tufte and Graves-Morris(1983)]%
        {tufte1983visual}
\bibfield{author}{\bibinfo{person}{Edward~R Tufte} {and}
  \bibinfo{person}{Peter~R Graves-Morris}.} \bibinfo{year}{1983}\natexlab{}.
\newblock \bibinfo{booktitle}{\emph{The visual display of quantitative
  information}}. Vol.~\bibinfo{volume}{2}.
\newblock \bibinfo{publisher}{Graphics press Cheshire, CT}.
\newblock


\bibitem[Tunnell~Wilson et~al\mbox{.}(2018)]%
        {tunnell2018behavior}
\bibfield{author}{\bibinfo{person}{Preston Tunnell~Wilson},
  \bibinfo{person}{Ben Greenman}, \bibinfo{person}{Justin Pombrio}, {and}
  \bibinfo{person}{Shriram Krishnamurthi}.} \bibinfo{year}{2018}\natexlab{}.
\newblock \showarticletitle{The behavior of gradual types: a user study}.
\newblock \bibinfo{journal}{\emph{ACM SIGPLAN Notices}} \bibinfo{volume}{53},
  \bibinfo{number}{8} (\bibinfo{year}{2018}), \bibinfo{pages}{1--12}.
\newblock


\bibitem[Tunnell~Wilson et~al\mbox{.}(2017)]%
        {tunnell2017can}
\bibfield{author}{\bibinfo{person}{Preston Tunnell~Wilson},
  \bibinfo{person}{Justin Pombrio}, {and} \bibinfo{person}{Shriram
  Krishnamurthi}.} \bibinfo{year}{2017}\natexlab{}.
\newblock \showarticletitle{Can we crowdsource language design?}. In
  \bibinfo{booktitle}{\emph{Proceedings of the 2017 ACM SIGPLAN International
  Symposium on New Ideas, New Paradigms, and Reflections on Programming and
  Software}}. \bibinfo{pages}{1--17}.
\newblock


\bibitem[Tversky(2001)]%
        {spatialschemastversky}
\bibfield{author}{\bibinfo{person}{Barbara Tversky}.}
  \bibinfo{year}{2001}\natexlab{}.
\newblock \showarticletitle{Spatial schemas in depictions}.
\newblock In \bibinfo{booktitle}{\emph{Spatial schemas and abstract thought}},
  \bibfield{editor}{\bibinfo{person}{M.~Gattis}} (Ed.). \bibinfo{publisher}{The
  MIT Press}, \bibinfo{pages}{79--112}.
\newblock


\bibitem[Werth and Leuschel(2020)]%
        {werth+leuschel:abz2020:visb}
\bibfield{author}{\bibinfo{person}{Michelle Werth} {and}
  \bibinfo{person}{Michael Leuschel}.} \bibinfo{year}{2020}\natexlab{}.
\newblock \showarticletitle{{VisB}: A Lightweight Tool to Visualize Formal
  Models with {SVG} Graphics}. In \bibinfo{booktitle}{\emph{Rigorous State
  Based Methods}}. \bibinfo{pages}{260--265}.
\newblock
\showISBNx{978-3-030-48077-6}
\urldef\tempurl%
\url{https://doi.org/10.1007/978-3-030-48077-6_21}
\showDOI{\tempurl}


\bibitem[White(1969)]%
        {spatialstroop}
\bibfield{author}{\bibinfo{person}{Benjamin~W White}.}
  \bibinfo{year}{1969}\natexlab{}.
\newblock \showarticletitle{Interference in identifying attributes and
  attribute names}.
\newblock \bibinfo{journal}{\emph{Perception \& Psychophysics}}
  \bibinfo{volume}{6} (\bibinfo{year}{1969}), \bibinfo{pages}{166--168}.
\newblock


\bibitem[Wren(2021)]%
        {wren2021animate}
\bibfield{author}{\bibinfo{person}{David Wren}.}
  \bibinfo{year}{2021}\natexlab{}.
\newblock \bibinfo{title}{animate-lean-proofs}.
\newblock
\newblock
\urldef\tempurl%
\url{https://github.com/dwrensha/animate-lean-proofs}
\showURL{%
\tempurl}
\newblock
\shownote{Accessed: 2024-11-22}.


\bibitem[Ye et~al\mbox{.}(2020)]%
        {ye2020penrose}
\bibfield{author}{\bibinfo{person}{Katherine Ye}, \bibinfo{person}{Wode Ni},
  \bibinfo{person}{Max Krieger}, \bibinfo{person}{Dor Ma'ayan},
  \bibinfo{person}{Jenna Wise}, \bibinfo{person}{Jonathan Aldrich},
  \bibinfo{person}{Joshua Sunshine}, {and} \bibinfo{person}{Keenan Crane}.}
  \bibinfo{year}{2020}\natexlab{}.
\newblock \showarticletitle{Penrose: from mathematical notation to beautiful
  diagrams}.
\newblock \bibinfo{journal}{\emph{ACM Transactions on Graphics (TOG)}}
  \bibinfo{volume}{39}, \bibinfo{number}{4} (\bibinfo{year}{2020}),
  \bibinfo{pages}{144--1}.
\newblock


\end{thebibliography}

\end{document}